\documentclass[10pt,twocolumn,letterpaper]{article}

\usepackage[pagenumbers]{cvpr} 

\usepackage[dvipsnames]{xcolor}

\newcommand\blfootnote[1]{%
  \begingroup
  \renewcommand\thefootnote{}\footnote{#1}%
  \addtocounter{footnote}{-1}%
  \endgroup
}

\usepackage{overpic}
\usepackage{arydshln}
\usepackage{fontawesome5}  
\usepackage{svrsymbols}  

\usepackage[accsupp]{axessibility} 

\definecolor{cvprblue}{rgb}{0.21,0.49,0.74}
\usepackage[pagebackref,breaklinks,colorlinks,citecolor=cvprblue]{hyperref}

\def\paperID{5} 
\def\confName{CVPR}
\def\confYear{2024}

\title{An Empty Room is All We Want:\\Automatic Defurnishing of Indoor Panoramas}

\author{
Mira Slavcheva$^\assumption$
\quad
Dave Gausebeck$^\assumption$
\quad
Kevin Chen$^\assumption$
\quad
David Buchhofer
\quad
Azwad Sabik
\quad
Chen Ma\\
Sachal Dhillon
\quad
Olaf Brandt
\quad
Alan Dolhasz \vspace*{2mm} \\ 
Matterport\\
}

\begin{document}
\maketitle
\blfootnote{
\hspace*{-5mm}$^\assumption$ denotes equal contribution.\\
\faEnvelope[regular] {\tt research@matterport.com}
}
\begin{abstract}
We propose a pipeline that leverages Stable Diffusion to improve inpainting results in the context of defurnishing---the removal of furniture items from indoor panorama images.
Specifically, we illustrate how increased context, domain-specific model fine-tuning, and improved image blending can produce high-fidelity inpaints that are geometrically plausible without needing to rely on room layout estimation. 
We demonstrate qualitative and quantitative improvements over other furniture removal techniques.
\end{abstract}
\section{Introduction}
\label{sec:intro}
The recent advancement of digital technologies has revolutionized various industries, and real estate is no exception~\cite{shahzad2022digital}. 
The emergence of \emph{digital twins}---virtual replicas of physical assets---has significantly impacted the way properties are marketed, managed, and visualized~\cite{halmetoja2022role}. Digital twins offer a plethora of benefits, including enhanced decision-making~\cite{kuehn2018digital}, improved design processes~\cite{delgado2021digital}, and immersive experiences for stakeholders~\cite{eyre2018immersive}.

While digital twins excel in replicating physical structures with remarkable accuracy, they often lack the flexibility to adapt to different downstream scenarios and requirements. One critical aspect that remains underexplored is \textit{defurnishing}: removing furnishings and objects from virtual representations of built environments~\cite{gkitsas2021panodr}. Defurnishing offers numerous advantages, such as enabling potential buyers or renters to envision personalized layouts~\cite{wang2023survey}, facilitating interior design experimentation~\cite{kalantari2020virtual}, and providing insights for property evaluation and renovation~\cite{daniotti2022development}. It is a fundamental component of any workflow dealing with digital twins, particularly when captured from a built environment rather than synthesized from the ground up. 

\begin{figure}[t]
\centering
 \begin{overpic}[width=\linewidth]{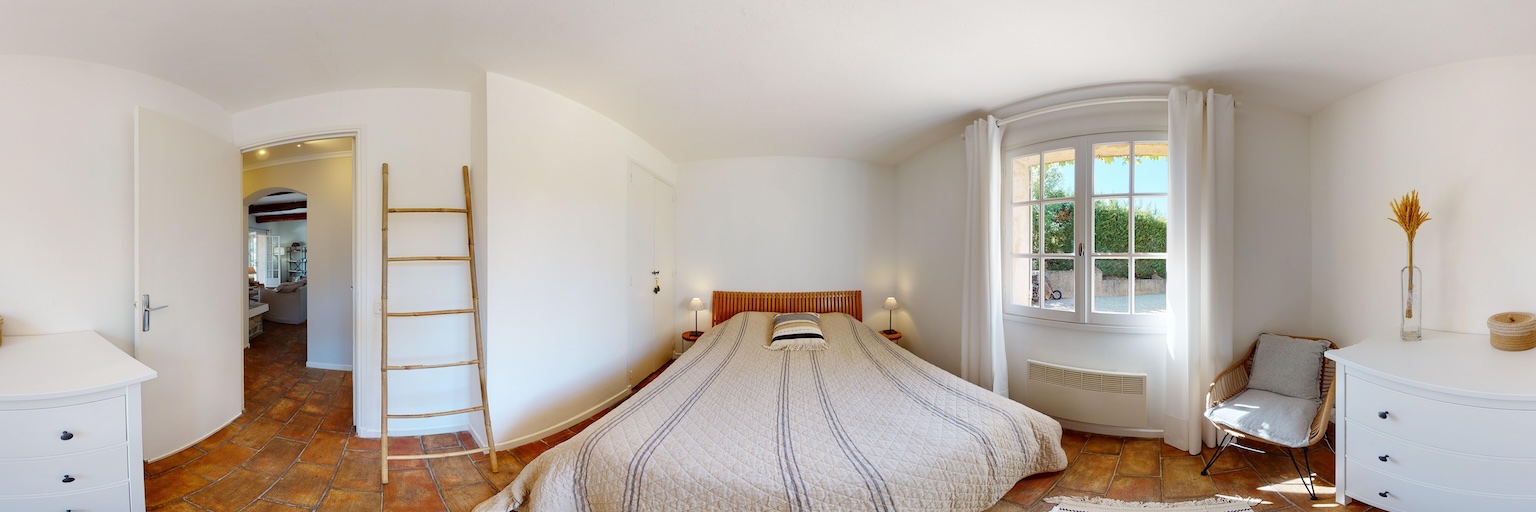}
    \put(2,28){\small{Input panorama}}
 \end{overpic}
 \begin{overpic}[width=\linewidth]{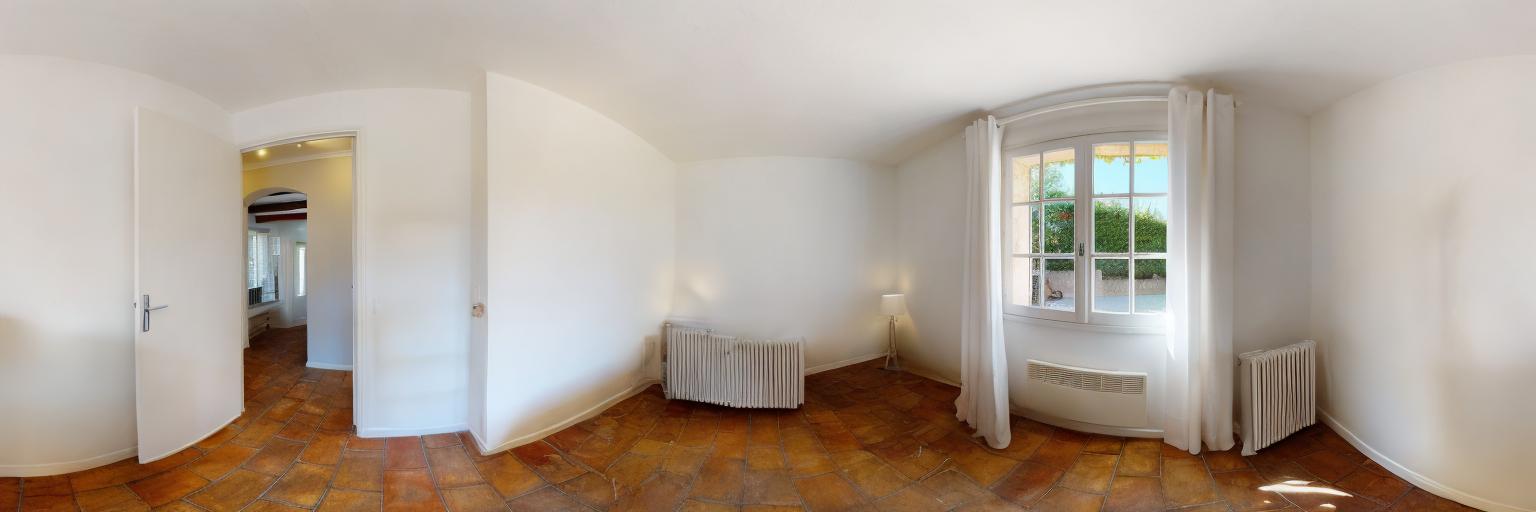}
    \put(2,28){\small{Stable Diffusion 2.0 inpainting}}
 \end{overpic}
 \begin{overpic}[width=\linewidth]{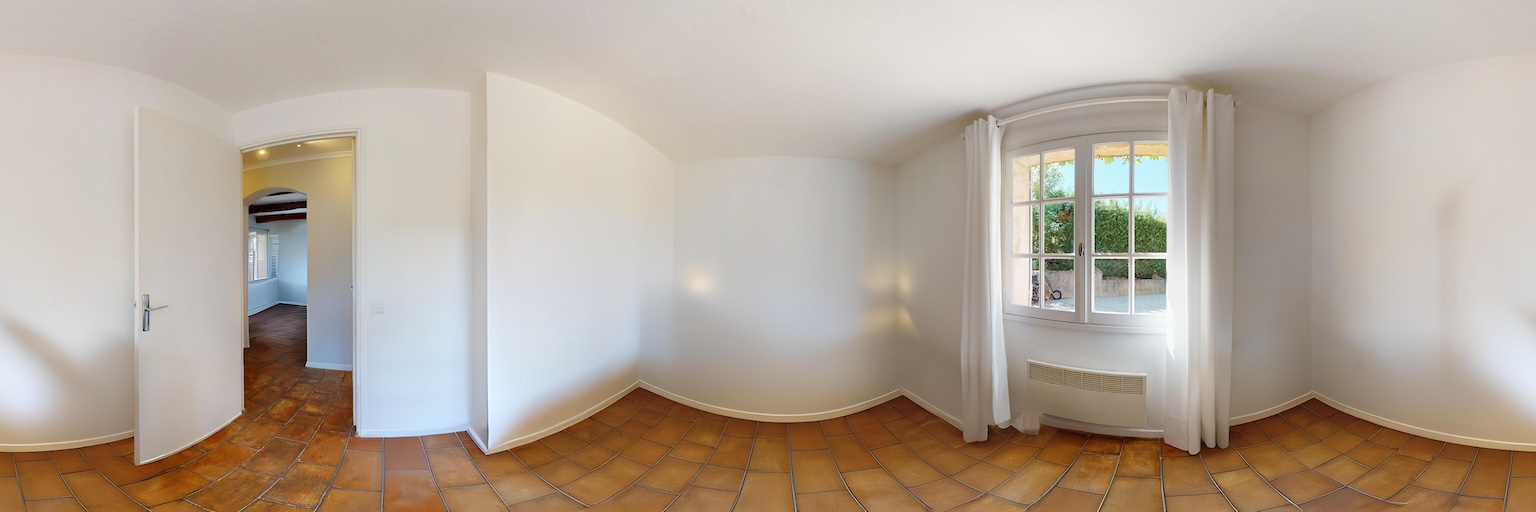}
    \put(2,28){\small{Our pipeline}}
 \end{overpic}
 \caption{\textbf{Indoor panorama defurnishing.} Our custom fine-tuning of Stable Diffusion inpainting reduces its tendency to hallucinate objects near shadows and reflections, such as the radiators on the walls and lamp in the corner.}
 \label{fig:teaser} \vspace*{-0.5mm}
\end{figure}

In this paper, we focus on defurnishing indoor panorama images, leaving the study of defurnishing other common representations of digital twins, such as polygon meshes, to future work.
Additionally, while a digital twin may be composed of several posed panorama images, we focus on defurnishing single images in isolation and do not consider consistency between multiple views.

Recent advances in generative models, such as latent diffusion models~\cite{rombach2022highresolution}, have dramatically improved the quality of image inpainting~\cite{dhariwal2021} by leveraging learned contextual priors to better complete the missing information in an image.
However, as these models are commonly trained to inpaint objects and scene content rather than empty space, spurious inpaints or \emph{hallucinations}~\cite{zhou2024hallucination,liu2024hallucination} tend to be a significant problem when using these models in the context of defurnishing.
Specifically, Stable Diffusion attempts to ``explain away'' contextual information such as shadows and object parts left behind from small imperfections in the inpainting mask by placing objects where empty space  was expected (see Figure~\ref{fig:teaser}). This issue can be partially resolved by applying morphological operations such as dilation to the inpainting mask~\cite{isogawa2018mask}, but this does not adequately handle all cases of shadows, light beams, and other similar effects, which are related to the furniture but not co-located with it \cite{zhang2021shadows}.

Consequently, we present a system for automatic furniture removal, at the core of which is a custom inpainting model that builds over Stable Diffusion in the following ways:
\begin{enumerate}
    \item Leverages a custom dataset of unfurnished panoramas and an augmentation strategy to make the model robust to imperfect masks, cast light beams, reflections, and shadows, significantly reducing hallucinations.
    \item Inpaints equirectangular images natively, significantly increasing the amount of available contextual information compared to perspective images.
    \item Incorporates superresolution and a custom blending strategy to achieve a high-quality, seamless result.
\end{enumerate}
\section{Related work}
\label{sec:related_work}

The key component of our method is an improved inpainting routine, so here we focus on works related to this task.

\subsection{Inpainting}

Digital inpainting is the task of generating pixels to complete missing regions of an image~\cite{bertalmio2000}.
Many inpainting methods have been developed over the past decades, ranging from approaches using classical techniques~\cite{bertalmio2000,criminisi2003,telea2004,osher2005,hays2007,Barnes2009PatchMatchAR} to those leveraging deep neural networks~\cite{pathak2016context,iizuka2017,yu2018generative,liu2018image,song2018spgnet,yu2019freeform,Nazeri_2019_ICCV,zeng2020highresolution, suvorov2021lama,rombach2022highresolution}.

Early learned approaches using Convolutional Neural Networks (CNNs) pioneered the use of Generative Adversarial Networks (GANs) for inpainting~\cite{pathak2016context,iizuka2017,yu2018generative}.
Partial convolutions~\cite{liu2018image} and their learned counterparts, gated convolutions~\cite{yu2019freeform}, allowed inpainting with arbitrary mask shapes by propagating masks through the network layers.
Many models employed a two-stage pipeline that generated intermediate representations to aid in the final inpainting step, such as coarse resolution inpaints~\cite{yu2018generative,yu2019freeform,zeng2020highresolution}, edge maps~\cite{Nazeri_2019_ICCV}, and semantic segmentation masks~\cite{song2018spgnet}.
To handle large masks, fast Fourier convolutions~\cite{suvorov2021lama} were shown to increase the effective receptive field of the neural network and aid in understanding the global structure of the image.

\subsection{Panoramic furniture removal}

Given a panoramic image of an indoor scene, furniture removal generally involves two important steps~\cite{gkitsas2021panodr}:

\begin{enumerate}
    \item Segment all instances of furniture to remove.
    \item Inpaint on the combined furniture masks.
\end{enumerate}

\noindent While an off-the-shelf semantic segmentation model~\cite{long2015fully,ronneberger2015unet,zhao2017pyramid,dosovitskiy2020image,chen2023vit} trained on a suitable ontology~\cite{zhou2017ade20k} is sufficient for the first step,
a number of different approaches~\cite{gkitsas2021panodr,gkitsas2021fulltoempty,geomagical2022,gao2022lgpn,ji2023virtual} have been explored to obtain realistic inpaints using CNN GAN inpainting models.
PanoDR~\cite{gkitsas2021panodr} proposed that in order to produce inpaints that preserve real-life structures, predicting the room layout (\eg~segmentation for floor, walls, and ceiling) is a crucial pre-processing step.
To address perspective distortion and mixing textures coming from different surfaces, Kulshreshtha~\etal\cite{geomagical2022} used the room layout to separate the room into distinct 3D planes~\cite{liu2019planercnn}, then inpainting each plane separately after orthographic reprojection.
Gao~\etal\cite{gao2022lgpn} improved upon this by performing inpainting on each 3D plane in a single end-to-end pipeline.

These approaches all leverage a room layout estimator~\cite{zhang2014,sun2019horizonnet} to explicitly or implicitly guide inpainting.
This makes it difficult to generalize these models to handle certain features common to real-life room layouts, such as curved walls, kitchen islands and half walls, and staircases.
With recent advances in image inpainting techniques, we study whether structural cues from room layout estimation are necessary for creating high-fidelity and geometrically plausible inpaints. 

\begin{figure*}[t]
  \centering
  \includegraphics[width=0.99\linewidth]{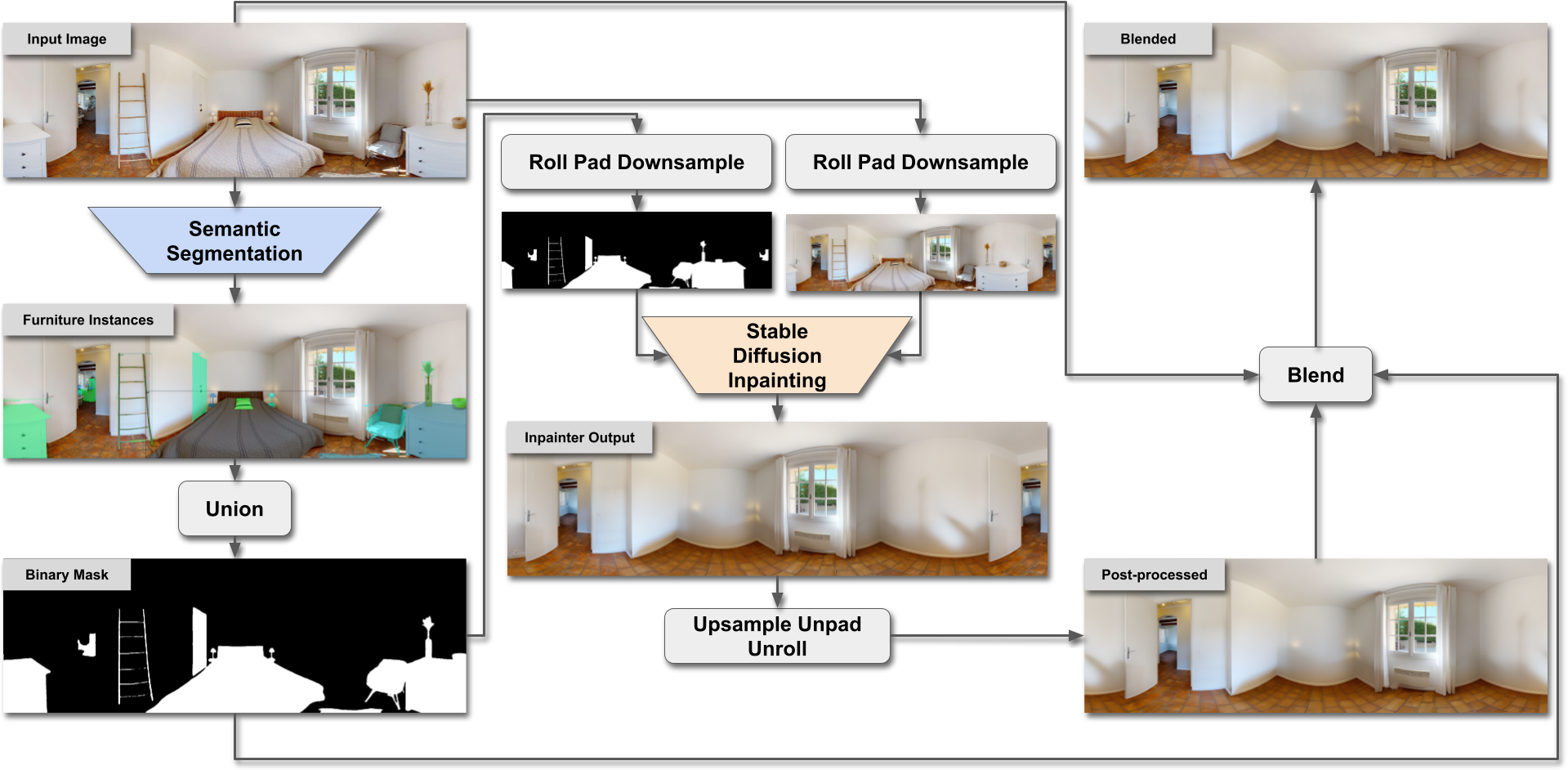}
  \caption{\textbf{Defurnishing pipeline.}
  The input to our system is a 8192$\times$4096 pixel equirectangular panorama, which we crop to 3:1 aspect ratio to exclude the poles. 
  \textbf{Pre-processing:} We obtain a binary furniture mask via semantic segmentation. Both input and mask are rolled, so that inpainting regions are in the center of the image, and padded, to ensure sufficient context (note the repeated doorway and cupboard in the example). The images are then downsampled to a height of 512 pixels.
  \textbf{Inpainting:} Our custom process is robust to inexact masks and remnant shadows, as detailed in the method section.
  \textbf{Post-processing:} We apply 4$\times$ superresolution to the inpainted output and then blend the original and inpainted panoramas into the final result using the mask, keeping as much of the original resolution details as possible.
  }
  \label{fig:pipeline}
\end{figure*}

\subsection{Stable Diffusion}
\label{sec:related_work:sd}

Latent diffusion models, such as Stable Diffusion (SD)~\cite{rombach2022highresolution}, have recently risen to the forefront of image generation and demonstrated results that out-perform their GAN counterparts~\cite{dhariwal2021}.
Diffusion models are readily scalable to better model the complex distribution of training data and can sample diverse inpaints at high fidelity~\cite{lugmayr2022repaint}.
SD is a text-to-image model trained on a large image dataset~\cite{schuhmann2022laion5b} that can also be conditioned by multi-modal inputs, including line contours, depth maps, and other images for image-to-image translation~\cite{zhang2023adding}.

Similar to other generative AI models, like large language models, SD suffers from \emph{hallucinations}---generation of plausible but incorrect information---due to issues such as training data impurities and bias, and attention dilution at inference~\cite{xu2024hallucination,tonmoy2024hallucination}.
Understanding why hallucinations occur and how they can be mitigated are research areas that are critically important for using these generative models for downstream tasks and in our daily lives.

In the case of inpainting for defurnishing, one root cause for hallucinations, which we hypothesize is very significant, is data bias and object co-occurrence~\cite{zhou2024hallucination,liu2024hallucination}.
If the dataset used to train SD contains more images of furnished indoor rooms than unfurnished ones, then when the trained model is presented with an image of an indoor room, it would be biased towards generating inpaints including furniture.
In other words, the dataset may contain a bias where an ``indoor room'' and ``furniture'' often co-occur leading the model to learn this association and spuriously hallucinate furniture in images of indoor rooms.

Regarding dataset bias, one remedy would be to choose text prompts that guide the inpaint towards the desired outcome.
Another approach is to fine-tune~\cite{hu2021lora,ruiz2023dreambooth} the diffusion model on another dataset where this bias has been rectified, trading diversity for fidelity in a particular use-case.
\section{Method}
\label{sec:method}
Our method takes a high resolution panorama image as input and outputs a defurnished version of it, as visualized in Figure~\ref{fig:pipeline}. We start by semantic segmentation of all furniture. Next, we run our custom unfurnished space inpainting. As it is SD-based, it is done at a lower resolution, so finally we run superresolution and blend the inpainted and original images in order to obtain a visually pleasing final result. We elaborate on each of these steps in the following.

\subsection{Pre-processing}

\paragraph{Furniture segmentation}
We use an off-the-shelf semantic segmentation network~\cite{chen2023vit, kirillov2023sam} to identify all instances of furniture classes, as defined in the ADE20K benchmark~\cite{zhou2017ade20k}. Their union is our inpainting mask. 
Note that, in our experience, modern semantic segmentation networks work equally well on equirectangular and perspective images. This spares us the back-and-forth conversion between a panorama and a set of perspective images that other methods necessitate~\cite{ji2023virtual}, yielding a continuous mask and faster processing in our case.

\paragraph{Context maximization}
Next, we ensure that our image is set up optimally for inpainting. To this end, we roll the panorama around horizontally so that the masked objects are as close to the center as possible. There is no guarantee that there will be no masked object on the image edge, so we additionally wrap-pad the image horizontally to provide sufficient context for the ensuing inpainting. 
We apply the same transformations to the corresponding mask image.

\subsection{Robust unfurnished space inpainting}
\label{sec:inpainting}

Figure~\ref{fig:teaser} points to three major issues with SD inpainting that we aim to solve:
\begin{enumerate}
    \item It is not readily applicable to non-square images.
    \item It is prone to hallucinating objects.
    \item It is low resolution and lacks high frequency details.
\end{enumerate}

\noindent Hallucinations are a known issue for large generative models~\cite{xu2024hallucination,liu2024hallucination}. 
The same prompt gives different results with different random seeds, some of which do not exhibit hallucinations, but this is unreliable in an automated system. 
This issue is multi-faceted in the case of inpainting. 
First, inpainting masks may be inaccurate, leaving visible parts of objects which SD picks up and completes into full objects. 
While this can be addressed with mask dilation, valuable context and detail is lost. 
Even with perfect object masks, there are shadows left, which SD aims to explain by inpainting objects that may have cast them.
Second, our case of inpainting empty space is especially challenging, because generative models are good at creating concrete objects with describable properties (such as round wooden kitchen tables), but struggle with generating imagery for vague concepts like \textit{emptiness}.
Negative prompts are another way, but while they can make hallucinated objects look less like the negative prompt, they do not remove the objects.

We fine-tune a version of SD inpainting to address the first two issues. For the third one, we develop a post-processing routine described in Section~\ref{sec:postproc}.

\paragraph{Dataset}
To be able to successfully inpaint \textit{empty space}, we train on a proprietary dataset of 160k equirectangular panoramas of unfurnished residential spaces, similar to Matterport3D~\cite{matterport3d,ramakrishnan2021hm3d}.
As discussed, the use of equirectangular panoramas ensures that we maximize the context at inpainting time. We found that SD quickly adapts to this kind of imagery after being exposed to fewer than 1000 panoramas. We find this strategy to be better than applying SD to a tiled version of the equirect, because all context is present in one image.

\begin{figure}[t]
 \centering
 \includegraphics[clip,trim={0 90 0 90},width=0.495\linewidth]{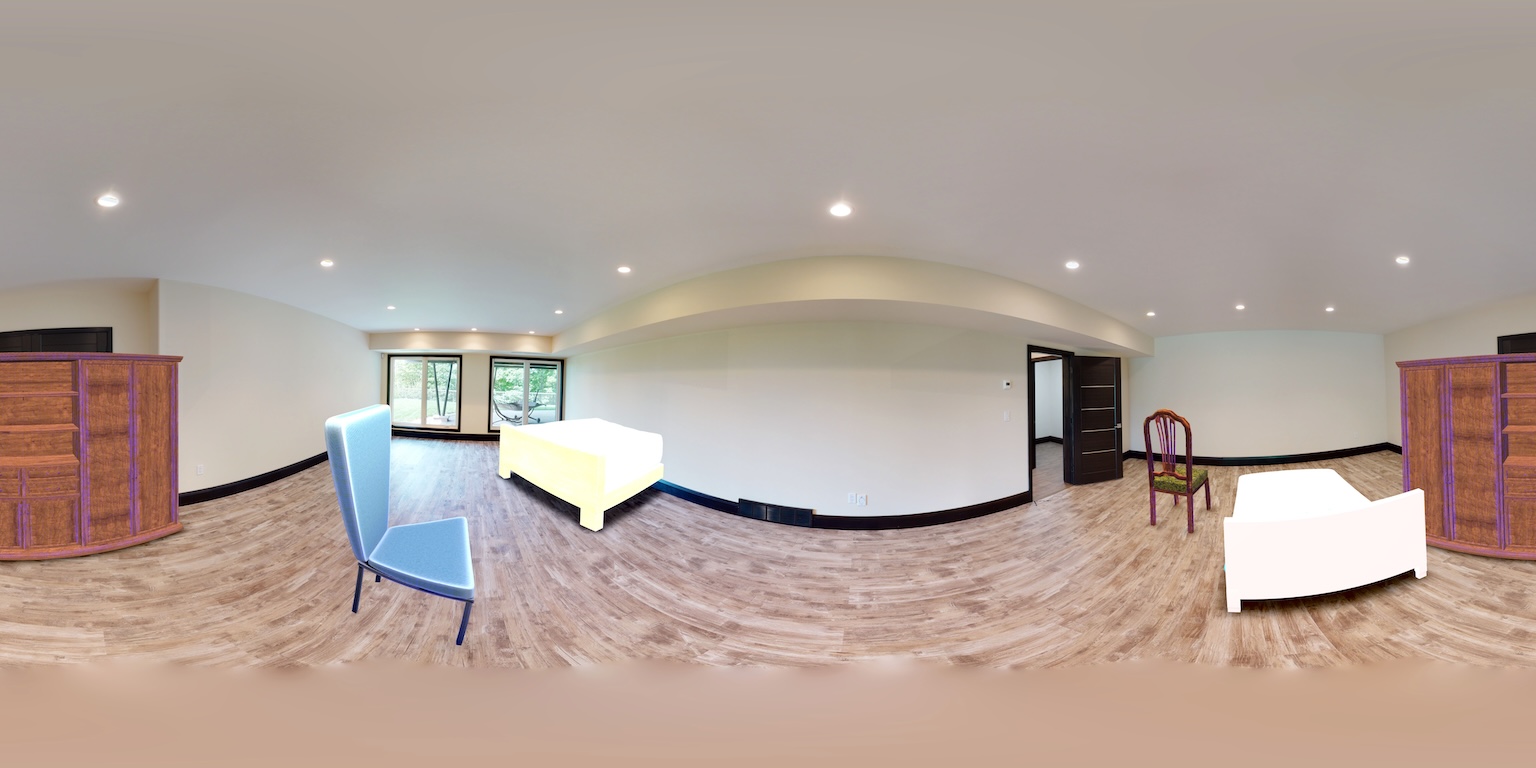}
 \includegraphics[clip,trim={0 90 0 90},width=0.495\linewidth]{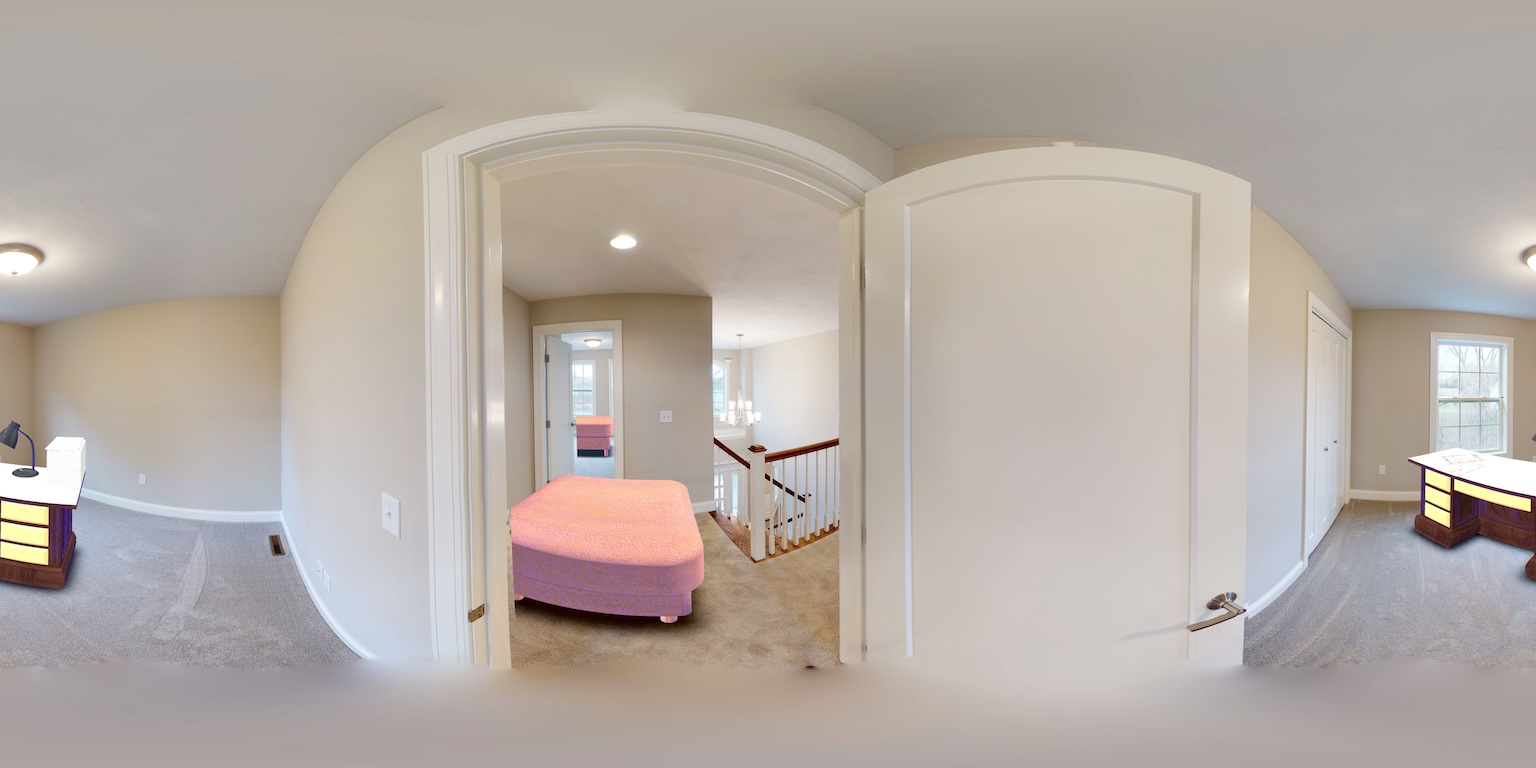}
 \caption{\textbf{Training dataset examples.} Synthetic furniture items and shadows are rendered over real unfurnished panoramas.}
 \label{fig:dataset_with_public_obj_on_public_panos}
\end{figure}

\paragraph{Synthetic data \& augmentations}
To make inpainting robust to shadows and inaccurate masks, we augment the empty spaces with synthetically rendered furniture objects from a dataset like Objaverse~\cite{deitke2022objaverse}. The rendering is not photorealistic, but includes shadows, as shown in Figure~\ref{fig:dataset_with_public_obj_on_public_panos}.
We fine-tune a pre-trained SD inpainting model using as inputs the unfurnished space panoramas with synthetically rendered objects and shadows, and masks that only cover the objects and not the shadows.
The target outputs are the original unfurnished panoramas.
This fine-tuning discourages SD from hallucinating objects when seeking to explain away effects of the furniture on the scene, such as shadows, reflections, or light beams. 
Additionally, to make our model robust to mask inaccuracies, we perturb the input masks to simulate imperfect semantic segmentation output.

\begin{figure}[t]
 \centering
 \begin{overpic}[clip,trim={1000 80 150 170},width=0.325\linewidth]
    {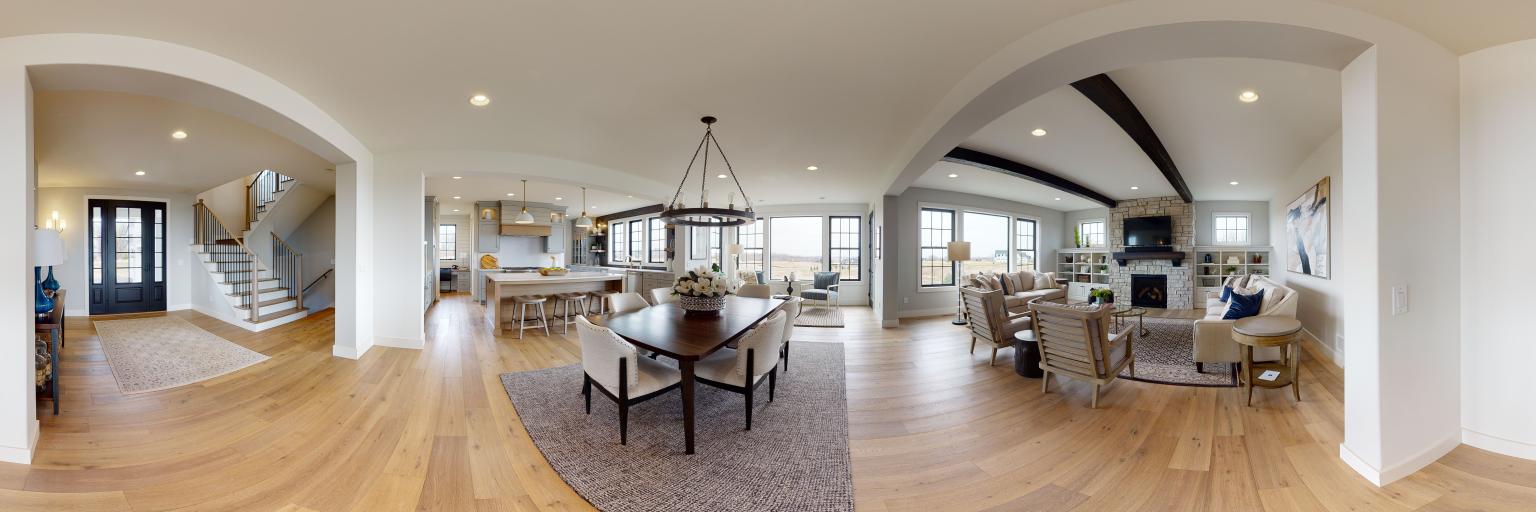}
    \put(5,5){\color{white}\scriptsize{Input}}
 \end{overpic}
 \begin{overpic}[clip,trim={1000 80 150 170},width=0.325\linewidth]
    {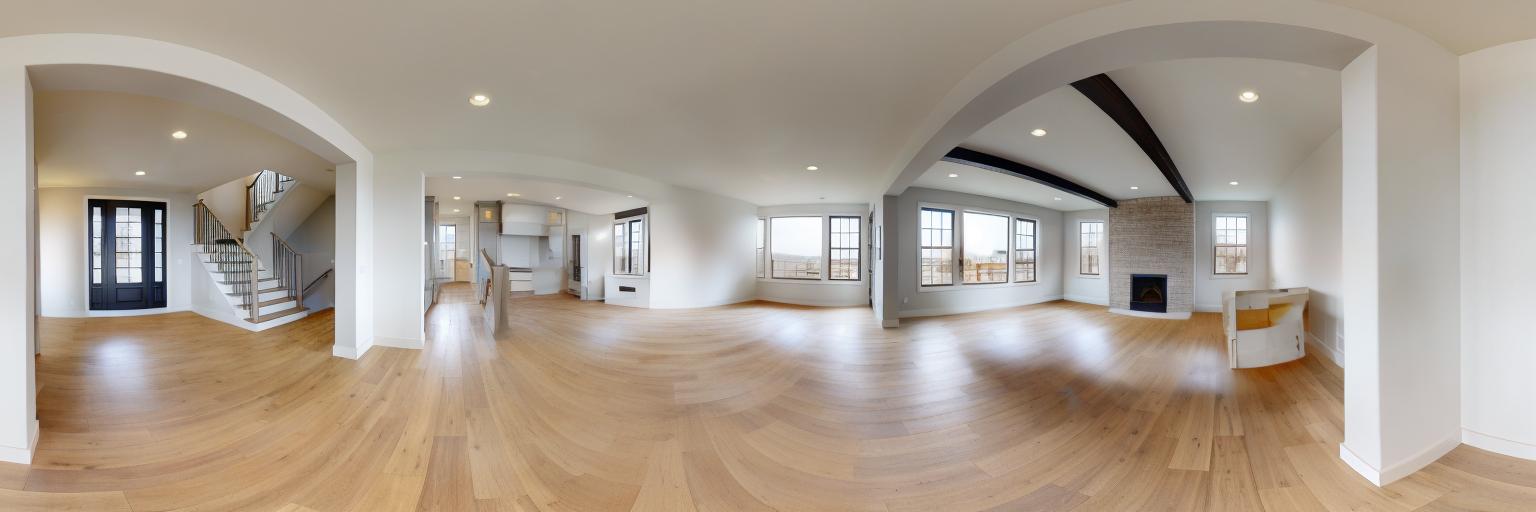}
    \put(5,5){\color{white}\scriptsize{Empty prompt}}
 \end{overpic}
 \begin{overpic}[clip,trim={1000 80 150 170},width=0.325\linewidth]
    {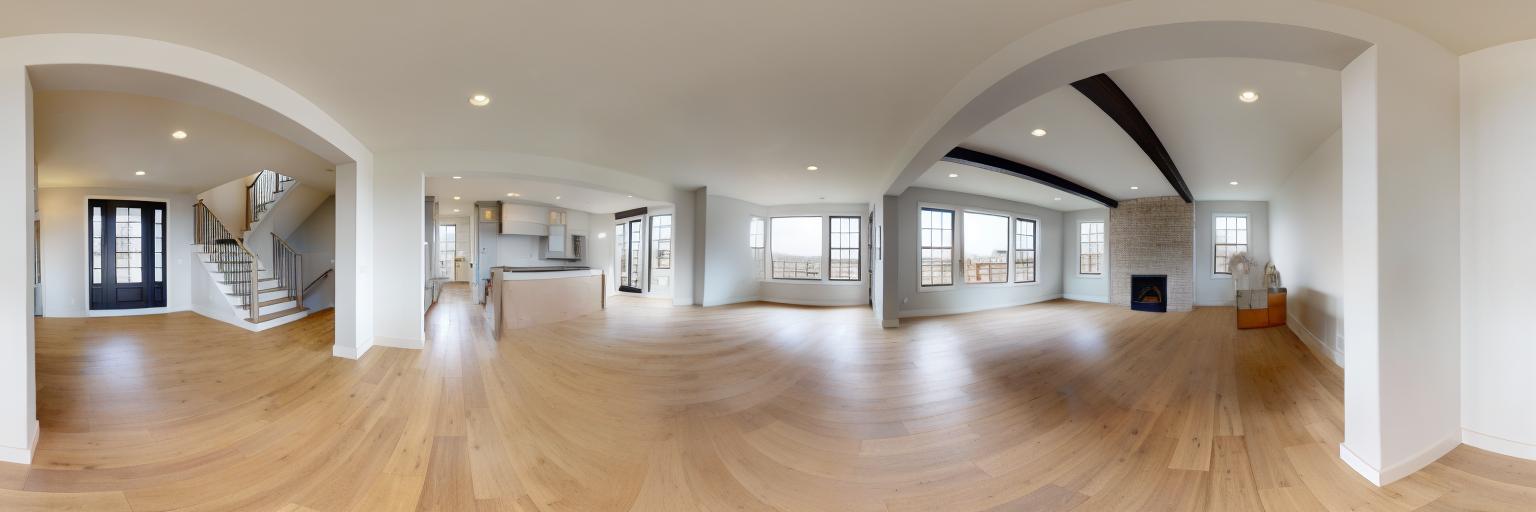}
    \put(5,5){\color{white}\scriptsize{1 prompt}}
 \end{overpic}\\
 \begin{overpic}[clip,trim={1000 80 150 170},width=0.325\linewidth]
    {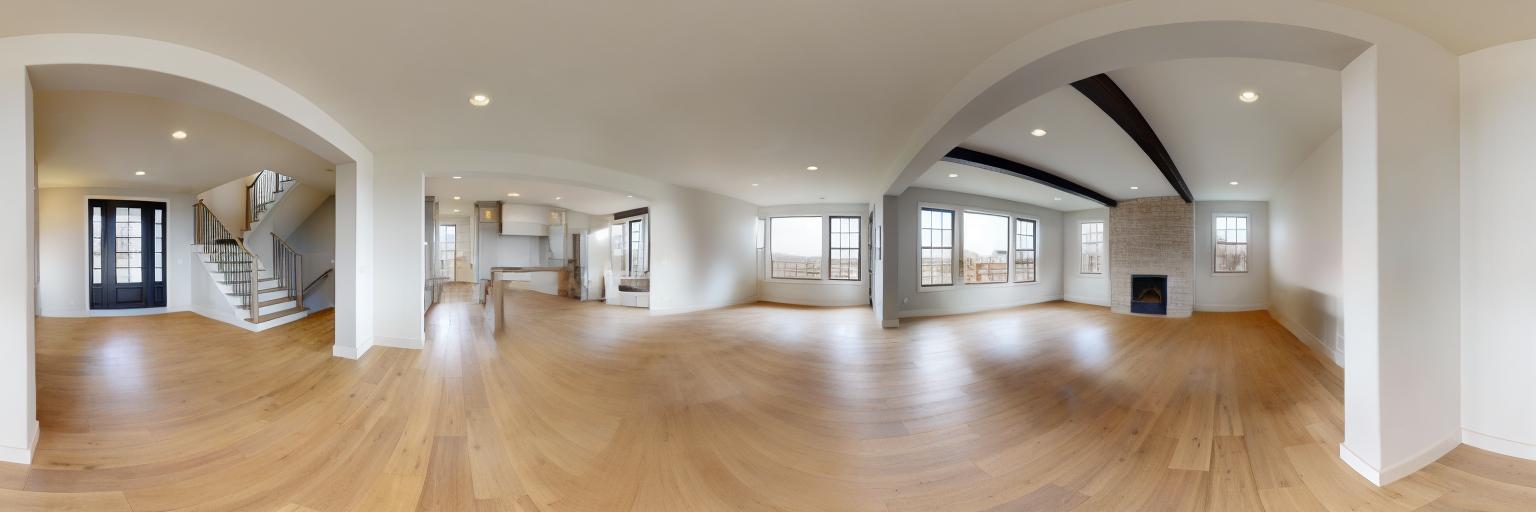}
    \put(5,5){\color{white}\scriptsize{32 prompts}}
 \end{overpic}
 \begin{overpic}[clip,trim={1000 80 150 170},width=0.325\linewidth]
    {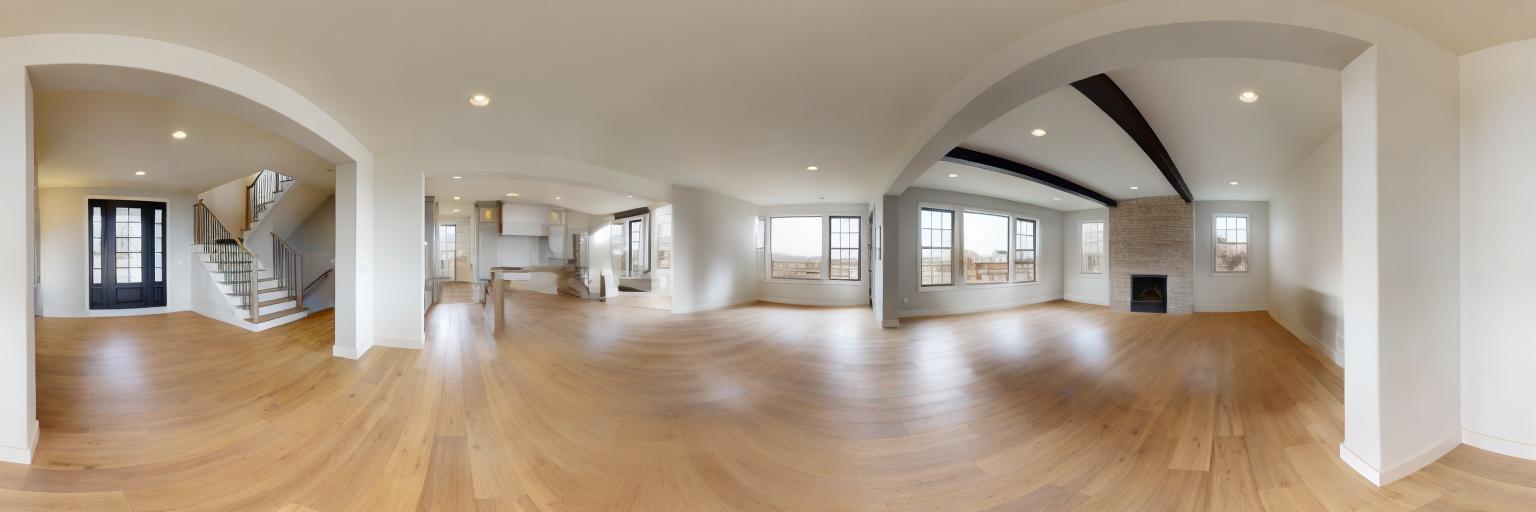}
    \put(5,5){\color{white}\scriptsize{72 prompts}}
 \end{overpic}
 \begin{overpic}[clip,trim={1000 80 150 170},width=0.325\linewidth]
    {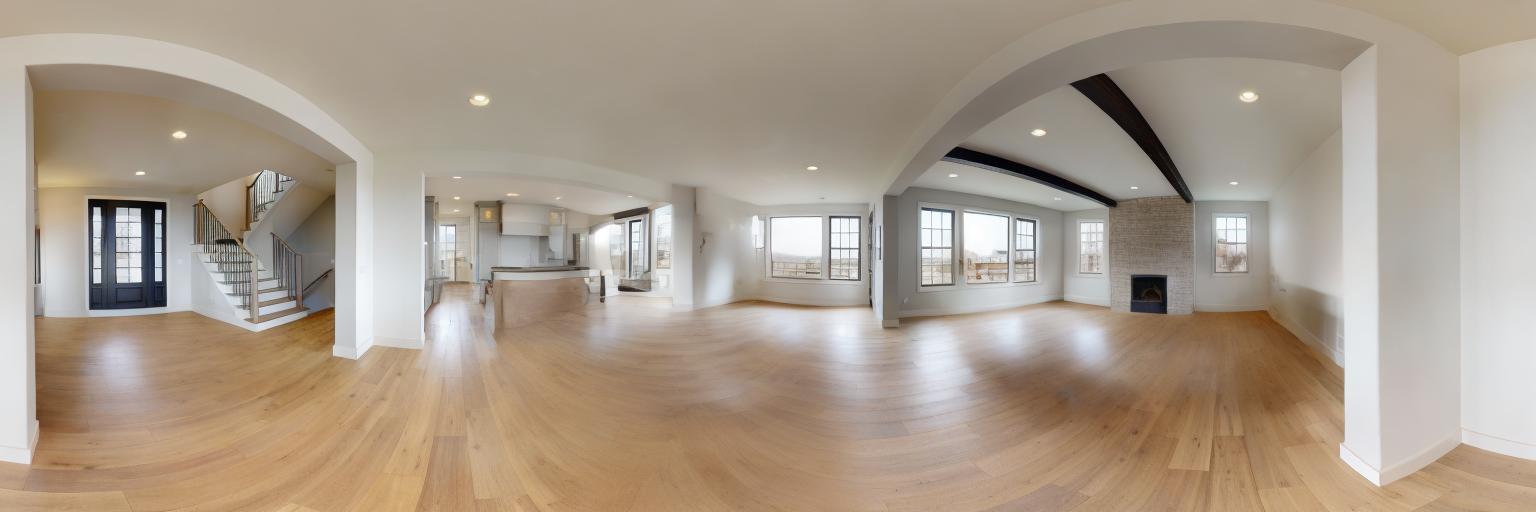}
    \put(5,5){\color{white}\scriptsize{180 prompts}}
 \end{overpic}
 \caption{\textbf{Effect of number of training prompts.} Fewer than 32 prompts lead to hallucinations near the shadow on the right wall.}
 \label{fig:prompts}
\end{figure}

\paragraph{Prompts}
SD is a text-to-image model, and consequently the text prompt(s) used for our inpainting task play(s) an important role. The prompt we use at inference time is \textit{empty room}, but we found that training with a set of similar prompts further reduces hallucinations, as shown in Figure~\ref{fig:prompts}. 
We randomly select one prompt from the set per training sample (prompts for the same sample may differ in different epochs).
The additional prompts contain synonyms, such as unfurnished; space, home, house; as well as appending descriptions such as ``uniformly blank''. We tested with 0 (empty prompt), 1, 8, 32, 72 and 180 prompts. In our experiments 0-8 prompts lead to models that are most prone to hallucination, 32 and 72 yield results with much fewer hallucinations, while more prompts result in increased hallucinations again. Thus, in our experiments we choose to train with 32 variants of the ``empty room'' prompt.

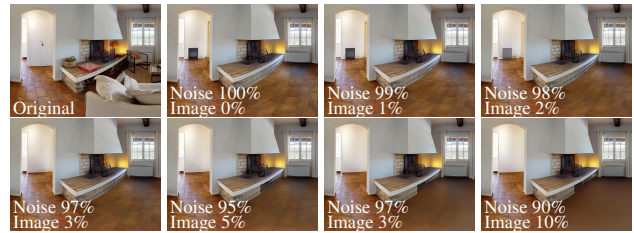
\begin{figure}[t]
 \centering
 \begin{overpic}[clip,trim={700 0 350 150},width=0.24\linewidth]{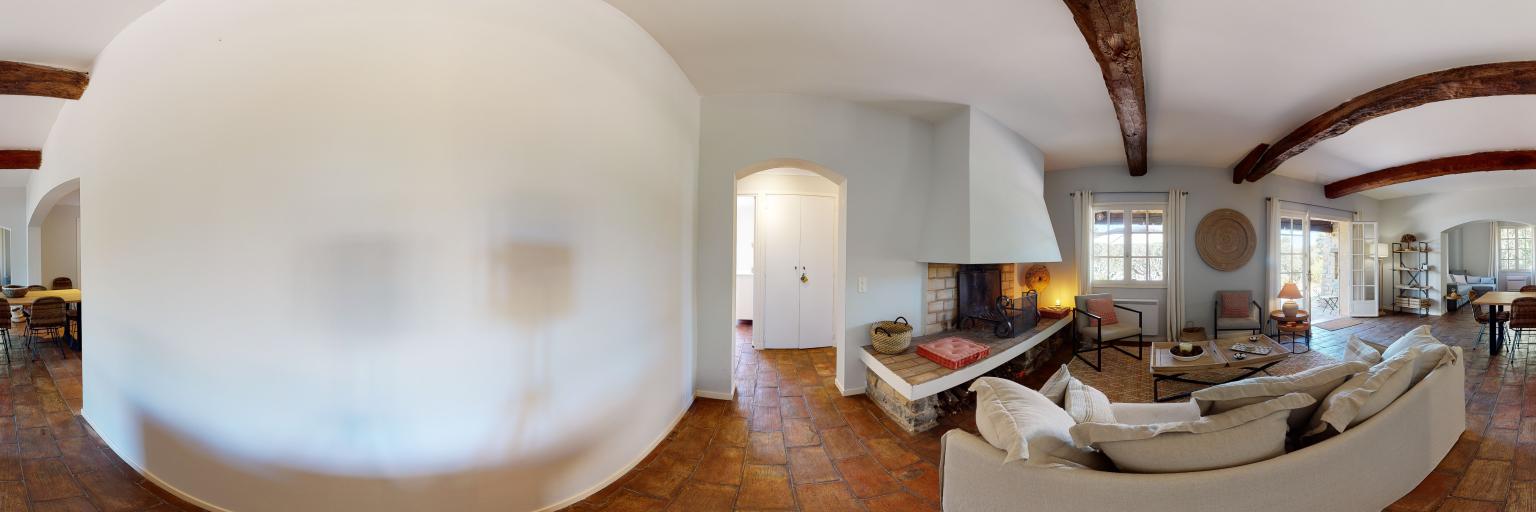}
  \put(2,2){\color{white}\scriptsize{Original}}
 \end{overpic}
 \begin{overpic}[clip,trim={700 0 350 150},width=0.24\linewidth]{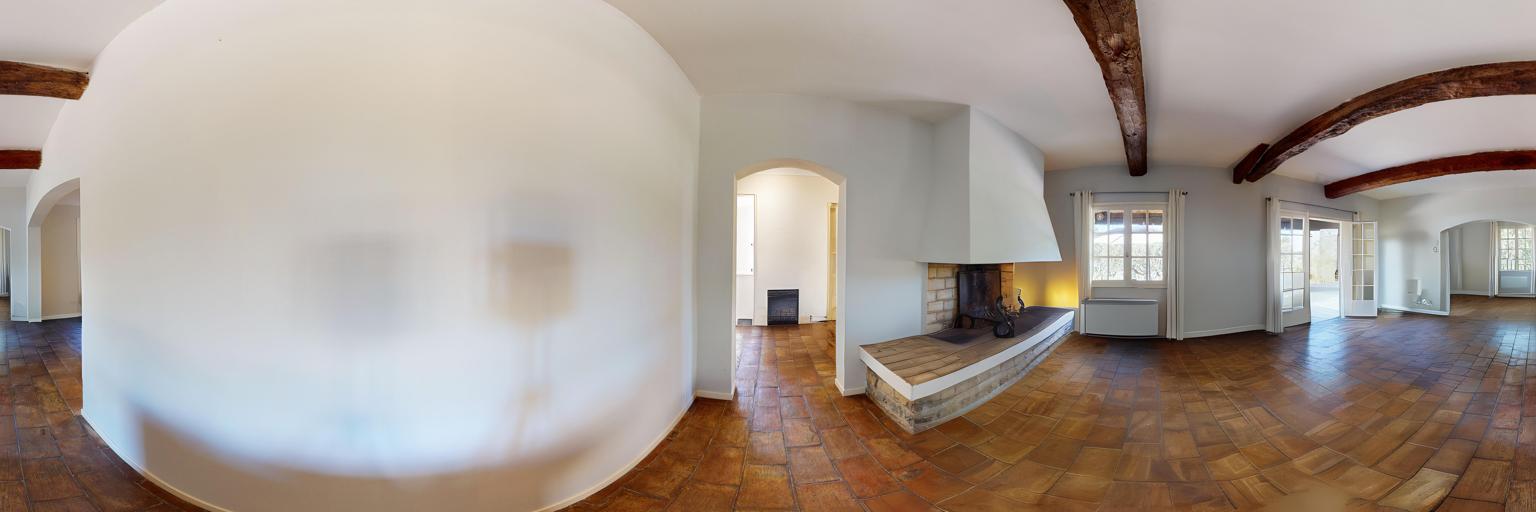}
  \put(2,12){\color{white}\scriptsize{Noise 100\%}}
  \put(2,2){\color{white}\scriptsize{Image   0\%}}
 \end{overpic}
 \begin{overpic}[clip,trim={700 0 350 150},width=0.24\linewidth]{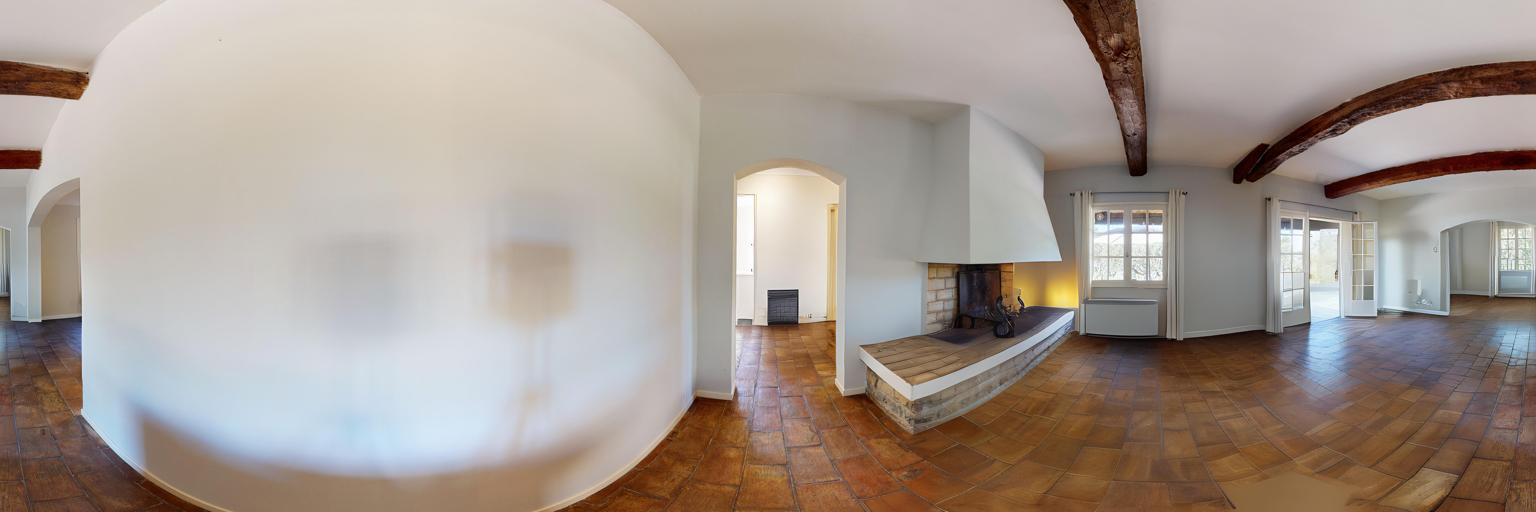}
  \put(2,12){\color{white}\scriptsize{Noise 99\%}}
  \put(2,2){\color{white}\scriptsize{Image  1\%}}
 \end{overpic}
 \begin{overpic}[clip,trim={700 0 350 150},width=0.24\linewidth]{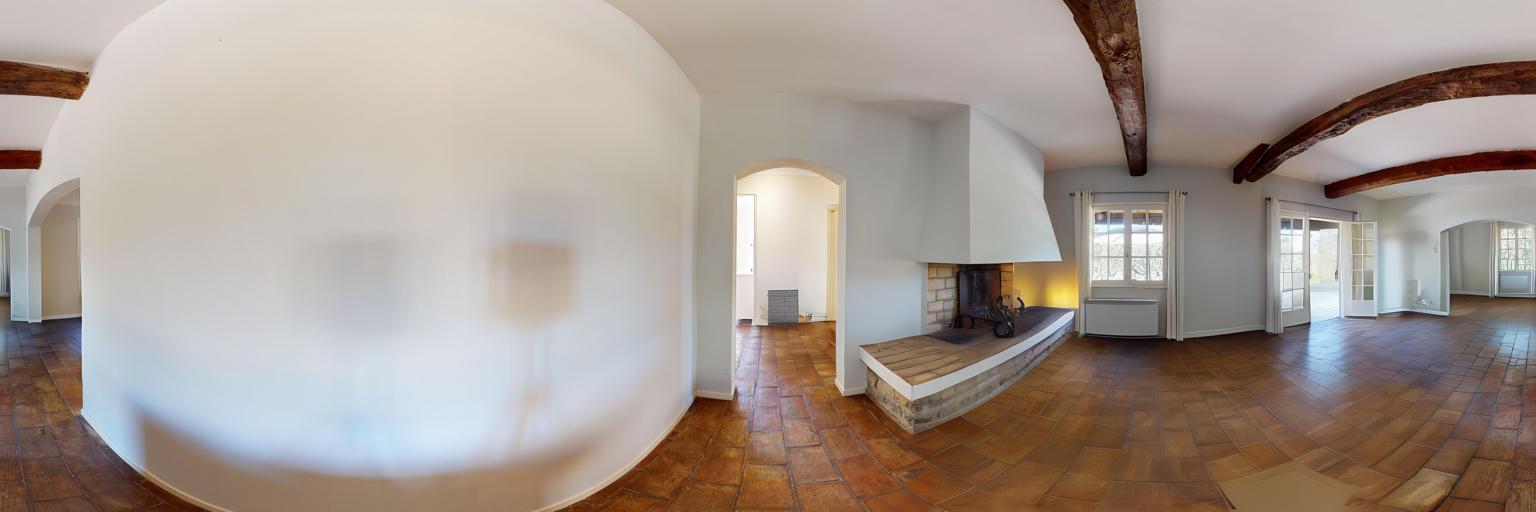}
  \put(2,12){\color{white}\scriptsize{Noise 98\%}}
  \put(2,2){\color{white}\scriptsize{Image  2\%}}
 \end{overpic}\\
 \begin{overpic}[clip,trim={700 0 350 150},width=0.24\linewidth]{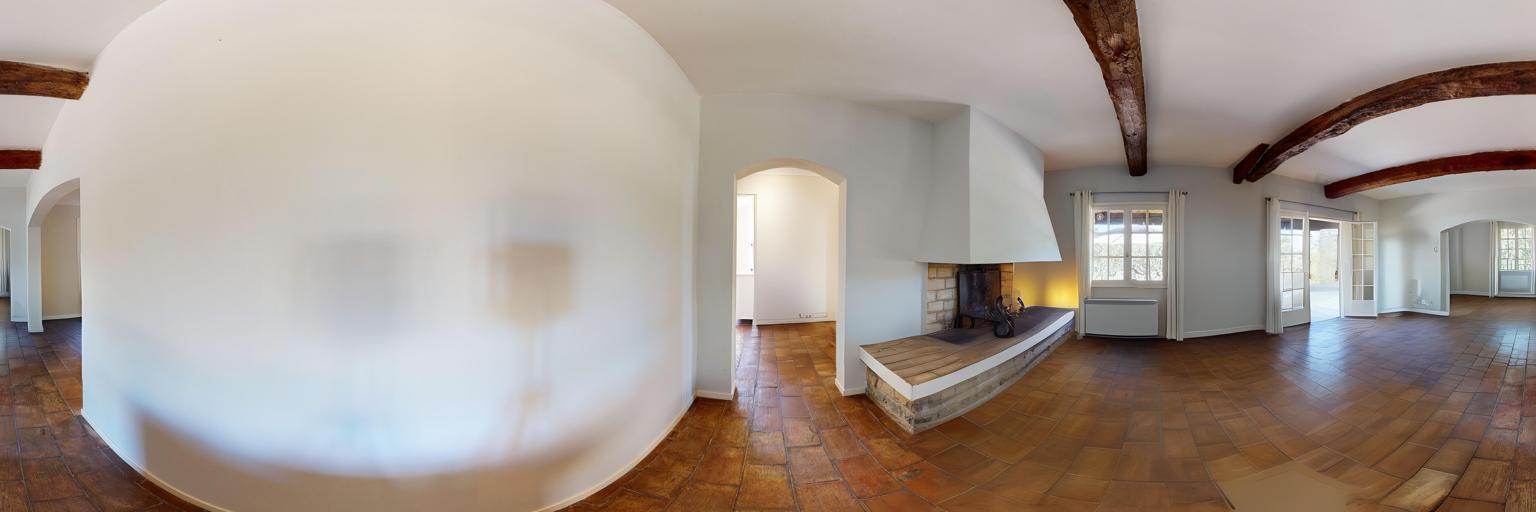}
  \put(2,12){\color{white}\scriptsize{Noise 97\%}}
  \put(2,2){\color{white}\scriptsize{Image  3\%}}
 \end{overpic}
 \begin{overpic}[clip,trim={700 0 350 150},width=0.24\linewidth]{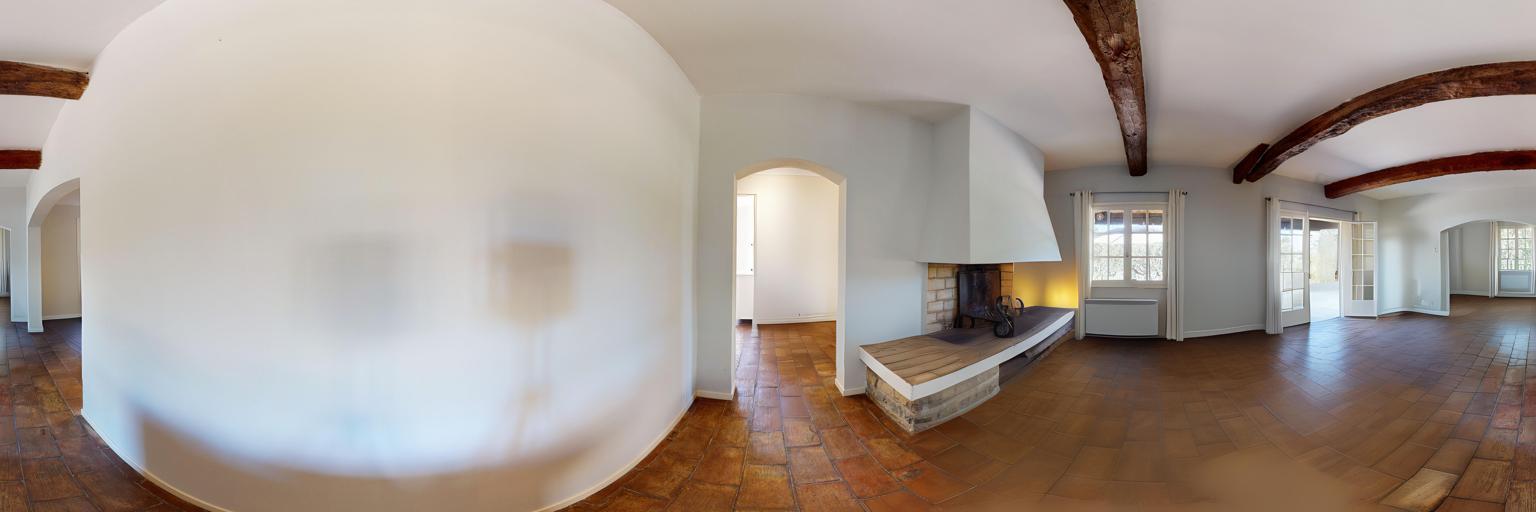}
  \put(2,12){\color{white}\scriptsize{Noise 95\%}}
  \put(2,2){\color{white}\scriptsize{Image  5\%}}
 \end{overpic}
 \begin{overpic}[clip,trim={700 0 350 150},width=0.24\linewidth]{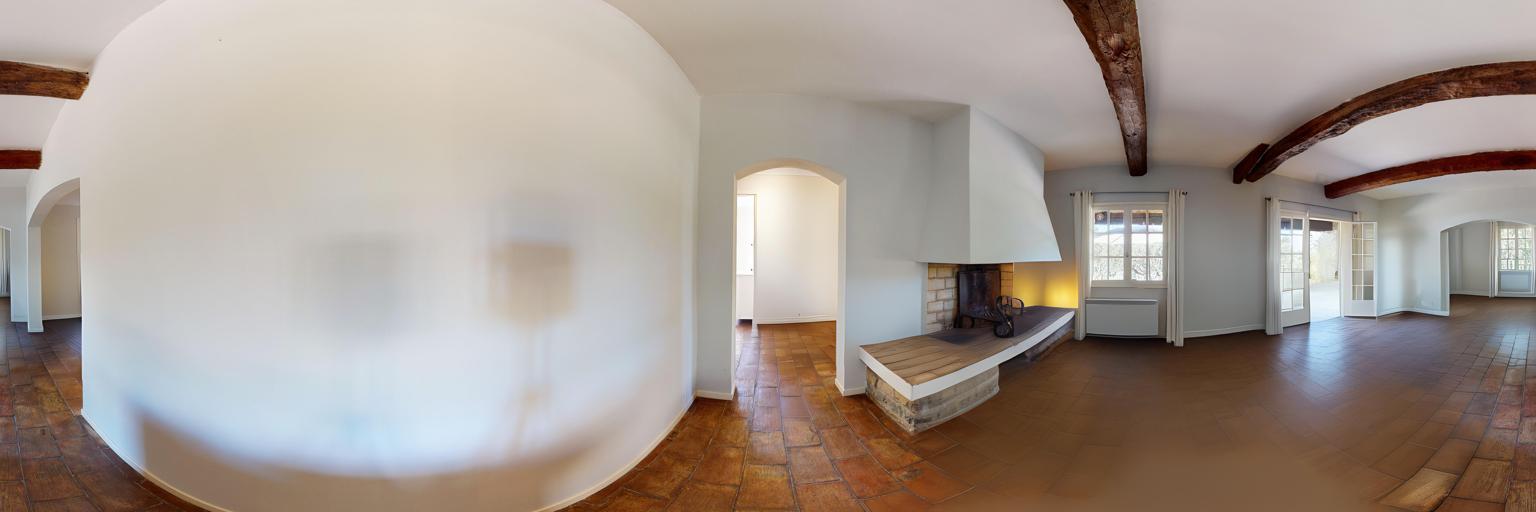}
  \put(2,12){\color{white}\scriptsize{Noise 97\%}}
  \put(2,2){\color{white}\scriptsize{Image  3\%}}
 \end{overpic}
 \begin{overpic}[clip,trim={700 0 350 150},width=0.24\linewidth]{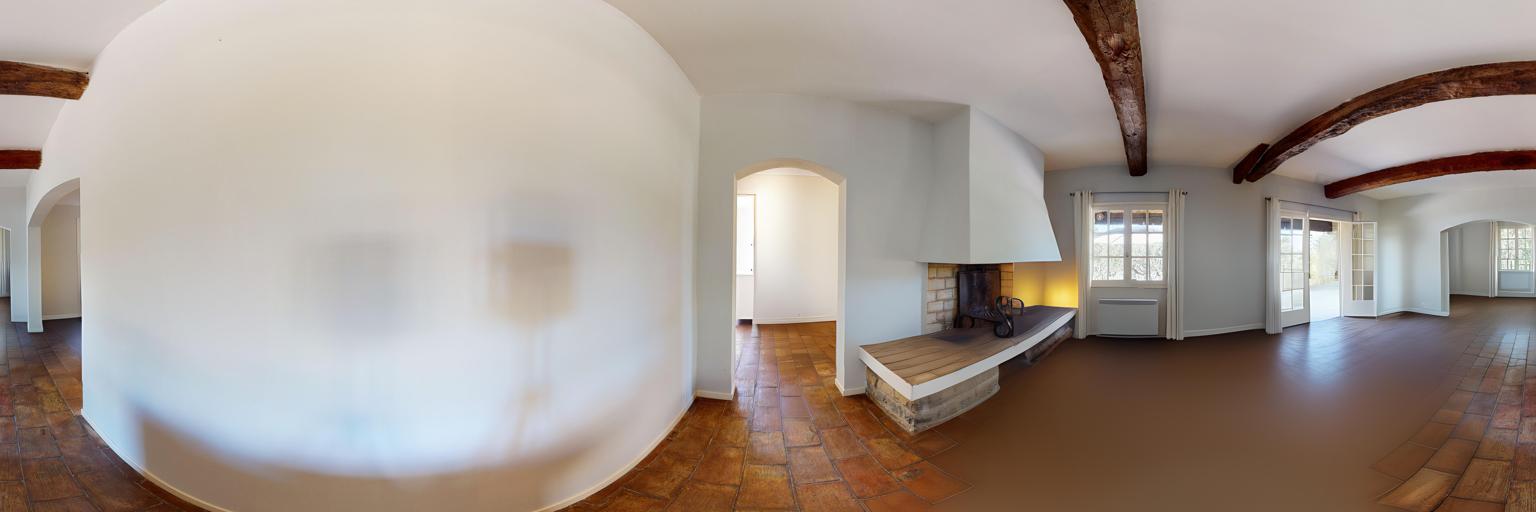}
  \put(2,12){\color{white}\scriptsize{Noise 90\%}}
  \put(2,2){\color{white}\scriptsize{Image 10\%}}
 \end{overpic}
 \caption{\textbf{Initialization of latents for inpainting.} With decreasing percentage of noise in the initialization, hallucinations in the hallway decrease, but blurriness on the inpainted floor increases.}
 \label{fig:ablate_latentinit}
\end{figure}

\paragraph{Initialization}
At inference time we initialize the inpainting latents as a weighted sum of 97\% random noise and 3\% latents based on the input image, blurred under the inpainting mask. 
We empirically found this combination to best minimize both hallucinations and blur, as shown in Figure~\ref{fig:ablate_latentinit}.

\paragraph{Resources}
We train for 96 hours on 8 NVIDIA A10G GPUs (24~GB vRAM) with an effective batch size of 96 (using gradient accumulation). 
The base SD version that we fine-tune is \textit{stabilityai/stable-diffusion-2-inpainting} and we only perform low-rank adaptation (LoRA)~\cite{hu2021lora}, whereby we calculate update matrices for its UNet, rather than training all its weights. 
We experimented with training the entire UNet and the VAE, each of which gave inferior results. This could be due to our dataset size or other factors, which we did not investigate further.
We also fine-tuned SD XL, but found it to be more prone to hallucinations.
 
We only use one GPU for inference. We found that 10 inference steps are sufficient. Our pipeline takes approximately 12 seconds to process a single panorama: 8s for pre-processing and semantic segmentation, 3.5s per image for inpainting alone, and 0.5s for post-processing.

\subsection{Post-processing}\label{sec:postproc}
After inpainting, we apply an off-the-shelf superresolution network~\cite{wang2021realesrgan} to upscale the inpainted image four times. We then undo the padding and rolling pre-processing transformations to restore the original panorama dimensions. 
Finally, we make sure that the inpainted and original textures match well via a custom blending routine.

\paragraph{Blending}
SD-based inpainting tends to lack high frequency details, which is problematic in our setting, where a high-resolution final output is required. Therefore, we aim to preserve as much detail as possible from the original furnished panorama, and develop a tailored blending procedure that uses the binary inpainting mask to select how to combine the original and inpainted images. As explained in Section~\ref{sec:inpainting}, our inpainting strategy is guided by the inpainting mask, but is trained to be robust to shadows by allowing for pixels nearby the mask to also be modified, as enforced by the dataset and training objective. In this way, we not only avoid hallucinating objects due to the remaining shadow, but we also remove the shadow itself from the image, leading to an overall higher-quality inpainted result. Conversely, if we were to directly use the mask for blending the original and inpainted panoramas, shadows would be re-introduced in the final image. Therefore, we use the pixels from the inpainted version not only where the mask indicates, but also in nearby regions where the inpainted image is significantly different from the original. Similarly, if those significant changes are far away from the inpainting mask, they are more likely to be spurious hallucinations, so we use pixels from the original image. Figure~\ref{ablate_2} visualizes the benefits of this strategy.

\section{Results}
\label{sec:results}

\begin{table}[t]
  \centering
  \begin{tabular}{@{}lcccc@{}}
    \toprule
    Method & PSNR $\uparrow$ & SSIM $\uparrow$ & JOD $\uparrow$ & LPIPS $\downarrow$ \\
    \midrule
    LaMa~\cite{suvorov2021lama} & \textbf{26.76} & \textbf{0.916} & 7.670 & 0.094 \\
    LGPN-Net~\cite{gao2022lgpn} & 25.57 & 0.867 & 7.439 & 0.206  \\
    SD-2-inpaint & 24.67 & 0.878 & 7.487 & 0.101 \\
    Ours-inpaint & 26.02 & 0.874 & \textbf{7.691} & \textbf{0.091} \\
    \hdashline
    Ours-full & {\color{blue}{\textbf{27.05}}} & {\color{blue}{\textbf{0.930}}} & {\color{blue}{\textbf{7.753}}} & {\color{blue}{\textbf{0.056}}} \\
    \bottomrule
  \end{tabular}
  \caption{\textbf{Quantitative evaluation.} Our inpainting module is slightly worse than LaMa~\cite{suvorov2021lama} on absolute difference metrics, because the underlying SD outputs lower-frequency textures. Together with superresolution and blending, our full pipeline outperforms all related techniques.}
  \label{tab:quant}
\end{table}

\begin{figure*}[t]
 \centering
 \includegraphics[width=0.195\linewidth]{fig/comp/ca9c4eca2d6f41aa8bf2d6a8a5407b15_eq}~
 \includegraphics[width=0.195\linewidth]{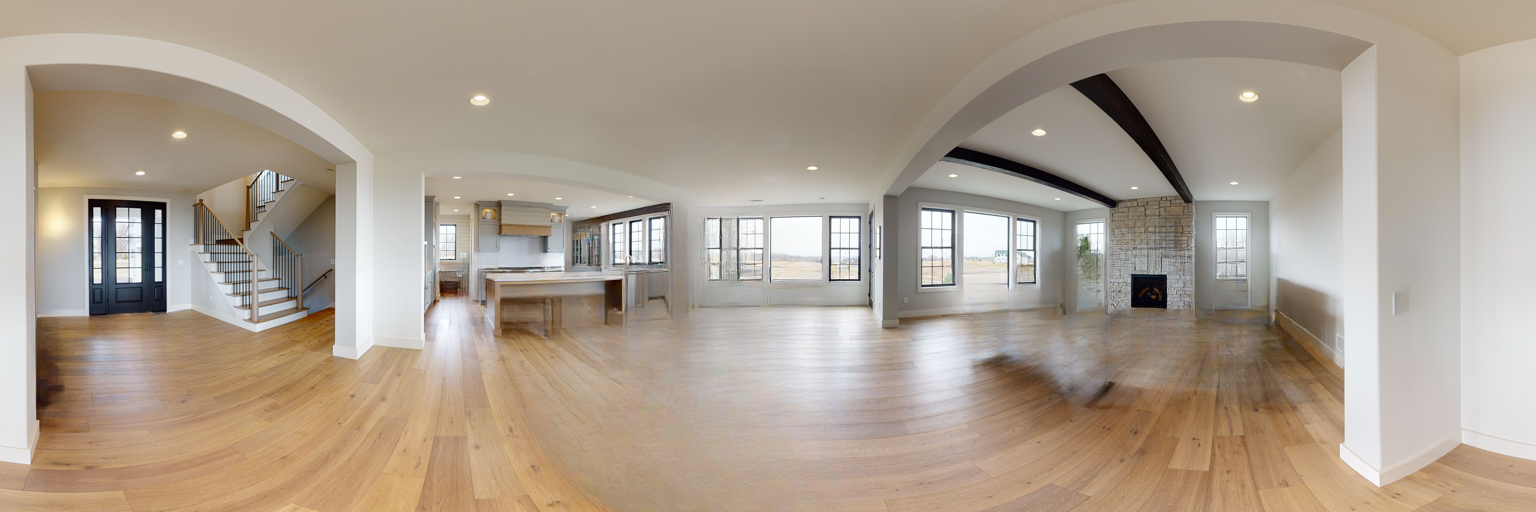}~
 \includegraphics[width=0.195\linewidth,height=0.065\linewidth]{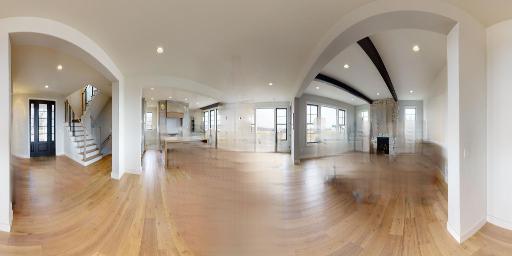}~
 \includegraphics[width=0.195\linewidth]{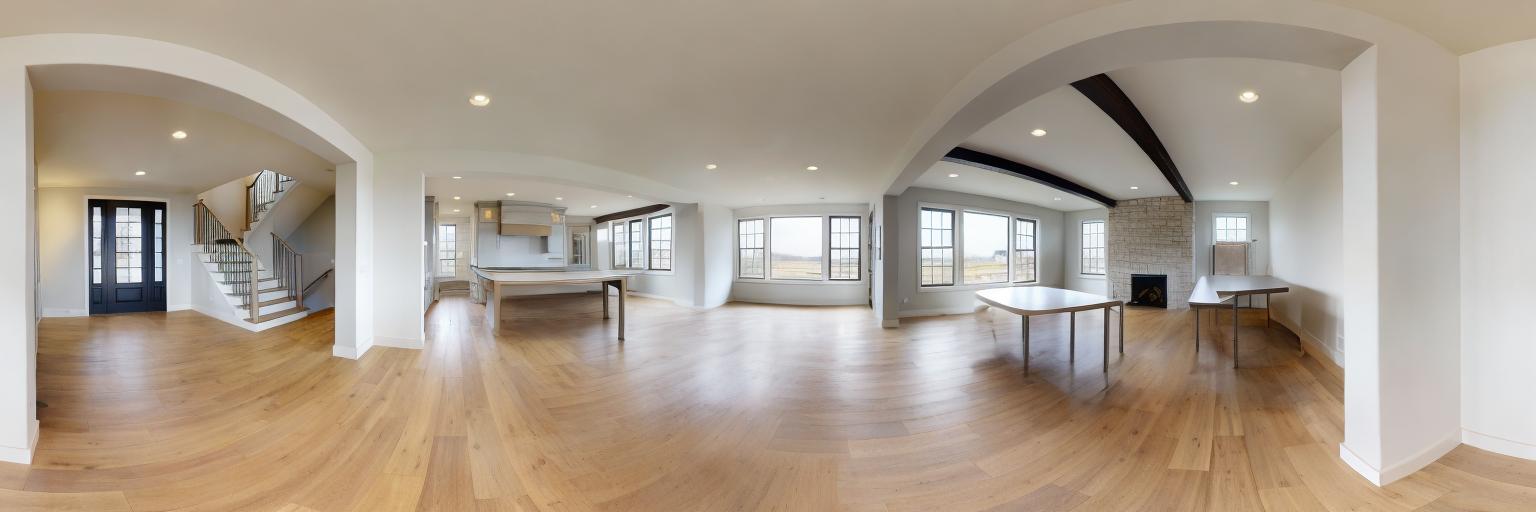}~
 \includegraphics[width=0.195\linewidth]{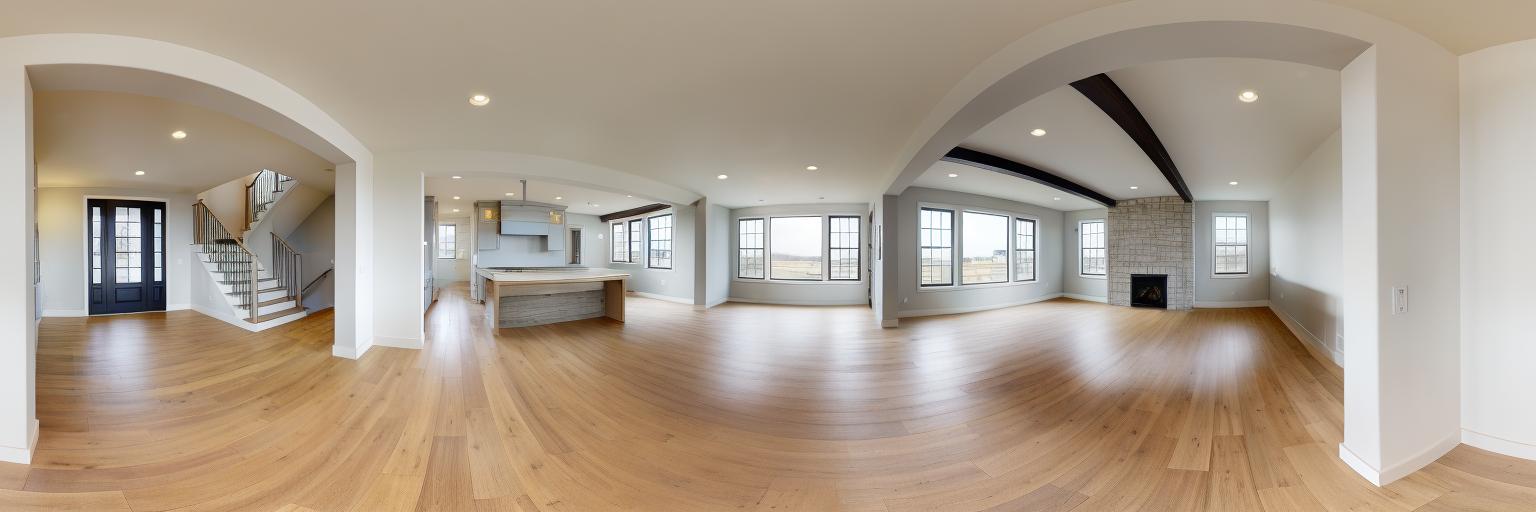}
 \\
 \includegraphics[clip,trim={900 50 200 150},width=0.195\linewidth]{fig/comp/ca9c4eca2d6f41aa8bf2d6a8a5407b15_eq}~
 \includegraphics[clip,trim={900 50 200 150},width=0.195\linewidth]{fig/comp/ca9c4eca2d6f41aa8bf2d6a8a5407b15_eq_mask_lama_dilate10_noref}~
 \includegraphics[clip,trim={300 25 66 75},width=0.195\linewidth,height=0.14\linewidth]{fig/comp/ca9c4eca2d6f41aa8bf2d6a8a5407b15_eq_inpainted_lgpn_dilate10}~
 \includegraphics[clip,trim={900 50 200 150},width=0.195\linewidth]{fig/comp/ca9c4eca2d6f41aa8bf2d6a8a5407b15_eq_vanilla_10dilate}~
 \includegraphics[clip,trim={900 50 200 150},width=0.195\linewidth]{fig/comp/ca9c4eca2d6f41aa8bf2d6a8a5407b15_eq_inpainted_comparesd}
 \\ \smallskip
 \includegraphics[width=0.195\linewidth]{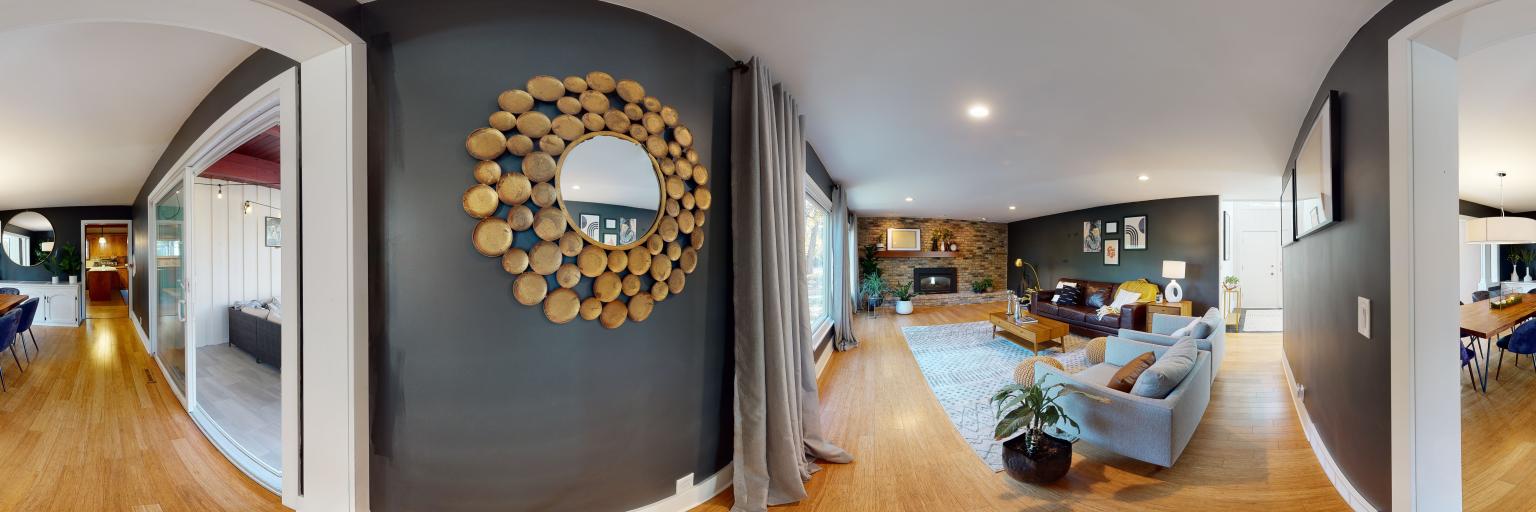}~
 \includegraphics[width=0.195\linewidth]{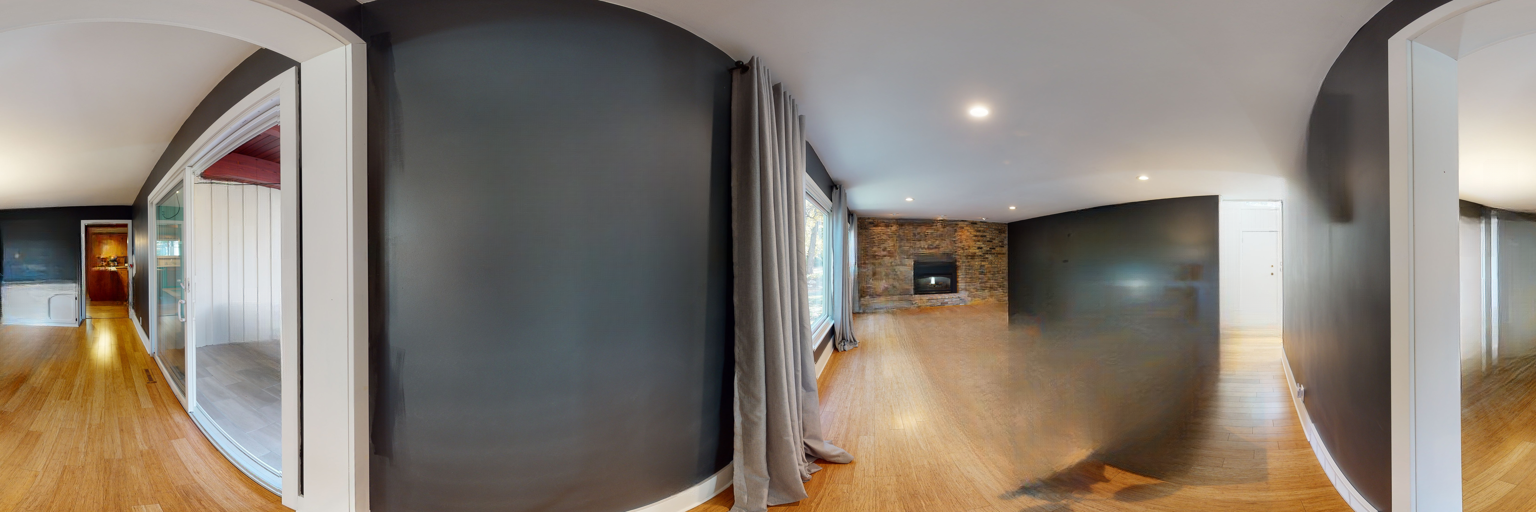}~
 \includegraphics[width=0.195\linewidth,height=0.065\linewidth]{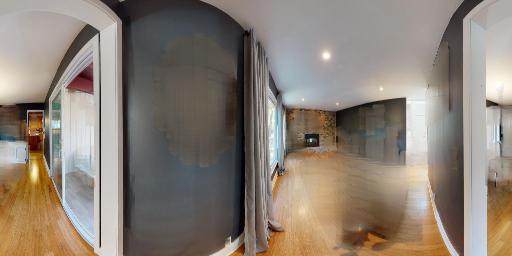}~
 \includegraphics[width=0.195\linewidth]{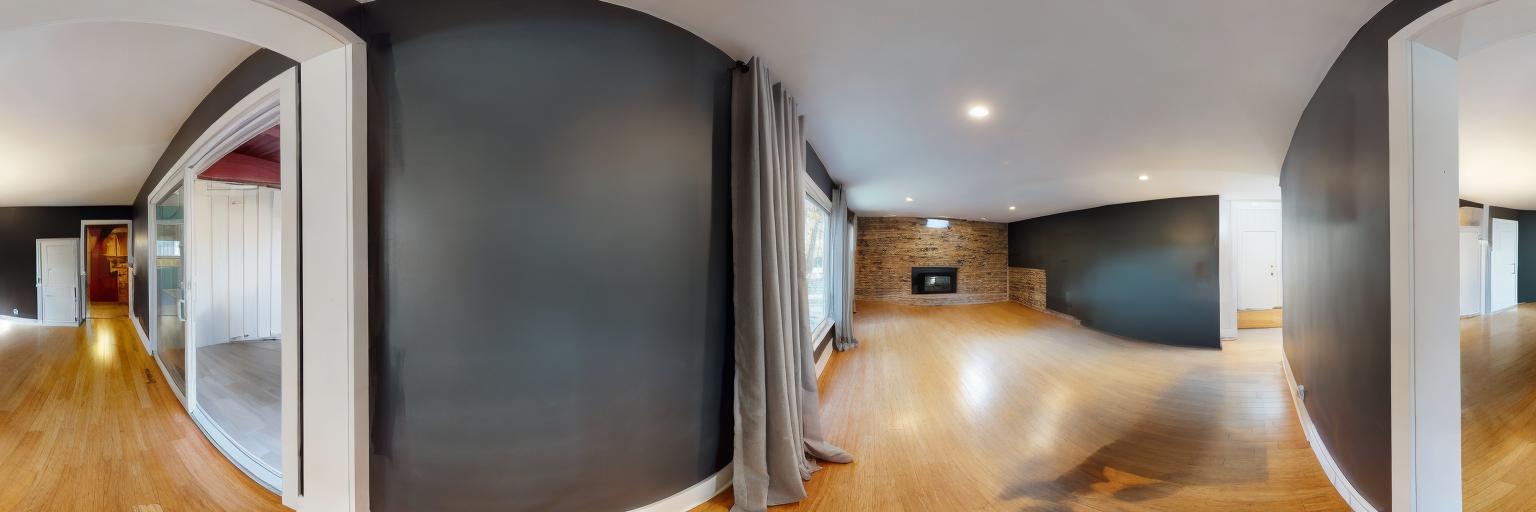}~
 \includegraphics[width=0.195\linewidth]{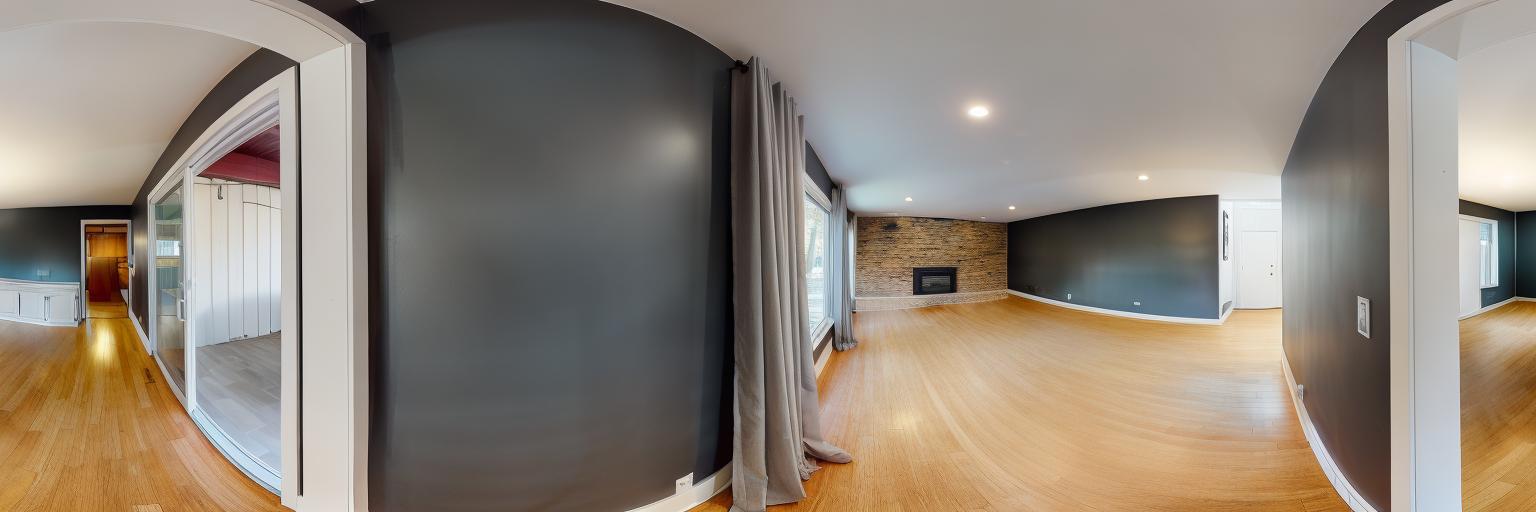}
 \\
 \includegraphics[clip,trim={800 0 300 200},width=0.195\linewidth]{fig/comp/a34d271e105744eabf60aa40f802c9d5_eq}~
 \includegraphics[clip,trim={800 0 300 200},width=0.195\linewidth]{fig/comp/a34d271e105744eabf60aa40f802c9d5_eq_mask_lama_dilate20_noref}~
 \includegraphics[clip,trim={267 0 100 100},width=0.195\linewidth,height=0.14\linewidth]{fig/comp/a34d271e105744eabf60aa40f802c9d5_eq_inpainted_lgpn_dilate10}~
 \includegraphics[clip,trim={800 0 300 200},width=0.195\linewidth]{fig/comp/a34d271e105744eabf60aa40f802c9d5_eq_vanilla_20dilate}~
 \includegraphics[clip,trim={800 0 300 200},width=0.195\linewidth]{fig/comp/a34d271e105744eabf60aa40f802c9d5_eq_inpainted_comparesd}
 \\ \smallskip
 \includegraphics[width=0.195\linewidth]{fig/comp/9985900c24be46bfa6af793fb34e393d_eq}~
 \includegraphics[width=0.195\linewidth]{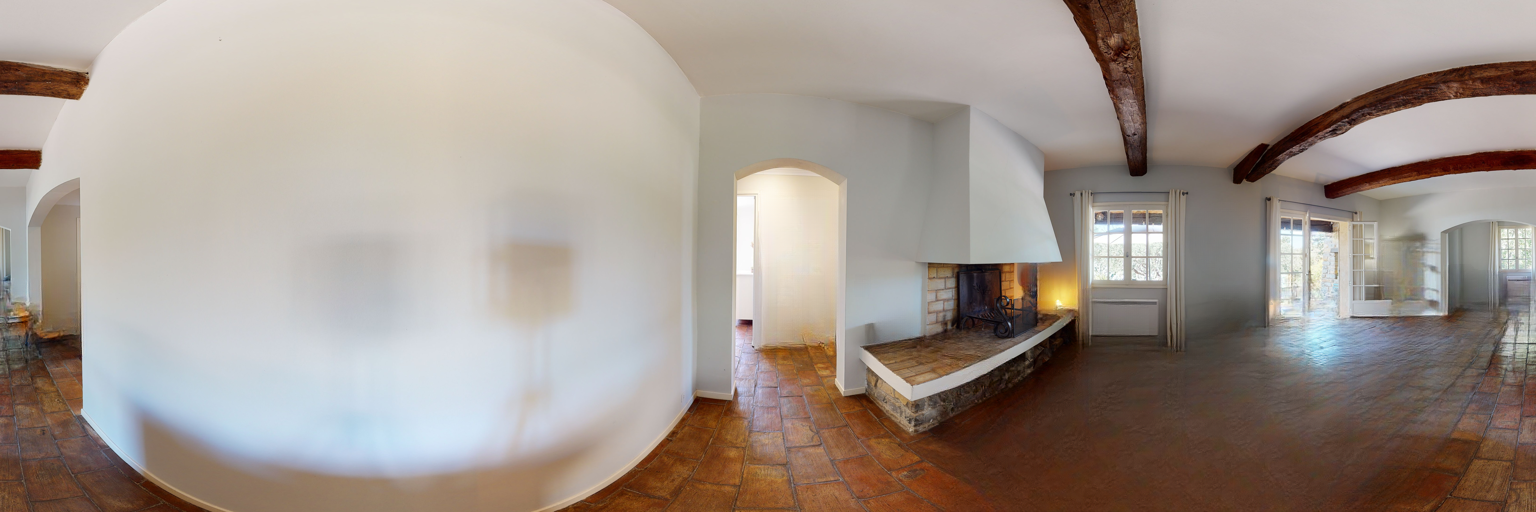}~
 \includegraphics[width=0.195\linewidth,height=0.065\linewidth]{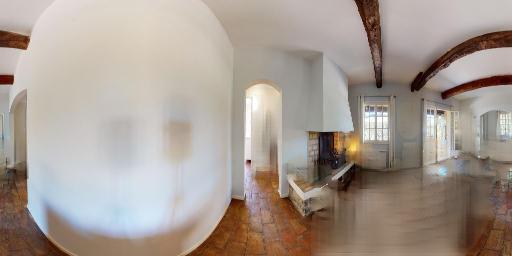}~
 \includegraphics[width=0.195\linewidth]{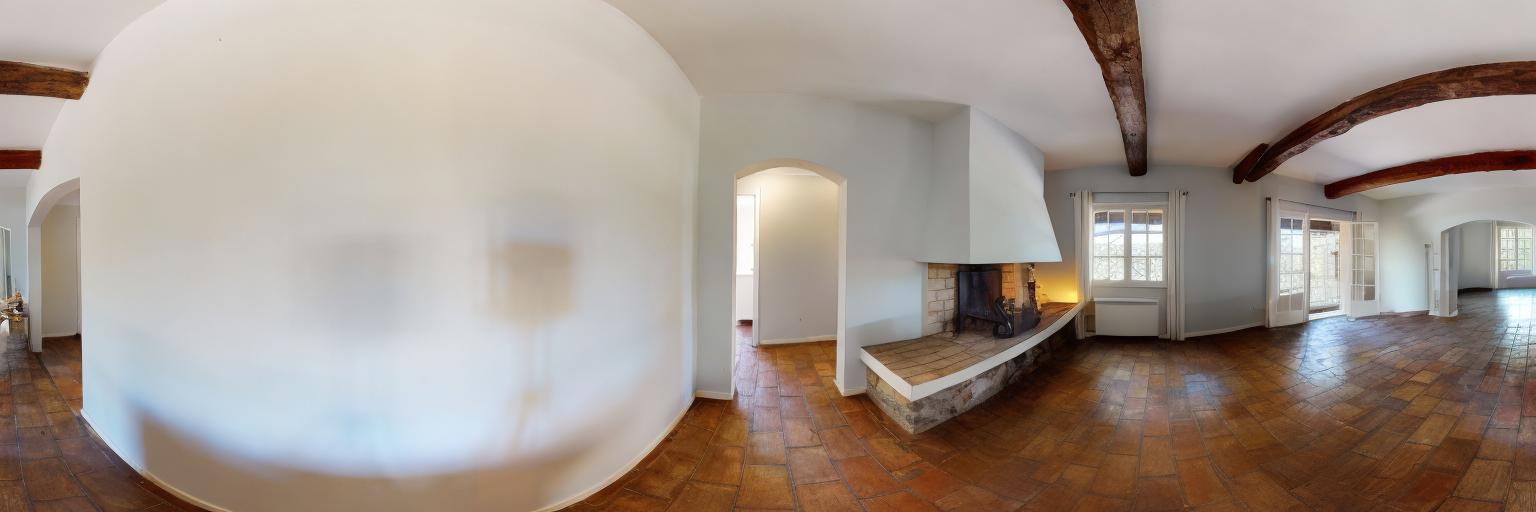}~
 \includegraphics[width=0.195\linewidth]{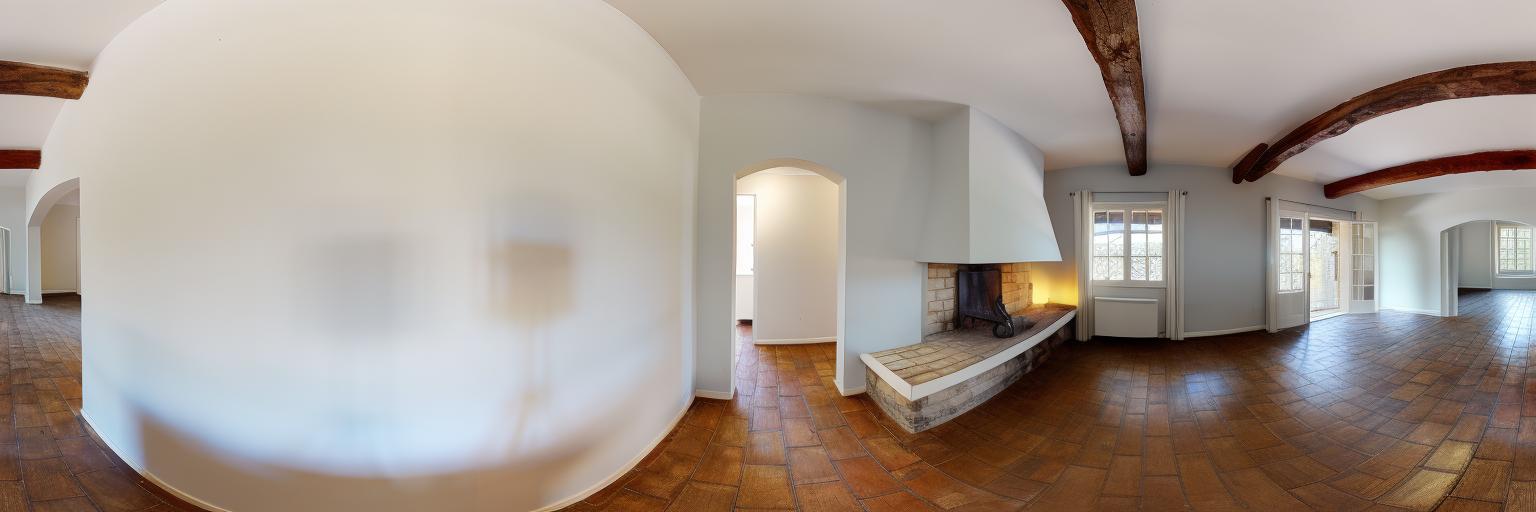}
 \\
 \includegraphics[clip,trim={900 0 180 150},width=0.195\linewidth]{fig/comp/9985900c24be46bfa6af793fb34e393d_eq}~
 \includegraphics[clip,trim={900 0 180 150},width=0.195\linewidth]{fig/comp/9985900c24be46bfa6af793fb34e393d_eq_mask_lama_dilate10_noref}~
 \includegraphics[clip,trim={300 0 60 75},width=0.195\linewidth,height=0.155\linewidth]{fig/comp/9985900c24be46bfa6af793fb34e393d_eq_inpainted_lgpn_dilate10}~
 \includegraphics[clip,trim={900 0 180 150},width=0.195\linewidth]{fig/comp/9985900c24be46bfa6af793fb34e393d_eq_vanilla_10dilate}~
 \includegraphics[clip,trim={900 0 180 150},width=0.195\linewidth]{fig/comp/9985900c24be46bfa6af793fb34e393d_eq_inpainted_comparesd}
 \\ \smallskip
 \includegraphics[width=0.195\linewidth]{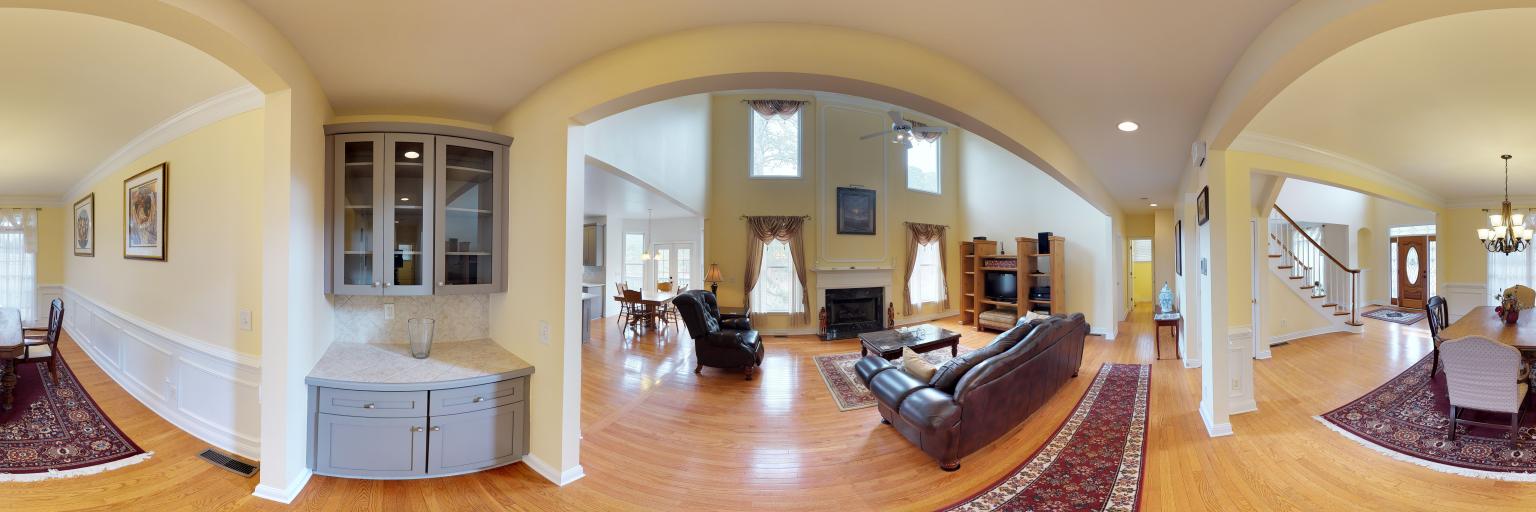}~
 \includegraphics[width=0.195\linewidth]{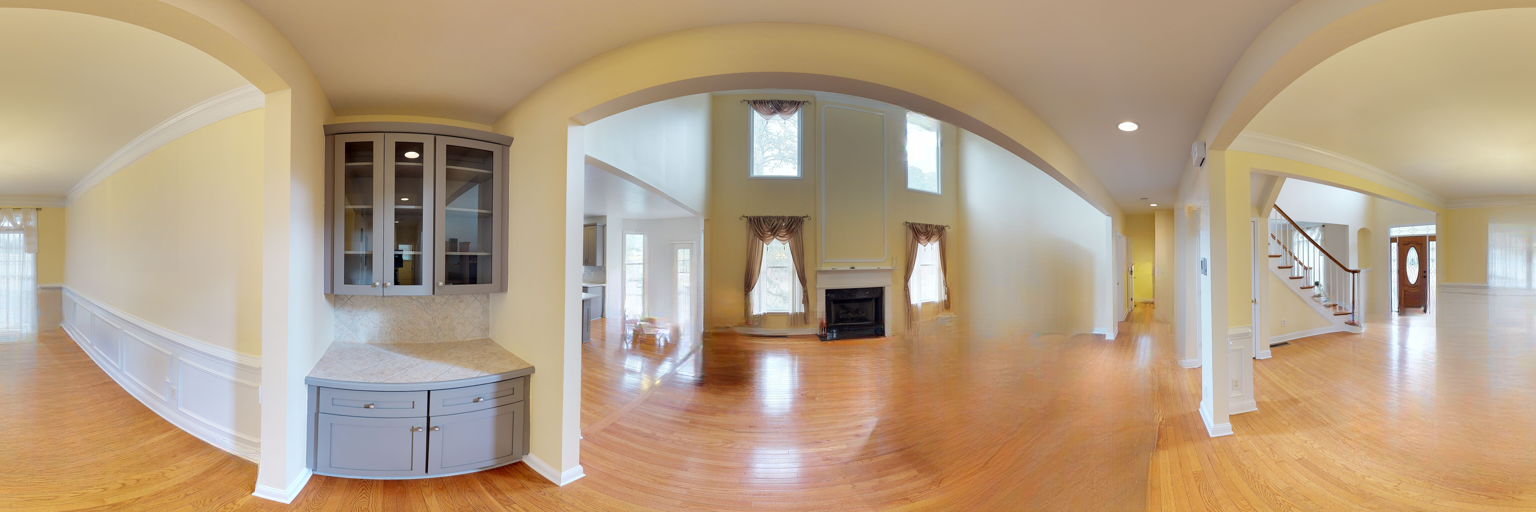}~
 \includegraphics[width=0.195\linewidth,height=0.065\linewidth]{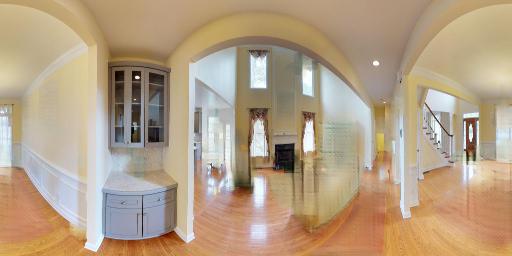}~
 \includegraphics[width=0.195\linewidth]{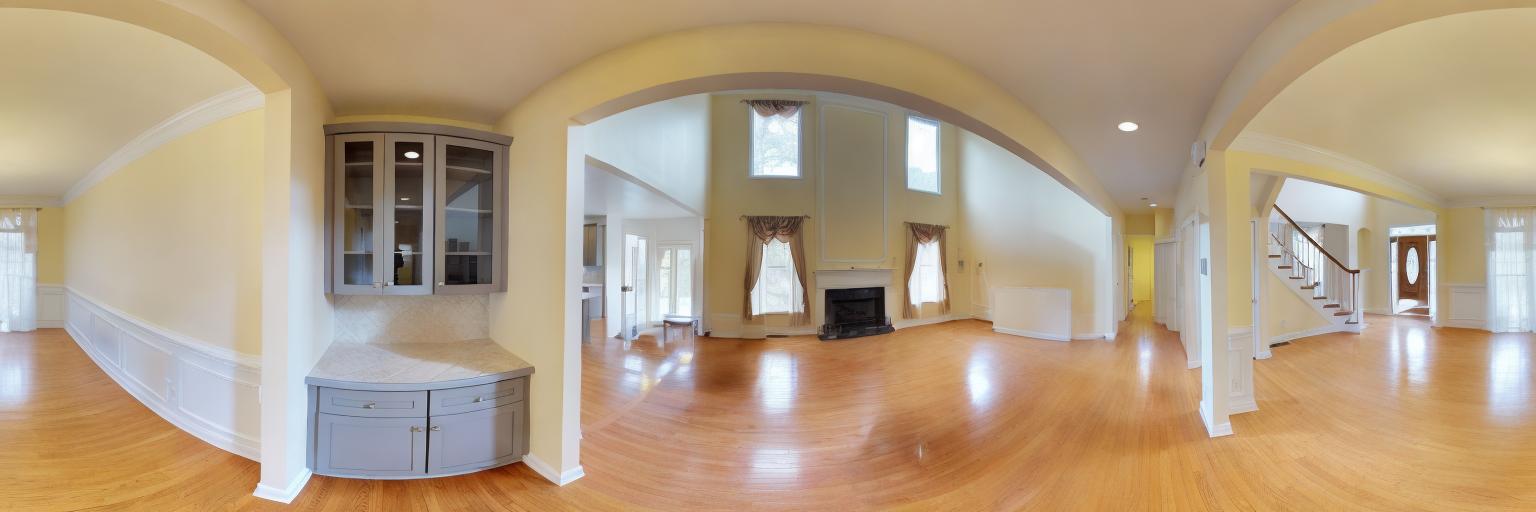}~
 \includegraphics[width=0.195\linewidth]{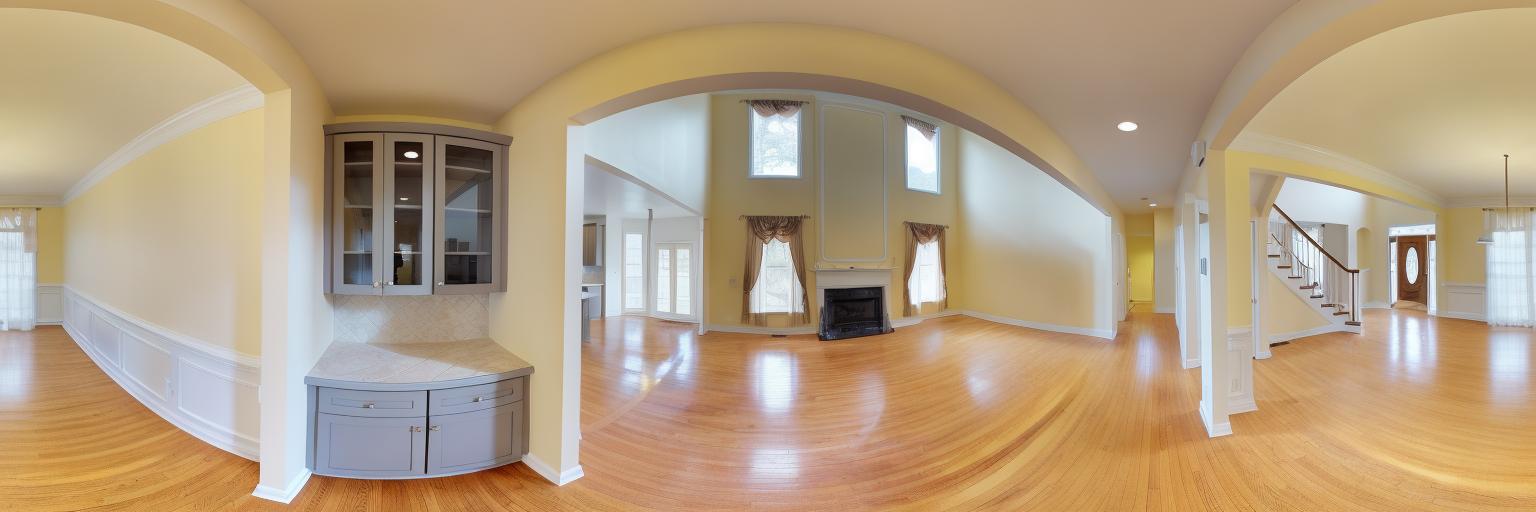}
 \\
 \includegraphics[clip,trim={600 0 400 0},width=0.195\linewidth]
    {fig/comp/eb59a0c1b01a4f7085fb3aec815cdde7_eq}
 \includegraphics[clip,trim={600 0 400 0},width=0.195\linewidth]     
    {fig/comp/eb59a0c1b01a4f7085fb3aec815cdde7_eq_mask_lama_dilate20_noref}
 \includegraphics[clip,trim={200 0 133 0},width=0.195\linewidth,height=0.186\linewidth]
    {fig/comp/eb59a0c1b01a4f7085fb3aec815cdde7_eq_inpainted_lgpn_dilate10}
 \includegraphics[clip,trim={600 0 400 0},width=0.195\linewidth]
    {fig/comp/eb59a0c1b01a4f7085fb3aec815cdde7_eq_vanilla_20dilate}
 \includegraphics[clip,trim={600 0 400 0},width=0.195\linewidth]
    {fig/comp/eb59a0c1b01a4f7085fb3aec815cdde7_eq_inpainted_comparesd}
 $\quad$ {\small{(a) Original}} $\qquad\qquad\quad$ {\small{(b) LaMa~\cite{suvorov2021lama}}} $\qquad\qquad$ {\small{(c) LGPN-Net~\cite{gao2022lgpn}}} $\qquad\quad\!$ {\small{(d) SD-2-inpaint}} $\qquad\qquad$ {\small{(e) Ours-inpaint}}
 \caption{
 \textbf{Defurnishing comparison.} Examples are arranged in row pairs of full images and zoomed-in patches. Results from GAN-based approaches, LaMa and LGPN-Net, are blurry and contain remnants of furniture objects, even though inpainting masks had to be dilated for these methods to run optimally. SD generates crisper images, but may seek to explain shadows by hallucinating furniture. Our custom fine-tuning removes the hallucinations, resulting in coherent textures, even though it is the only one that does not require mask dilation. 
 }
 \label{fig:comp}
\end{figure*}

In this section, we compare to related works and analyze the contributions of the various components of our pipeline.

\subsection{Furniture removal} \label{sec:comparisons}
We first analyze only the custom-trained inpainting component of our pipeline. The input consists of a 1536$\times$512 pixel panorama image and a corresponding mask, and the output is of the same resolution,~\ie no pre-processing to roll and pad or post-processing to upsample and blend is done.

We compare to the following related approaches:
\begin{itemize}
    \item LaMa~\cite{suvorov2021lama}: a ResNet-like inpainting network that relies on fast Fourier convolutions~\cite{chi2020ffc} to handle large masks;
    \item LGPN-Net~\cite{gao2022lgpn}: estimates the room layout via HorizonNet~\cite{sun2019horizonnet} and uses its edges to guide a GAN for inpainting;
    \item Stable Diffusion 2.0 inpainting~\cite{rombach2022highresolution}.
\end{itemize}
Note that masks are dilated by 10-20 pixels for these methods, as they are not specifically trained to handle mask inaccuracies like ours, and respectively their results were worse without the dilation. No dilation is applied for our method.

\paragraph{Qualitative comparison}
We test on furnished panoramas from the Habitat dataset~\cite{ramakrishnan2021hm3d}. Figure~\ref{fig:comp} compares the results on a few examples. The ResNet- and GAN-based approaches, LaMa and LGPN-Net, tend to generate very blurry results. Moreover, they leave pieces of furniture objects un-inpainted. 
This is an even more prevalent effect with smaller or no mask dilation, as shown in the supplementary document. 
On the other hand, SD and our custom modification of it generate much crisper textures, even in very complex scenes. While plain SD inpainting hallucinates objects, like the tables in the first example, our method does not. In addition, SD leaves shadows behind as in the second example. The training of our method enables it to modify pixels outside of the inpainting mask, which is why it manages to remove the shadow and inpaint it with plausible floor texture.

\begin{figure*}[t]
 \centering
 \begin{subfigure}[t]{0.24\textwidth}
    \centering
    \includegraphics[width=\linewidth]{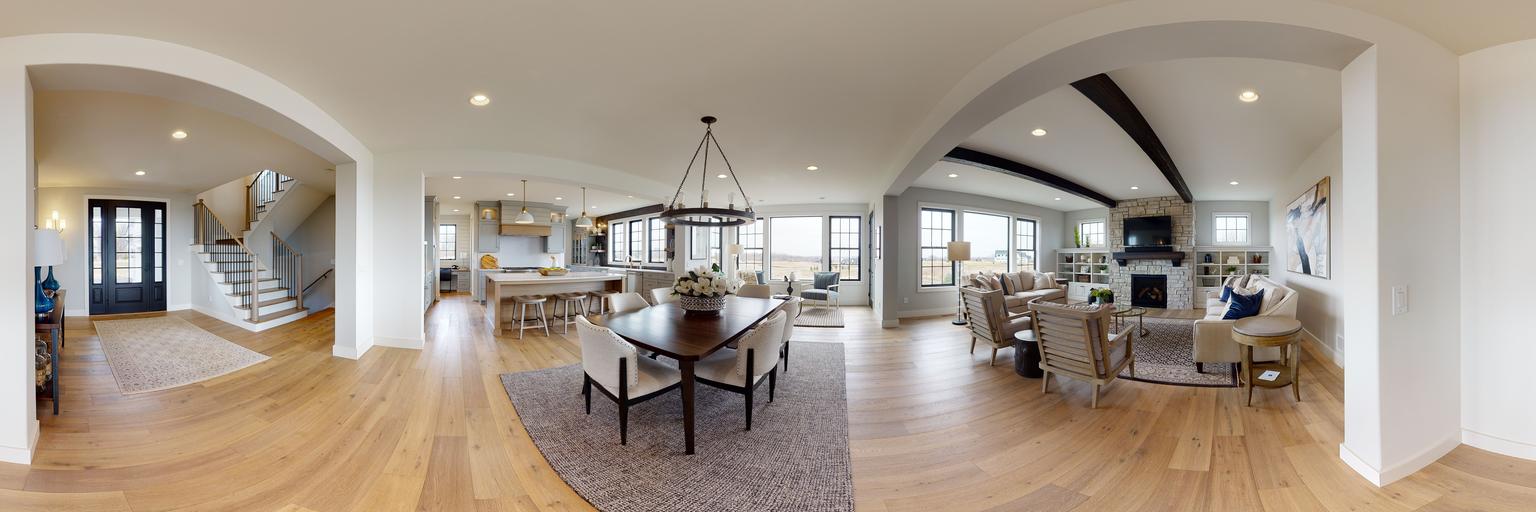}\\
    \includegraphics[clip,trim={450 50 700 100},width=\linewidth]{fig/ablate_0/cropped_ca9c4eca2d6f41aa8bf2d6a8a5407b15_eq.jpg}
    \caption{Original}
 \end{subfigure}%
 ~
 \begin{subfigure}[t]{0.24\textwidth}
    \centering
    \includegraphics[width=\linewidth]{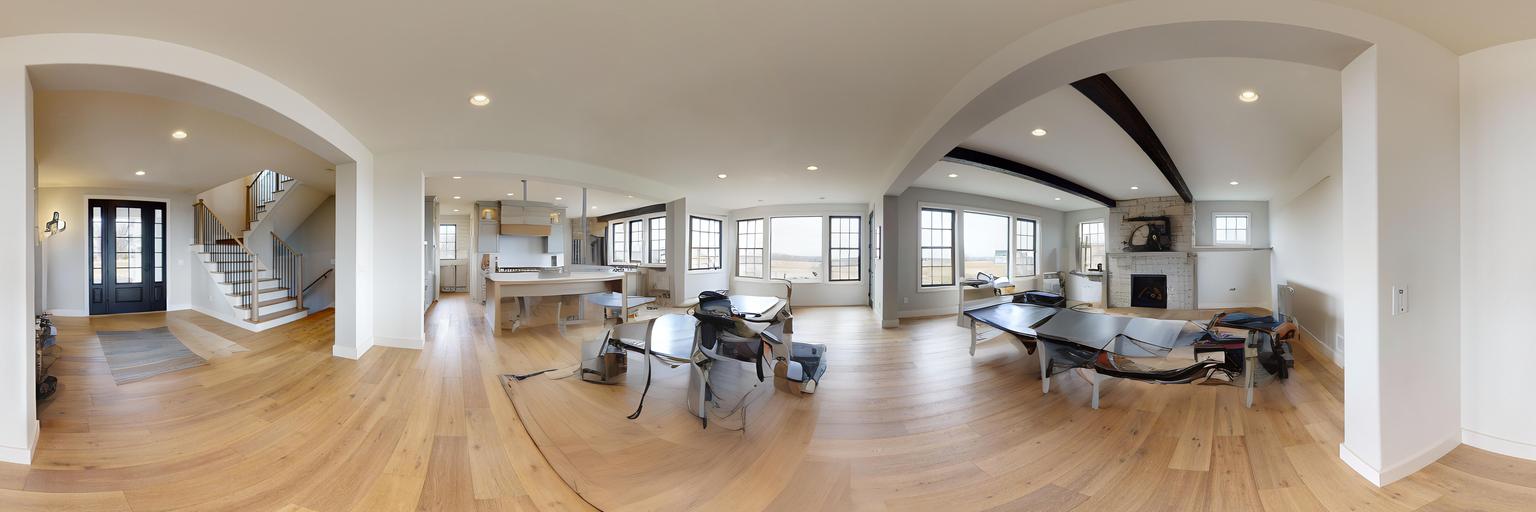}\\
    \includegraphics[clip,trim={450 50 700 100},width=\linewidth]{fig/ablate_0/cropped_ca9c4eca2d6f41aa8bf2d6a8a5407b15_eq_vanilla_nodilate.jpg}
    \caption{SD-2-inpaint, no mask dilation}
 \end{subfigure}%
 ~
 \begin{subfigure}[t]{0.24\textwidth}
    \centering
    \includegraphics[width=\linewidth]{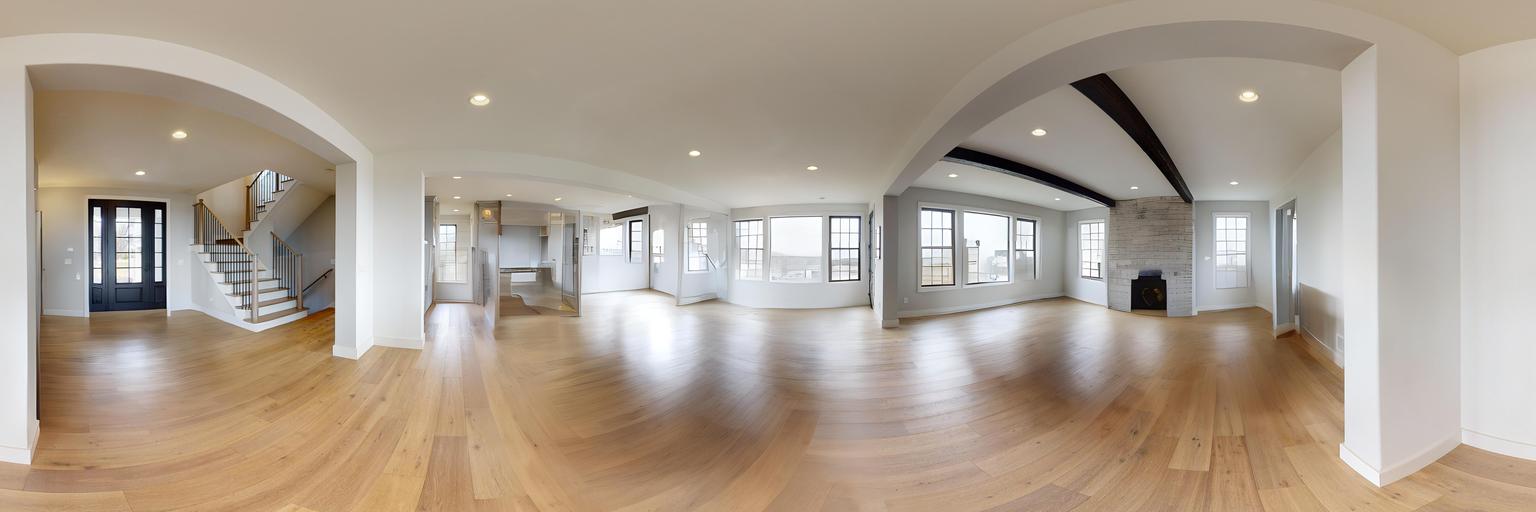}\\
    \includegraphics[clip,trim={450 50 700 100},width=\linewidth]{fig/ablate_0/cropped_ca9c4eca2d6f41aa8bf2d6a8a5407b15_eq_vanilla.jpg}
    \caption{SD-2-inpaint, dilated mask}
    \label{ablate_0:c}
 \end{subfigure}%
 ~
 \begin{subfigure}[t]{0.24\textwidth}
    \centering
    \includegraphics[width=\linewidth]{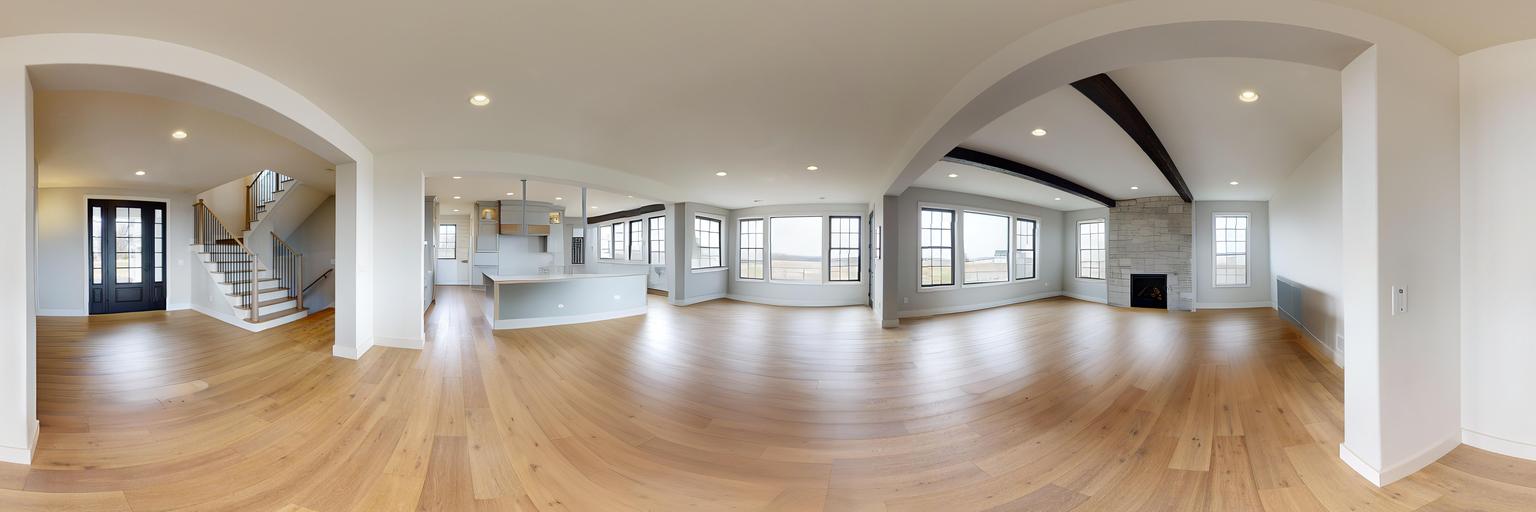}\\
    \includegraphics[clip,trim={450 50 700 100},width=\linewidth]{fig/ablate_0/cropped_ca9c4eca2d6f41aa8bf2d6a8a5407b15_eq_inpainted.jpg}
    \caption{Ours, no mask dilation}
 \end{subfigure}%
 \caption{\textbf{LoRA fine-tuning.} Comparing off-the-shelf weights for Stable Diffusion v2 inpainting and our fine-tuned weights.}
 \label{ablate_0}
\end{figure*}

\begin{figure*}[t]
 \centering
 \begin{subfigure}[t]{0.24\textwidth}
    \centering
    \includegraphics[width=\linewidth]{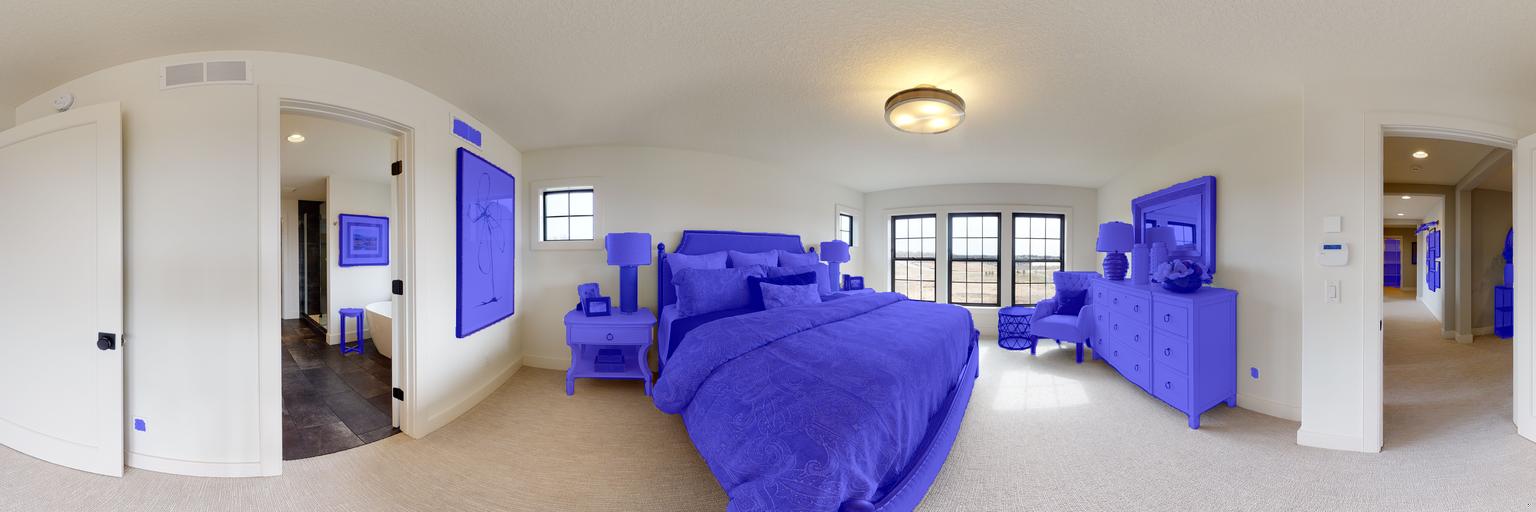}\\
    \includegraphics[width=\linewidth]{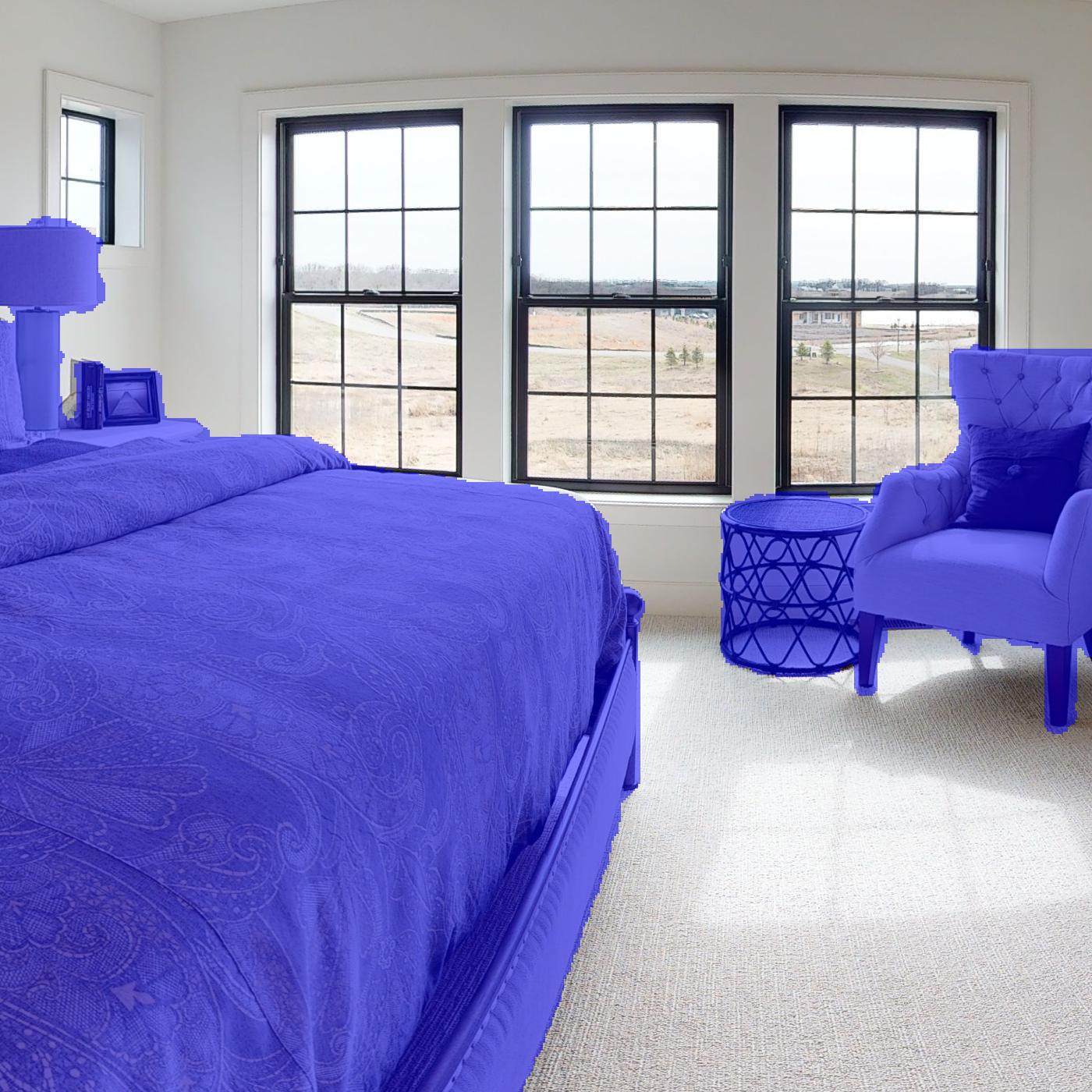}
    \caption{Original; mask overlaid in blue}
 \end{subfigure}%
 ~
 \begin{subfigure}[t]{0.24\textwidth}
    \centering
    \includegraphics[width=\linewidth]{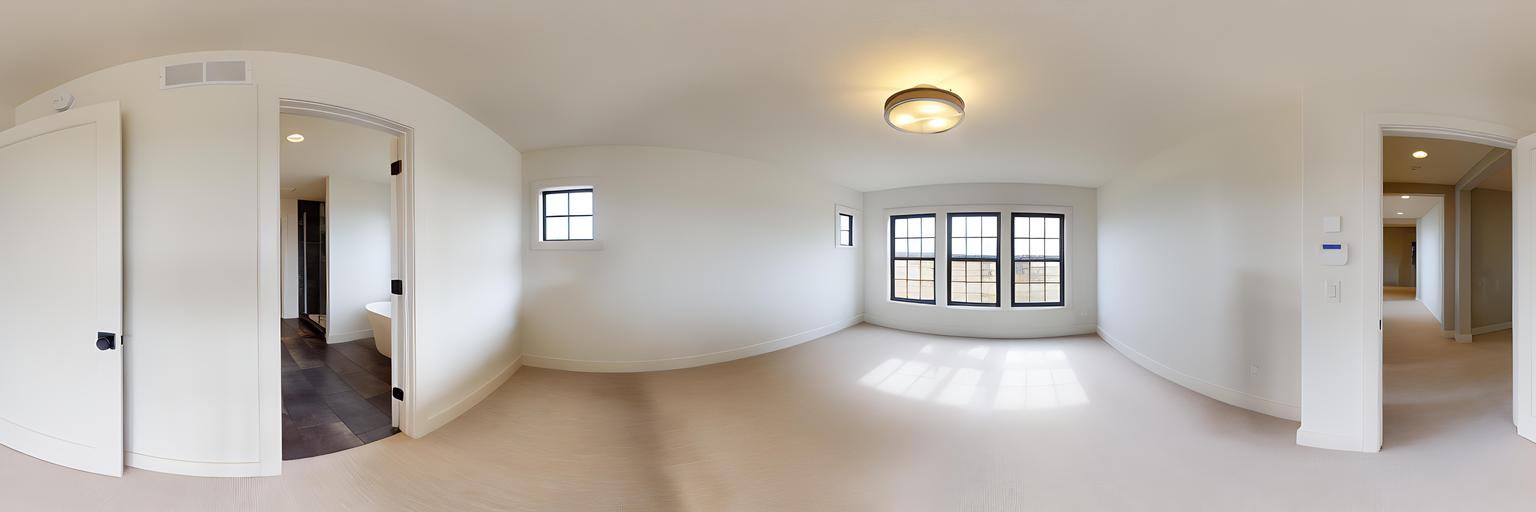}\\
    \includegraphics[width=\linewidth]{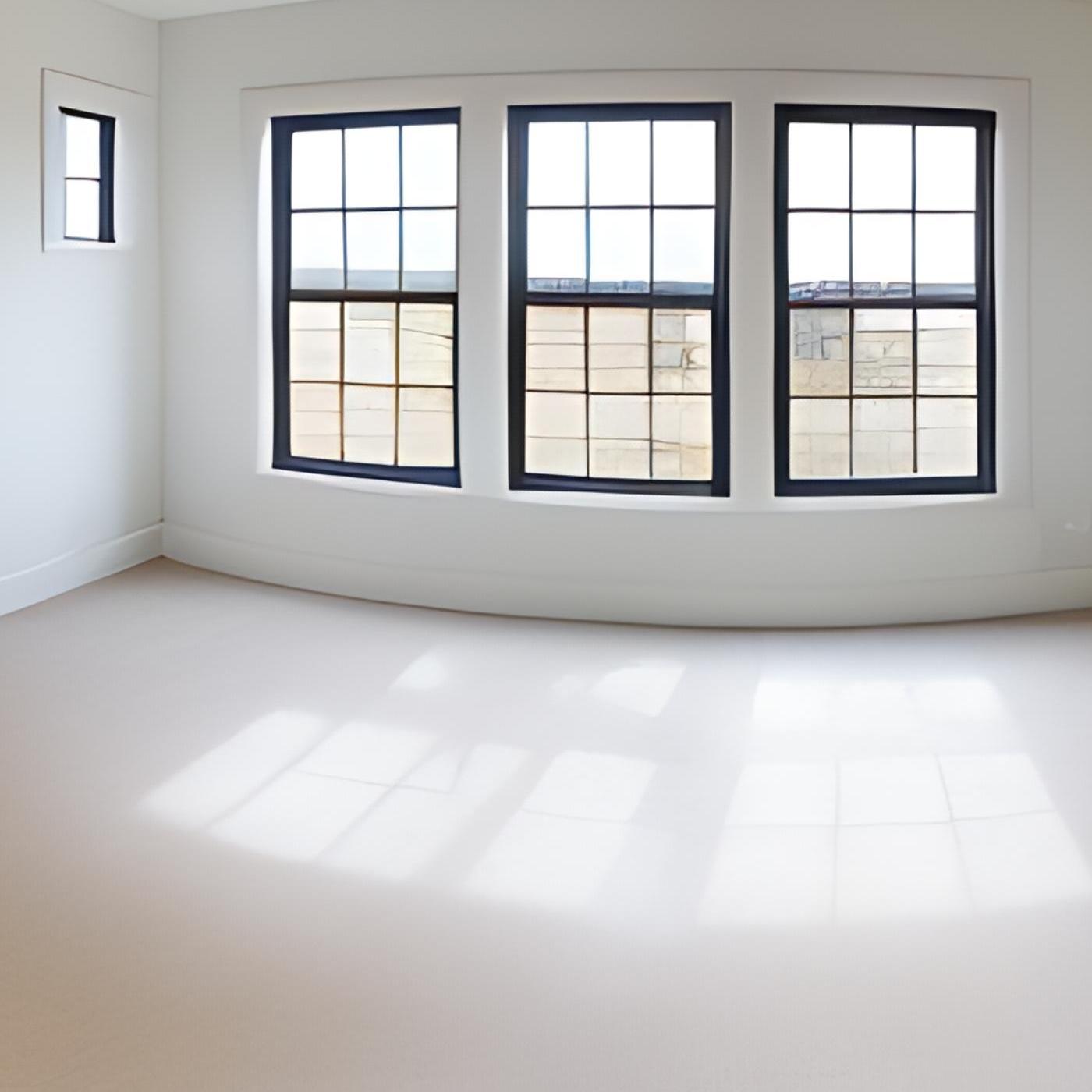}
    \caption{Generated image}
    \label{ablate_2:b}
 \end{subfigure}%
 ~
 \begin{subfigure}[t]{0.24\textwidth}
    \centering
    \includegraphics[width=\linewidth]{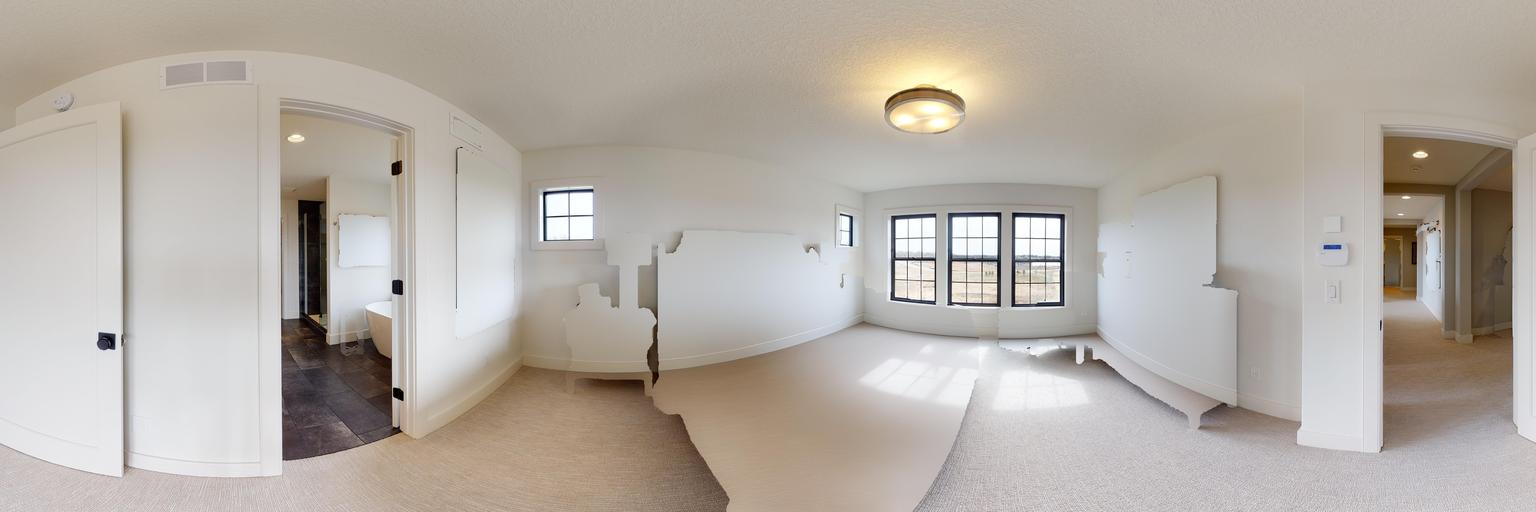}\\
    \includegraphics[width=\linewidth]{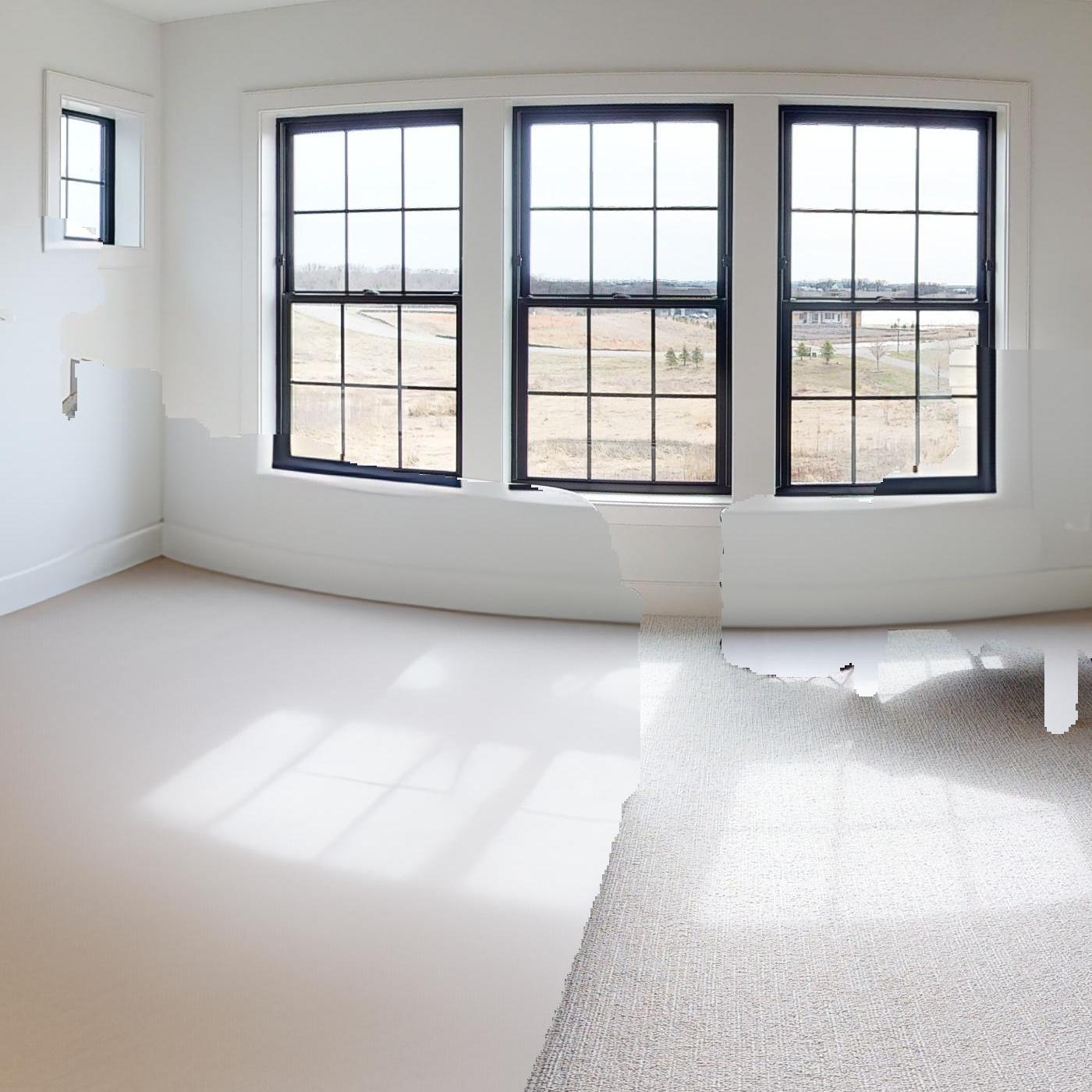}
    \caption{Na\"ive replacement}
 \end{subfigure}%
 ~
 \begin{subfigure}[t]{0.24\textwidth}
    \centering
    \includegraphics[width=\linewidth]{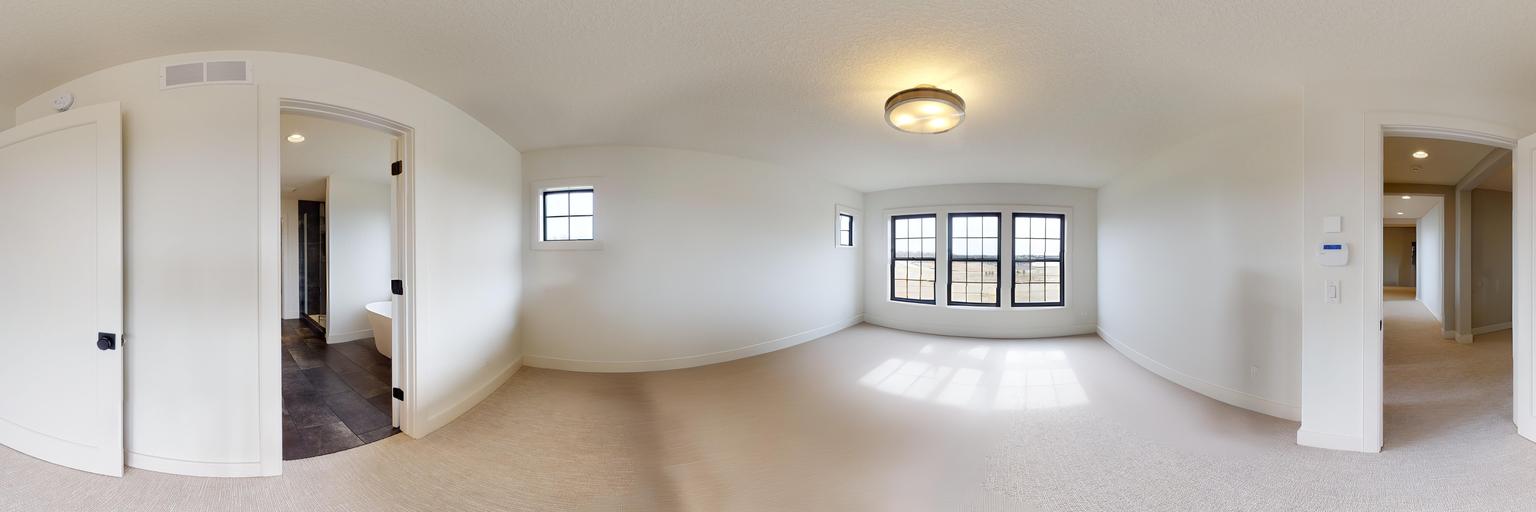}
    \includegraphics[width=\linewidth]{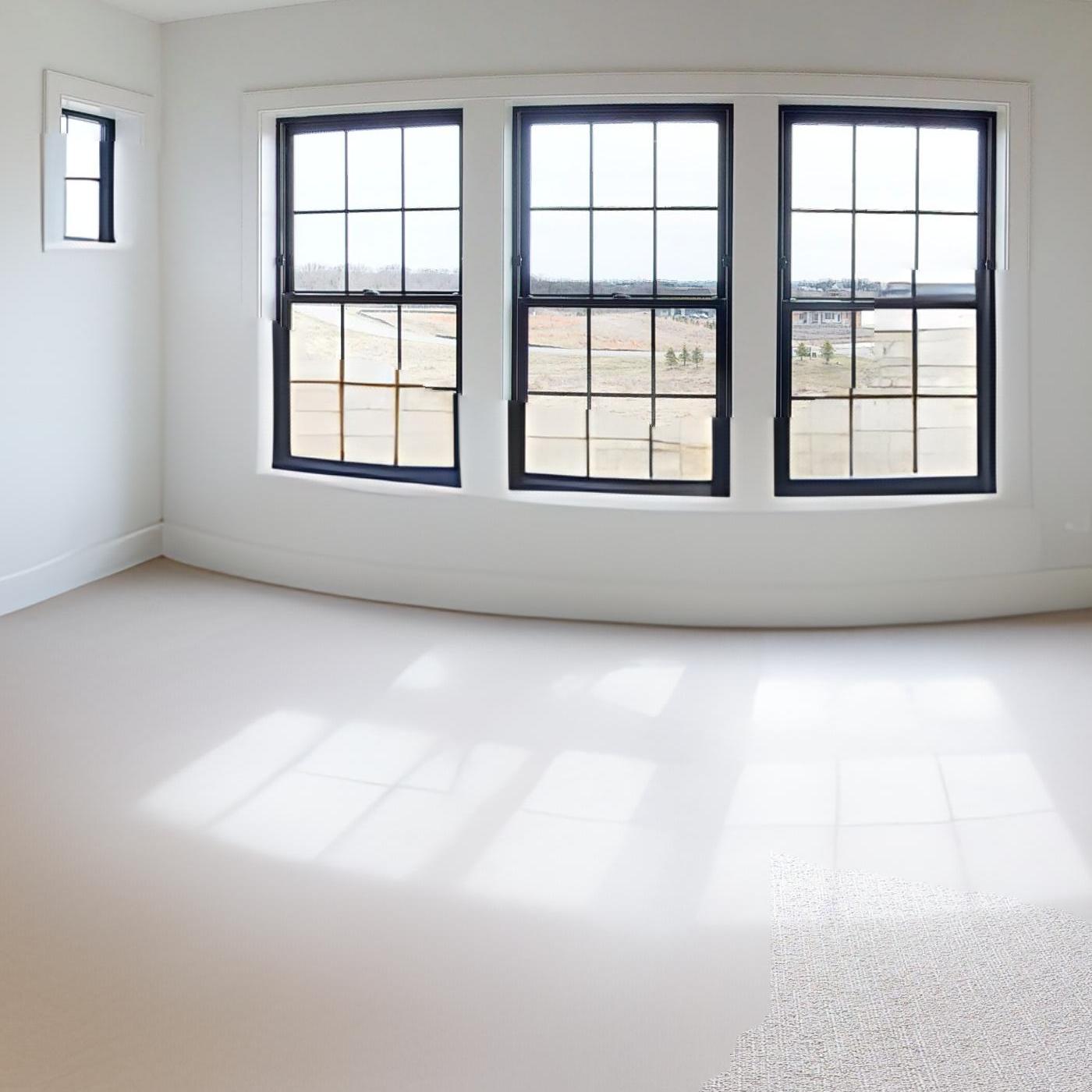}
    \caption{Our blending}
 \end{subfigure}%
 \caption{\textbf{Blending.} Our blending technique combines the original and generated images according to the inpainting mask, shown in blue.}
 \label{ablate_2}
\end{figure*}

\paragraph{Quantitative comparison}
To quantitatively compare results, we gather a selection of unfurnished panoramas from the Habitat dataset~\cite{ramakrishnan2021hm3d}, insert synthetic furniture such as tables and ottomans on the floors, and measure the difference between the defurnished images produced by each method and the original unfurnished images. 
Note that the spaces produced this way have simpler furniture setups than in the natural images we used in the previous section. 
We evaluate absolute differences via PSNR, and perceptual differences via SSIM~\cite{wang2004ssim}, LPIPS~\cite{zhang2018lpips}, and FovVideoVDP~\cite{mantiuk2021fovvideovdp}, a kind of just-objectionable-difference (JOD) measure for wide field-of-view images or video. Here, in addition to our inpainting network (Ours-inpaint), we evaluate our full pipeline (Ours-full), for a fairer comparison to a complete defurnishing method like LGPN-Net. The results are summarized in Table~\ref{tab:quant}. 
As the synthetic furniture does not include rugs and leaves large areas from the original flooring visible (shown in supplementary document), LaMa and LGPN-Net manage to propagate these textures well into the inpainted regions, and thus achieve good PSNR and SSIM scores. However, the borders between inpainted and uninpainted regions are blurry and disjoint, leaving these methods with low perceptual scores. On the other hand, our method and plain SD have somewhat lower-resolution textures, leading to worse PSNR. Conversely, they generate continuous textures with higher LPIPS and JOD. The post-processing component of our pipeline is developed exactly to remedy the low-resolution imagery---indeed, its better scores across all metrics demonstrate that we have accomplished this goal.

\subsection{Ablation studies}\label{sec:ablation}

Figure~\ref{ablate_0} shows an example inpainting result using off-the-shelf SD weights, with and without mask dilation.
Note that without any mask dilation, the network hallucinates objects due to imperfect semantic segmentation and context from shadows. If the mask is generously dilated, the network does not have enough context to generate realistic inpaints for the kitchen island. Therefore, it is necessary to fine-tune SD on a custom dataset to generate inpaints of unfurnished spaces without requiring mask dilation.


\begin{figure*}[t]
 \centering
 \begin{subfigure}[t]{0.245\textwidth}
    \centering
    \includegraphics[clip,trim={600 100 650 150},width=\linewidth]{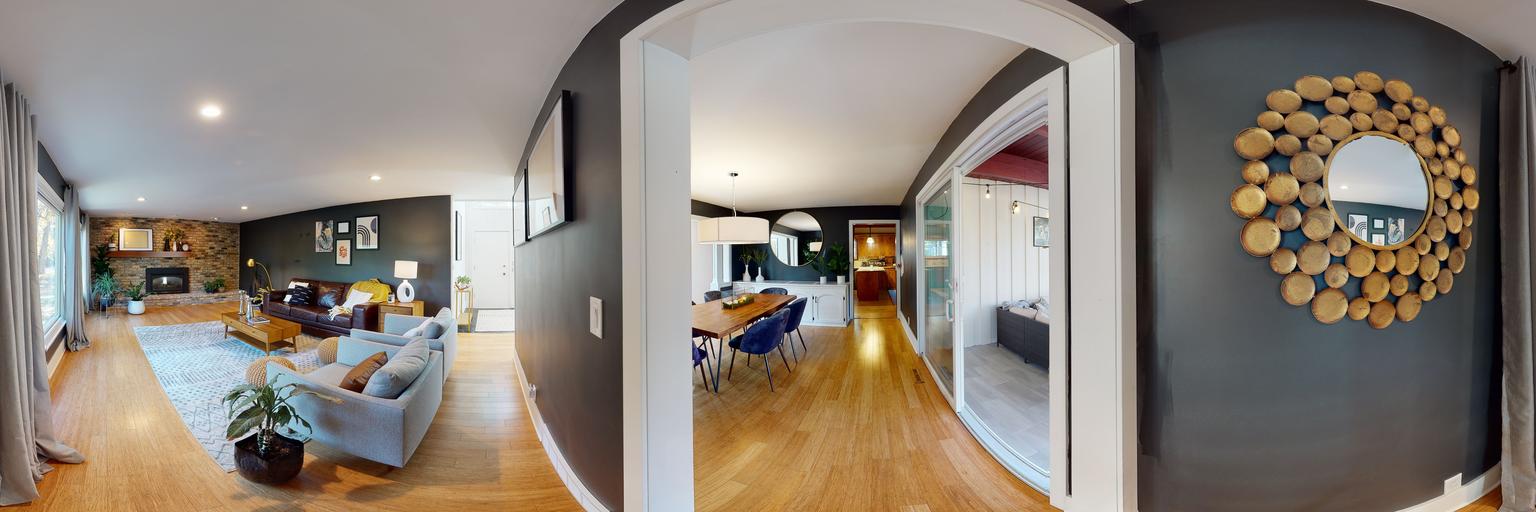}
    \caption{Original}
 \end{subfigure}%
 ~
 \begin{subfigure}[t]{0.245\textwidth}
    \centering
    \includegraphics[clip,trim={600 100 650 150},width=\linewidth]{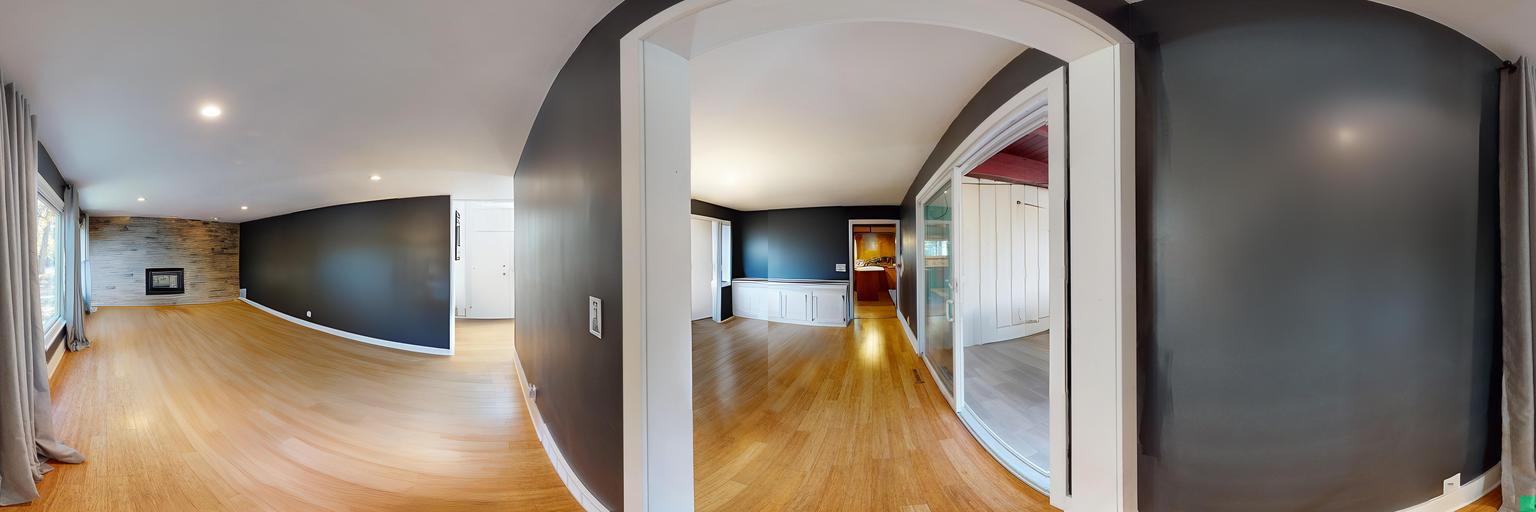}
    \caption{No rolling}
 \end{subfigure}%
 ~
 \begin{subfigure}[t]{0.245\textwidth}
    \centering
    \includegraphics[clip,trim={600 100 650 150},width=\linewidth]{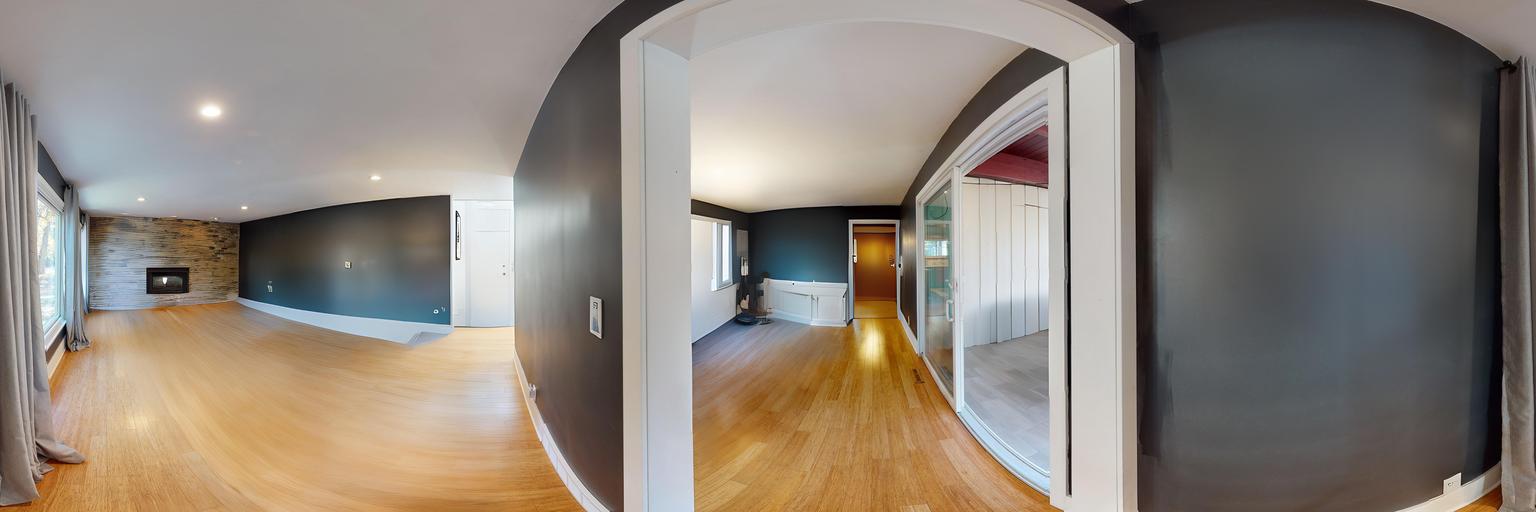}
    \caption{No padding}
 \end{subfigure}%
 ~
 \begin{subfigure}[t]{0.245\textwidth}
    \centering
    \includegraphics[clip,trim={600 100 650 150},width=\linewidth]{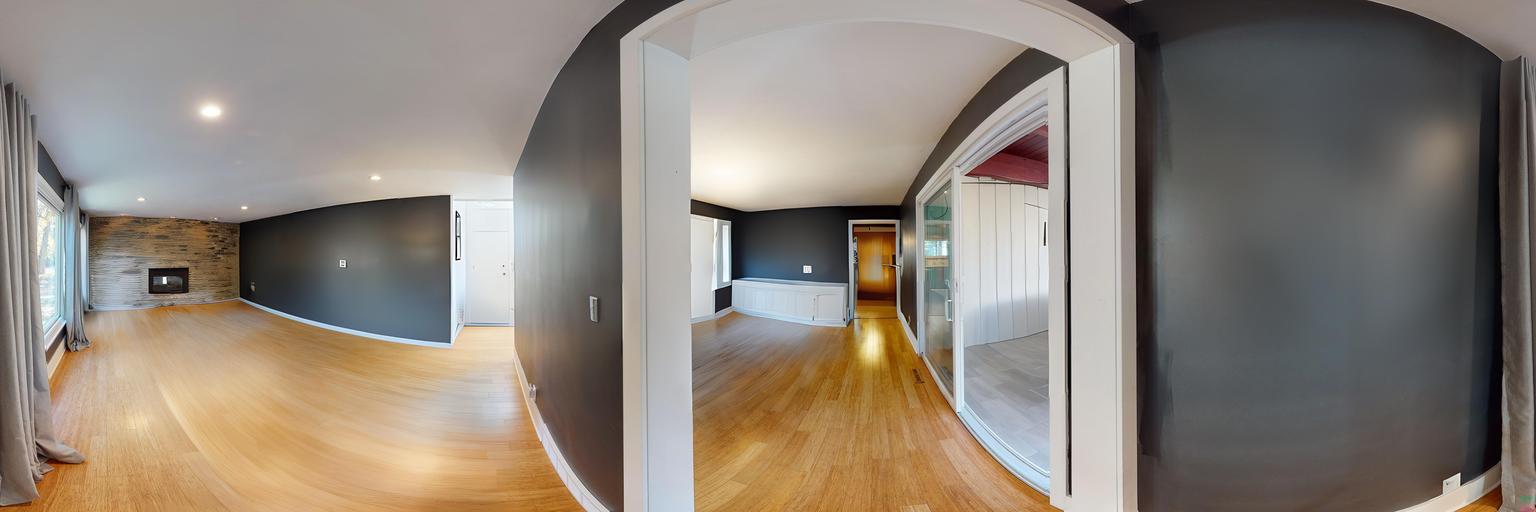}
    \caption{With rolling and padding}
 \end{subfigure} \vspace*{-1.1mm}
 \caption{\textbf{Rolling and Padding.} Inpainting results at the panorama seam with and without certain pre-processing steps.} \vspace*{-1mm}
 \label{ablate_1}
\end{figure*}


\begin{figure}[t]
 \centering
 \includegraphics[clip,trim={300 0 250 0},width=0.49\linewidth]{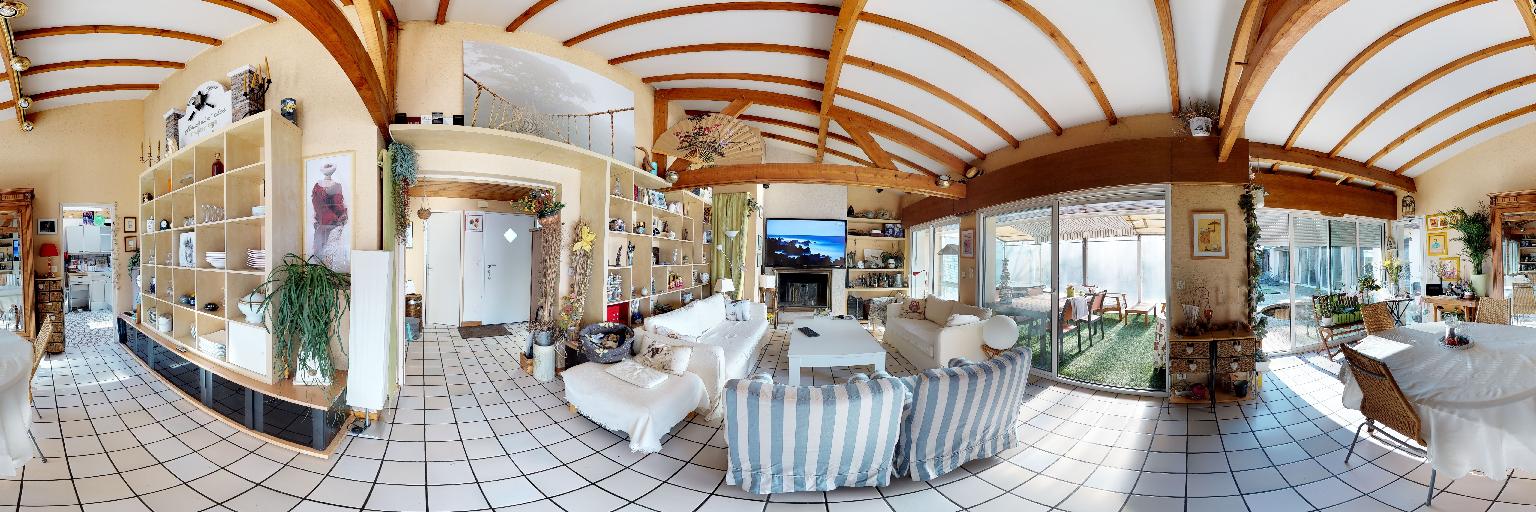}~
 \includegraphics[clip,trim={300 0 250 0},width=0.49\linewidth]{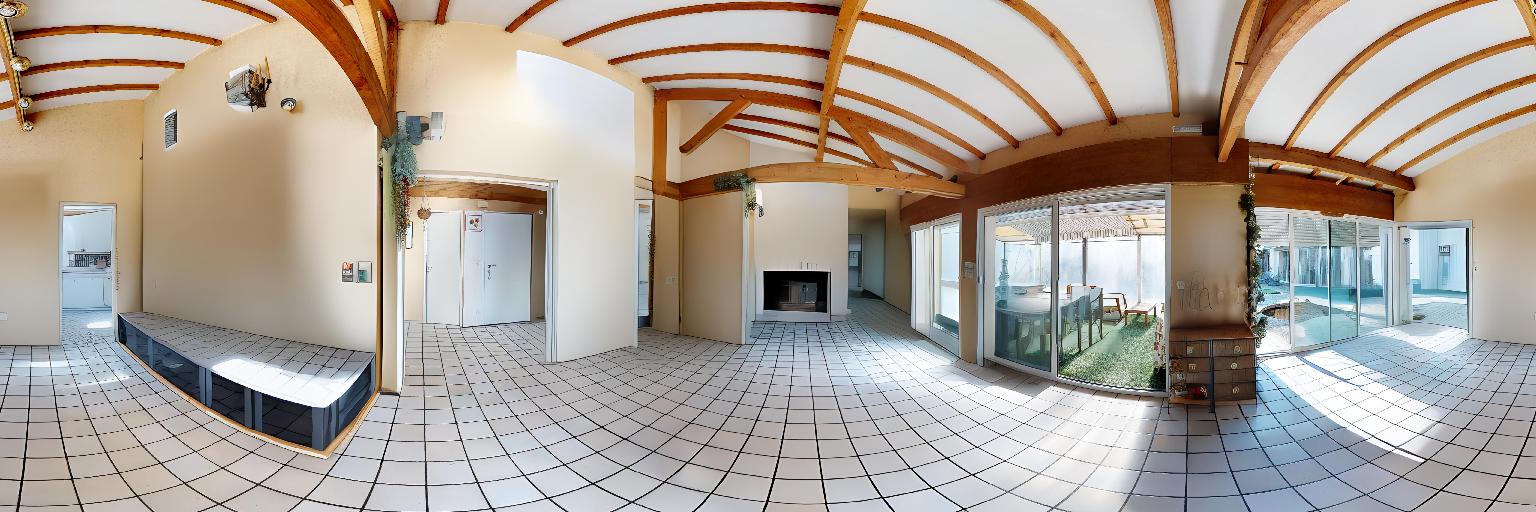}\\
 \includegraphics[clip,trim={150 0 400 0},width=0.49\linewidth]{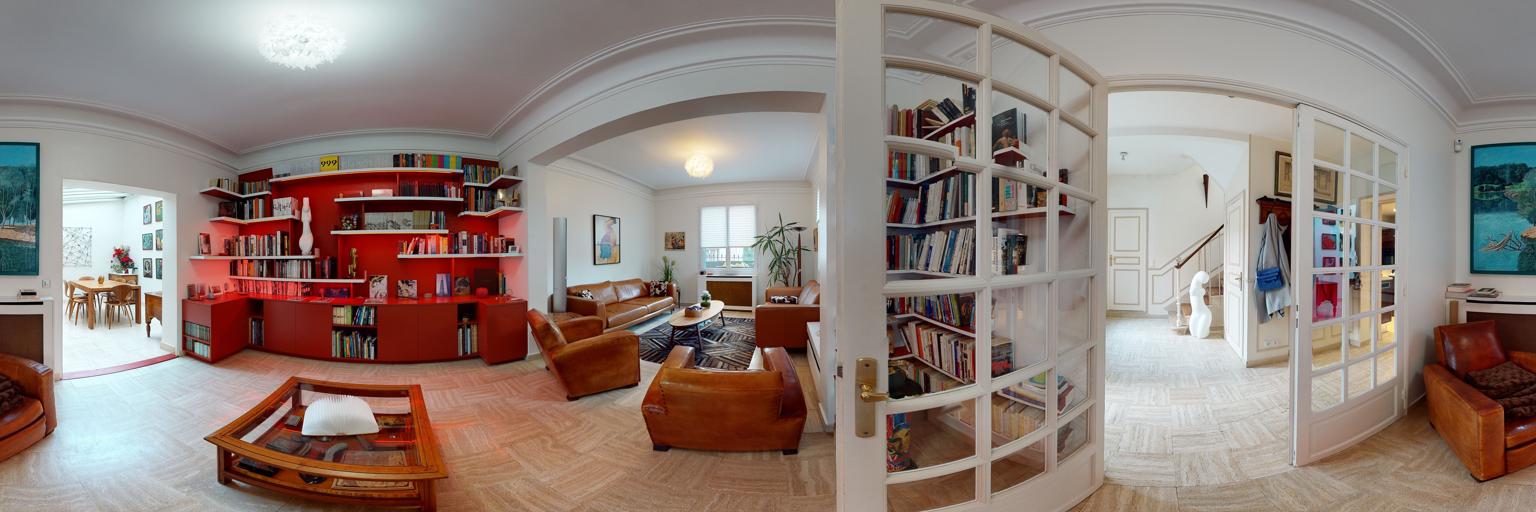}~
 \includegraphics[clip,trim={150 0 400 0},width=0.49\linewidth]{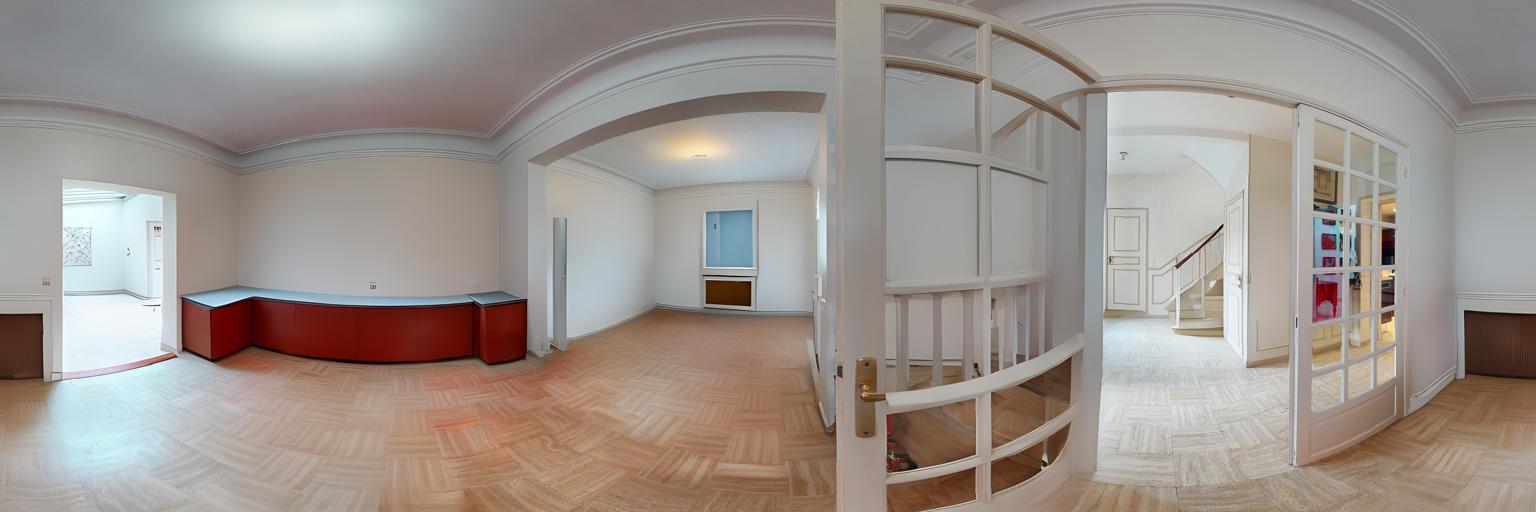}\\
 \includegraphics[clip,trim={400 0 150 0},width=0.49\linewidth]{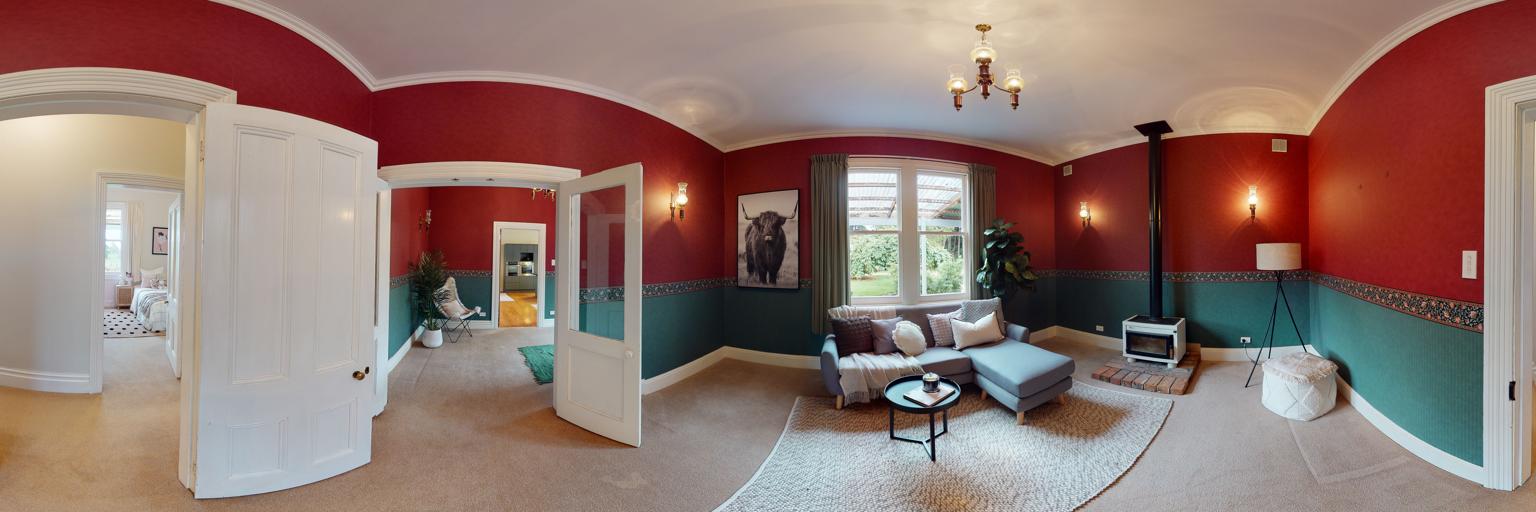}~
 \includegraphics[clip,trim={400 0 150 0},width=0.49\linewidth]{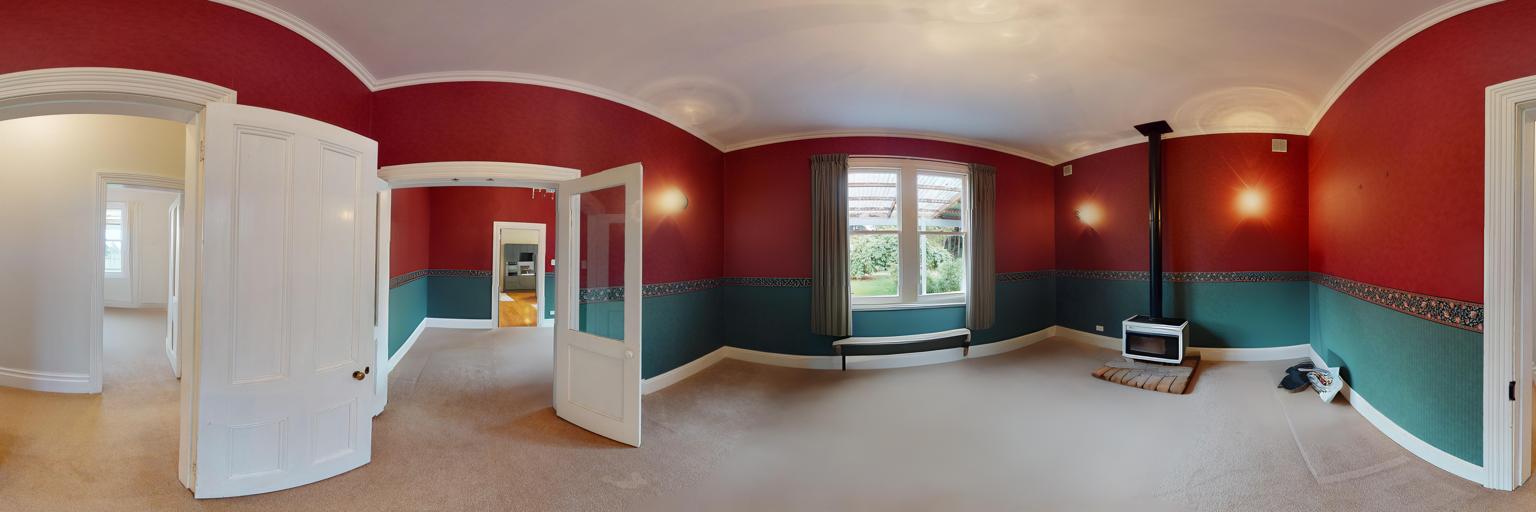}\\
 {\small{(a) Original}} $\qquad\qquad\qquad$ {\small{(b) Inpainted}}
 \caption{\textbf{Failure cases.} Structural changes or hallucinations may occur.}
 \label{fig:fail}
 \vspace{-2mm}
\end{figure}

Figure~\ref{ablate_2} shows how the image generated by our inpainting network is combined with the original image according to the inpainting mask. The generated image is low-resolution and often contains undesirable artifacts (\eg~there are no trees through the windows of \ref{ablate_2:b}), so we would prefer to use pixels of the original image when possible.
However, na\"ively replacing the generated image into the original image according to the inpainting mask (\ie~$\text{result} = \text{original} \cdot (1 - \text{mask}) + \text{generated} \cdot \text{mask}$) creates visually jarring outlines due to physical shadows and differences in white-balance.
While it is difficult to completely reconcile the differences between the detail frequencies of the original and generated images, our blending technique smoothly combines the two images.

Figure~\ref{ablate_1} demonstrates the effects of our rolling and padding pre-processing steps.
As the goal of these steps is to ensure that inpaints are consistent across the ``seam'' of the panorama image, we show the crops at opposite edges of the panorama joined together, so the panorama seam lies in the center of the image.
With no rolling, we obtain a vertical discontinuity where the seam lies.
With no padding, we obtain low-quality inpaints due to the reduced context.
While rolling and padding do not guarantee consistent inpainting around the entire panorama (\eg~this may fail when the mask spans the entire image), these two simple processing steps cover most cases.

\subsection{Limitations and future work}

While our method represents a step forward in defurnishing, it has limitations. 
Structural alterations and lingering hallucinations may occur, like the bench beneath the window in Figure~\ref{fig:fail}.
Our training methodology, which involves rendering synthetic furniture into unfurnished panoramas, introduces potential domain shift issues.
This discrepancy between synthetic and real-world data may impact the quality of results.
While we fixed the number of training prompts based on empirical evidence, the optimal prompting strategy remains an open question. 
Augmenting the fixed inference prompt ``empty room'' with additional contextual information could improve content generation accuracy. 

Our reliance on the Stable Diffusion 2.0 inpainting architecture imposes constraints, necessitating modifications for higher resolution training and inference. The resolution of the output of Stable Diffusion model is too low - only 512 pixels in height, and the superresolution model sometimes made the output warped or unnatural.
One area of improvement would be to choose an optimal superresolution model or to train one specifically on equirectangular images.

The applicability of our strategy to other generative models or imagery types besides indoor panoramas is out of the scope of this work and remains open for future research.

Finally, our method lacks consideration for multi-view consistency, which is important for applications like digital twins; each panorama is inpainted independently, overlooking valuable contextual information from other views. Future research should explore efficient and scalable approaches to incorporate geometric priors into the inpainting pipeline~\cite{zhang2023adding}.
Being able to inpaint consistently across different views of the same underlying physical space may help with defurnishing other representations of digital twins, such as textured polygon meshes.
\section{Conclusion}
\label{sec:conclusion}

We have presented a novel approach to panorama defurnishing by utilizing domain-specific fine-tuning for Stable Diffusion inpainting.
We have observed a notable reduction in undesirable hallucinations and improved the model's robustness to imperfect segmentation masks, making it much less likely to explain away shadows, light beams, reflections, and other similar effects. 
When compared to existing approaches, our method produces higher-quality results without the need for room layout estimation, as indicated in both qualitative and quantitative comparisons.
\section*{Acknowledgements}
\label{sec:ack}

We are greateful to Dorra Larnaout, Gregor Miller, Gunnar Hovden, Ky Waegel, Mykhaylo Kurinnyy, and Neil Jassal for their contributions.
We also thank Alexander Demidko and Kevin Balkoski for useful discussions.
\clearpage 
{
    \small
    \bibliographystyle{ieeenat_fullname}
    \bibliography{main}
}

\clearpage
\label{sec:additional}

\begin{figure*}
 \centering
 \includegraphics[width=0.33\linewidth]{fig/comp/ca9c4eca2d6f41aa8bf2d6a8a5407b15_eq_inpainted_comparesd}~
 \includegraphics[width=0.33\linewidth]{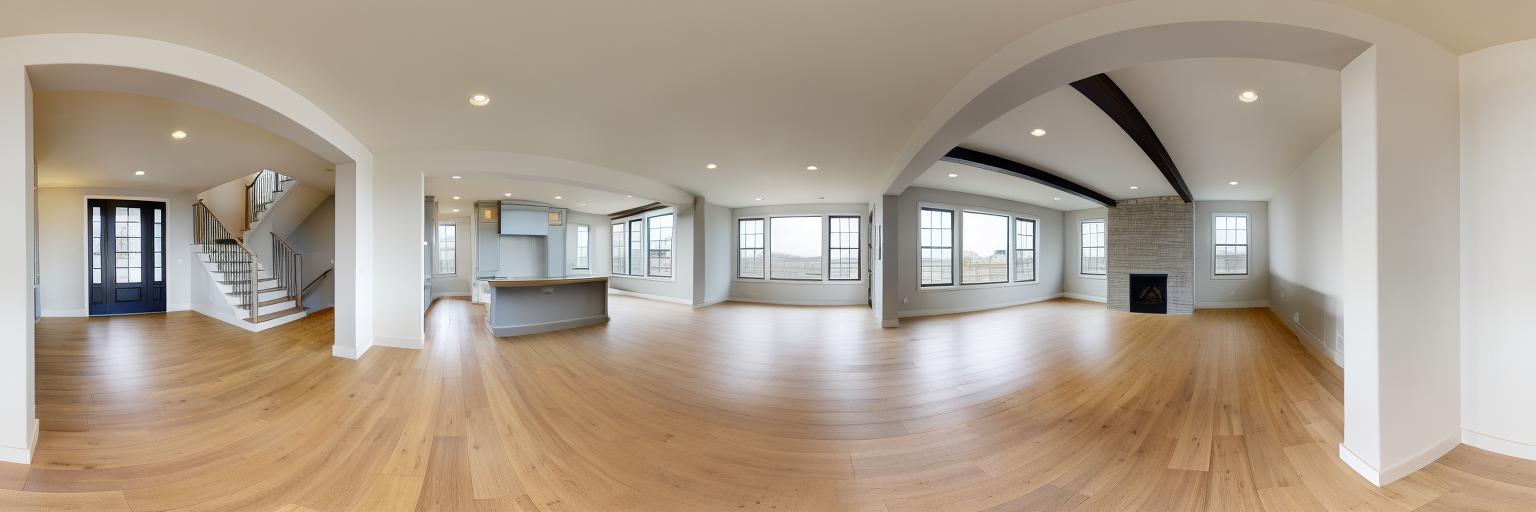}~
 \includegraphics[width=0.33\linewidth]{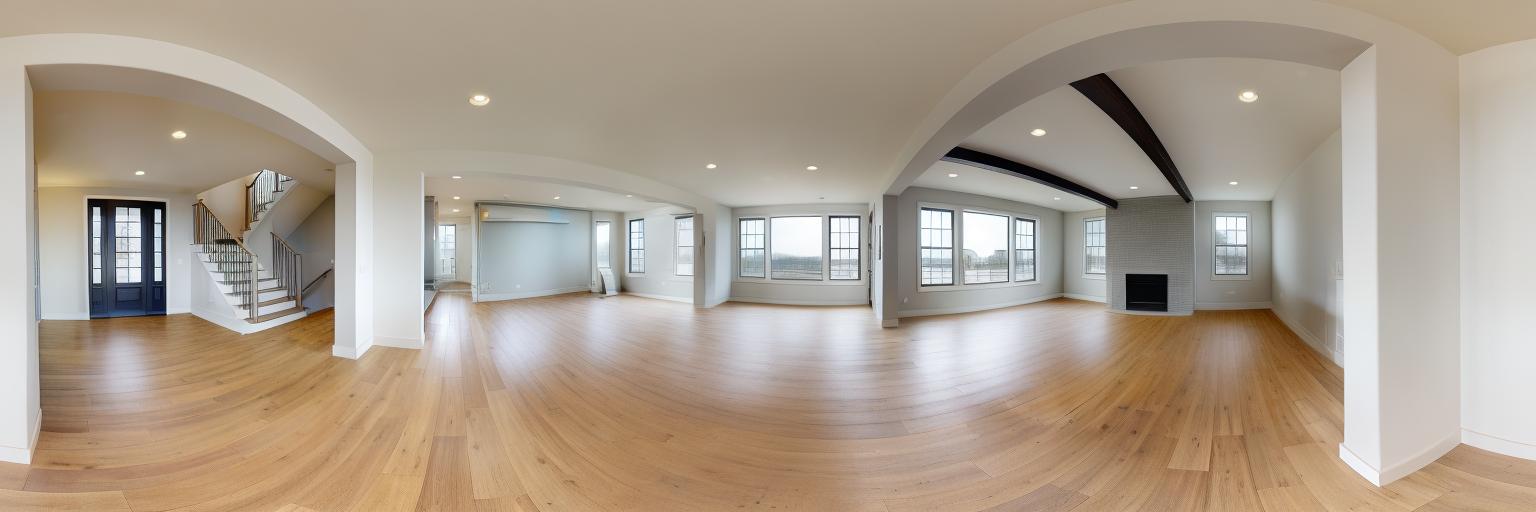}\\
 \includegraphics[width=0.33\linewidth]{fig/comp/a34d271e105744eabf60aa40f802c9d5_eq_inpainted_comparesd}~
 \includegraphics[width=0.33\linewidth]{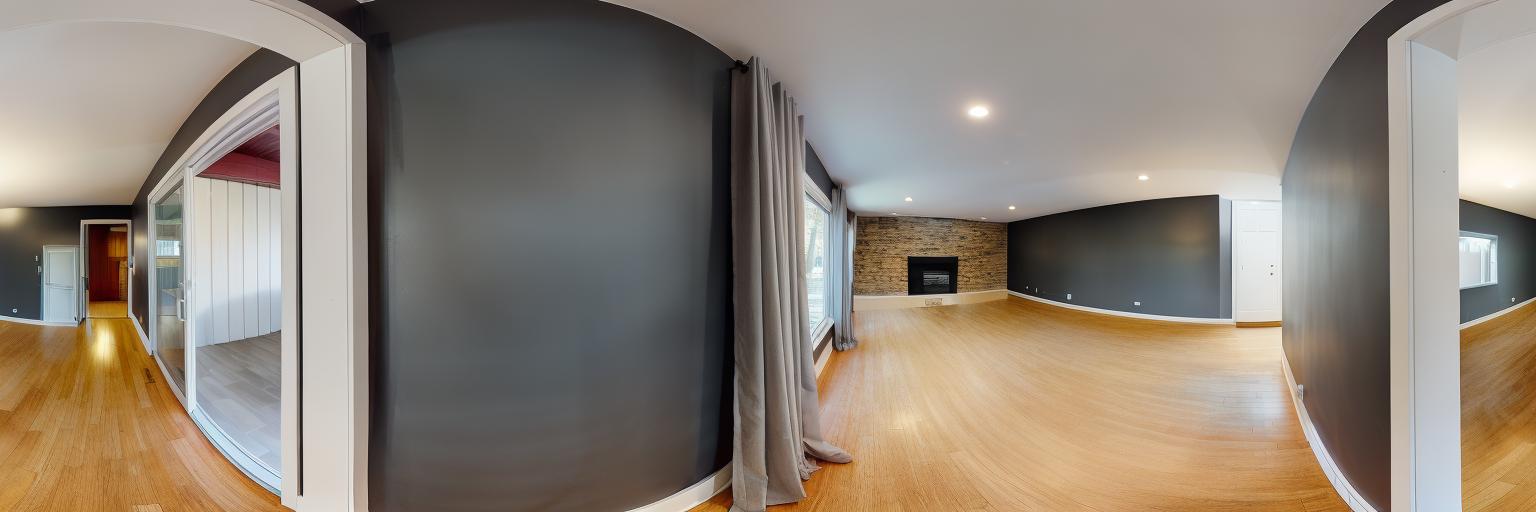}~
 \includegraphics[width=0.33\linewidth]{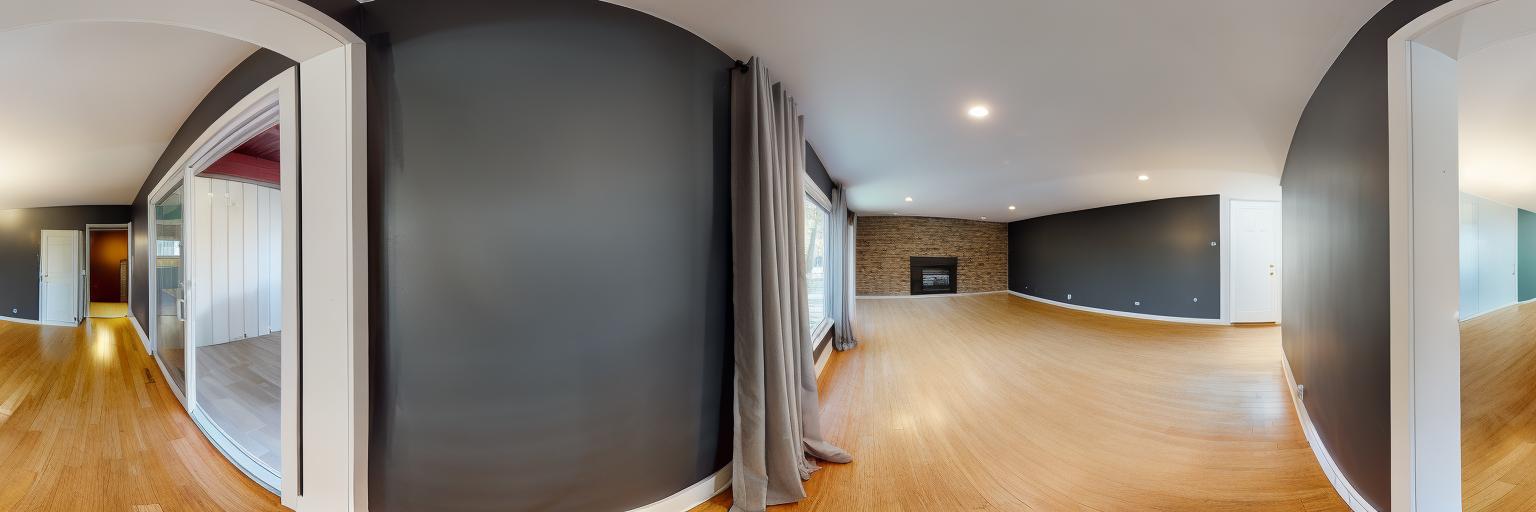}\\
 \includegraphics[width=0.33\linewidth]{fig/comp/9985900c24be46bfa6af793fb34e393d_eq_inpainted_comparesd}~
 \includegraphics[width=0.33\linewidth]{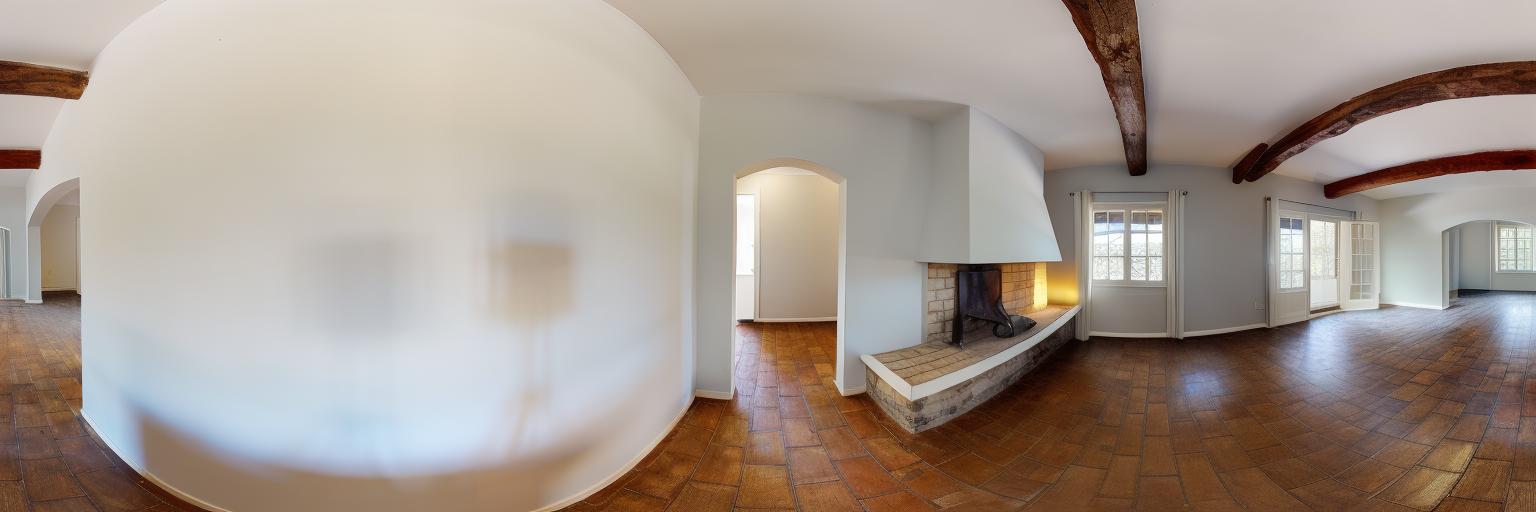}~
 \includegraphics[width=0.33\linewidth]{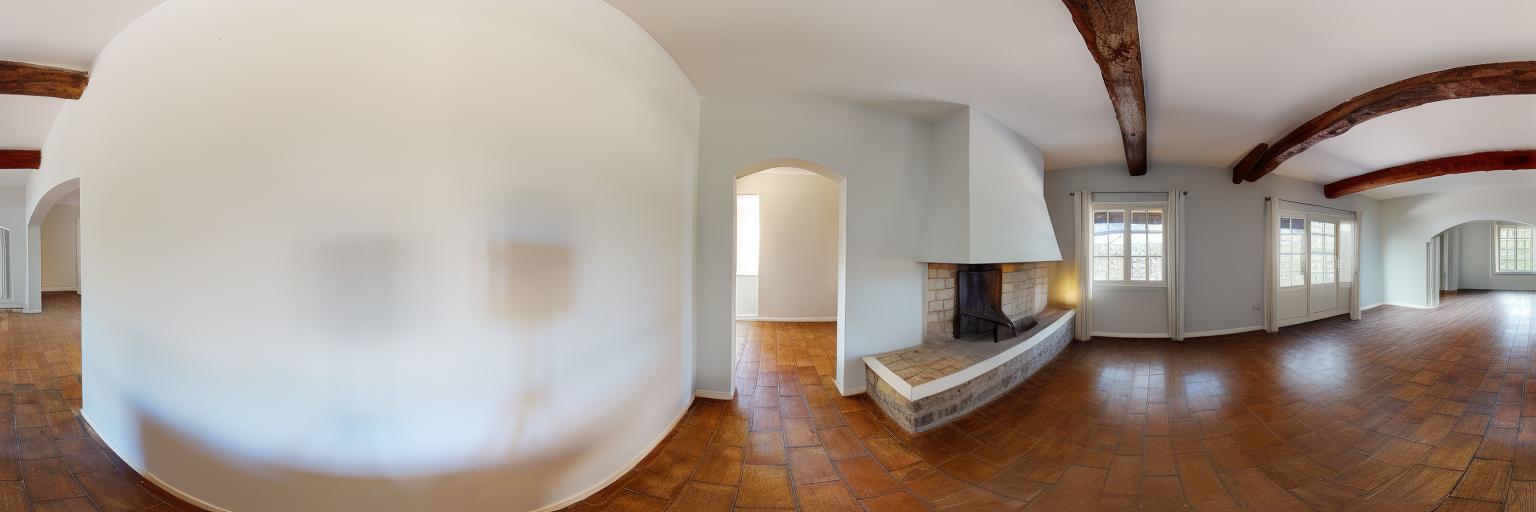}\\
 \includegraphics[width=0.33\linewidth]{fig/comp/eb59a0c1b01a4f7085fb3aec815cdde7_eq_inpainted_comparesd}~
 \includegraphics[width=0.33\linewidth]{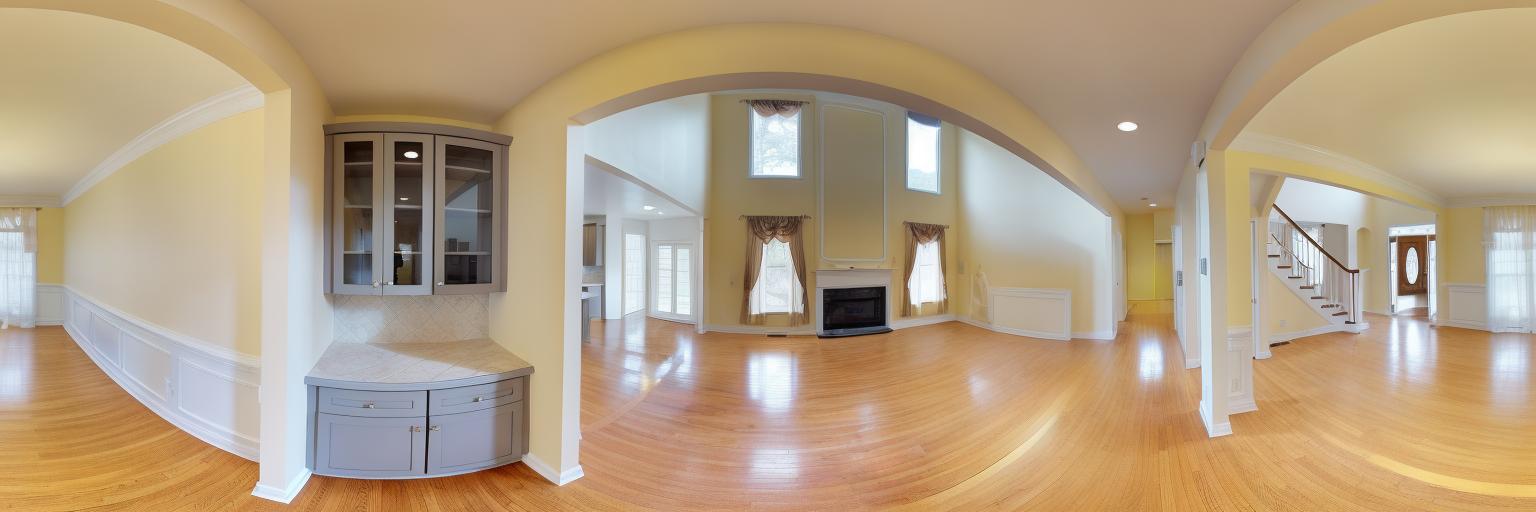}~
 \includegraphics[width=0.33\linewidth]{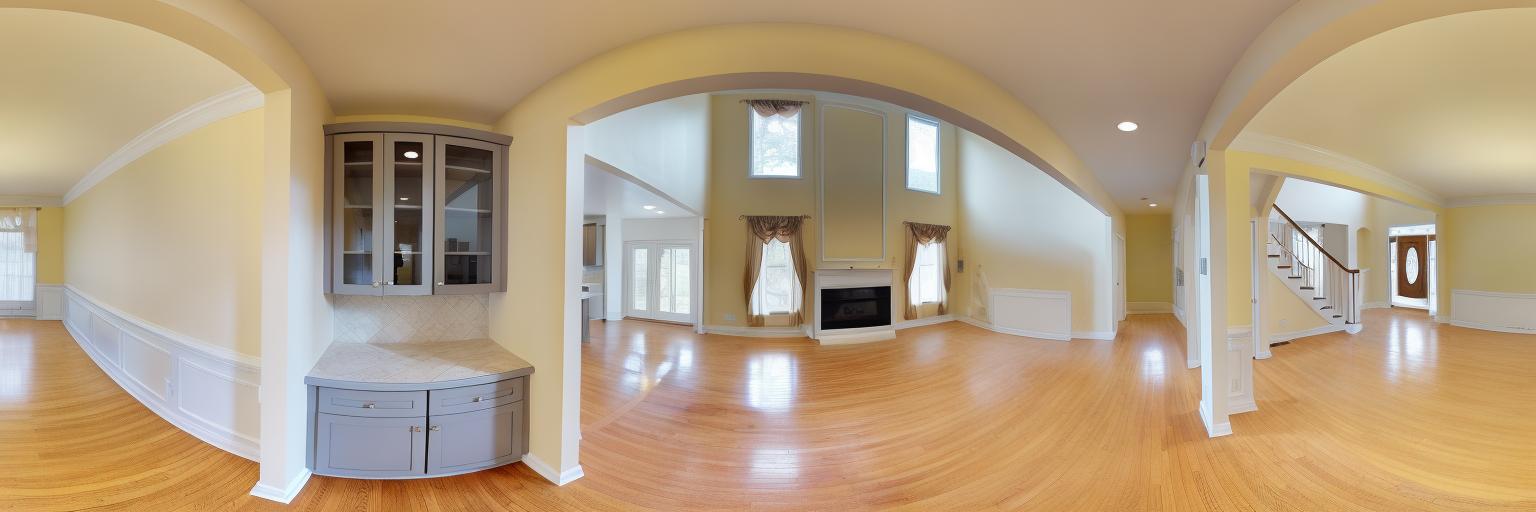}\\
 $\quad$ {\small{(a) No mask dilation}} $\qquad\qquad\qquad\qquad$ {\small{(b) Dilation 10 pixels}} $\qquad\qquad\qquad\qquad$ {\small{(c) Dilation 20 pixels}}
 \caption{\textbf{Effect of mask dilation on our inpainting.} We show results from our inpainting component only. Our results are nearly identical regardless of the amount of mask dilation, apart from far away details like the kitchen island in the first example.}
 \label{fig:supp_ours_dilation}
\end{figure*}

\begin{figure*}[t]
 \centering
 \begin{overpic}[width=0.495\linewidth]{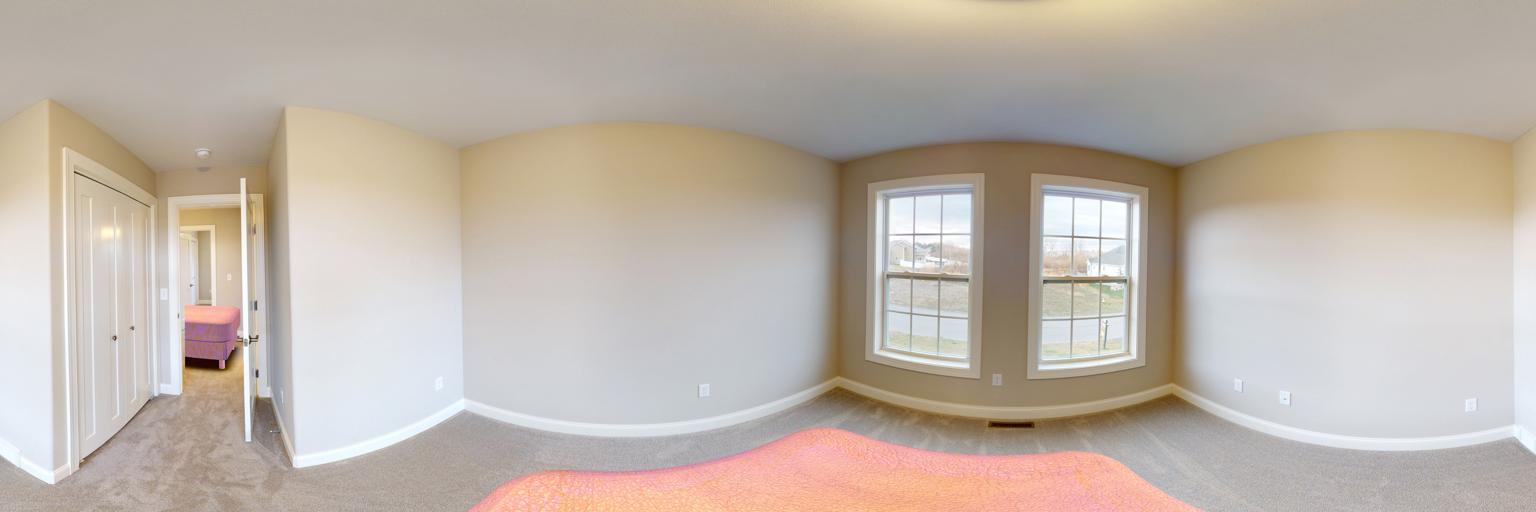}
  \put(2,2){\small{Original}}
 \end{overpic}~
 \begin{overpic}[width=0.495\linewidth]{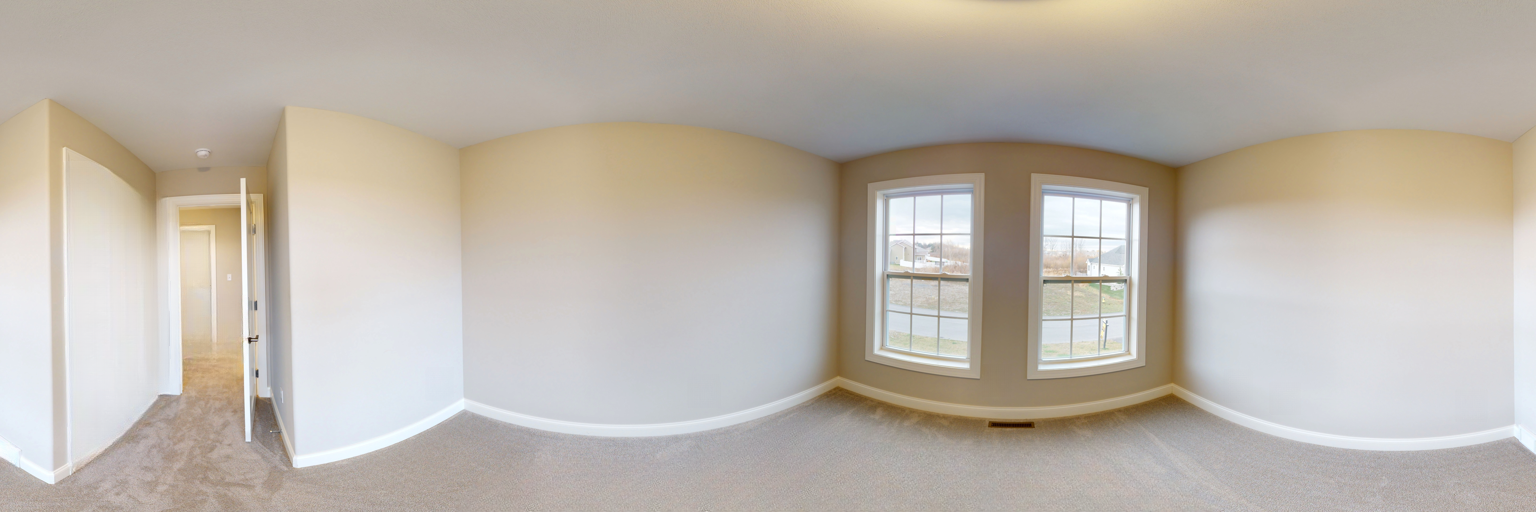}
  \put(2,2){\small{LaMa~\cite{suvorov2021lama}}}
 \end{overpic}\\
 \begin{overpic}[width=0.495\linewidth]{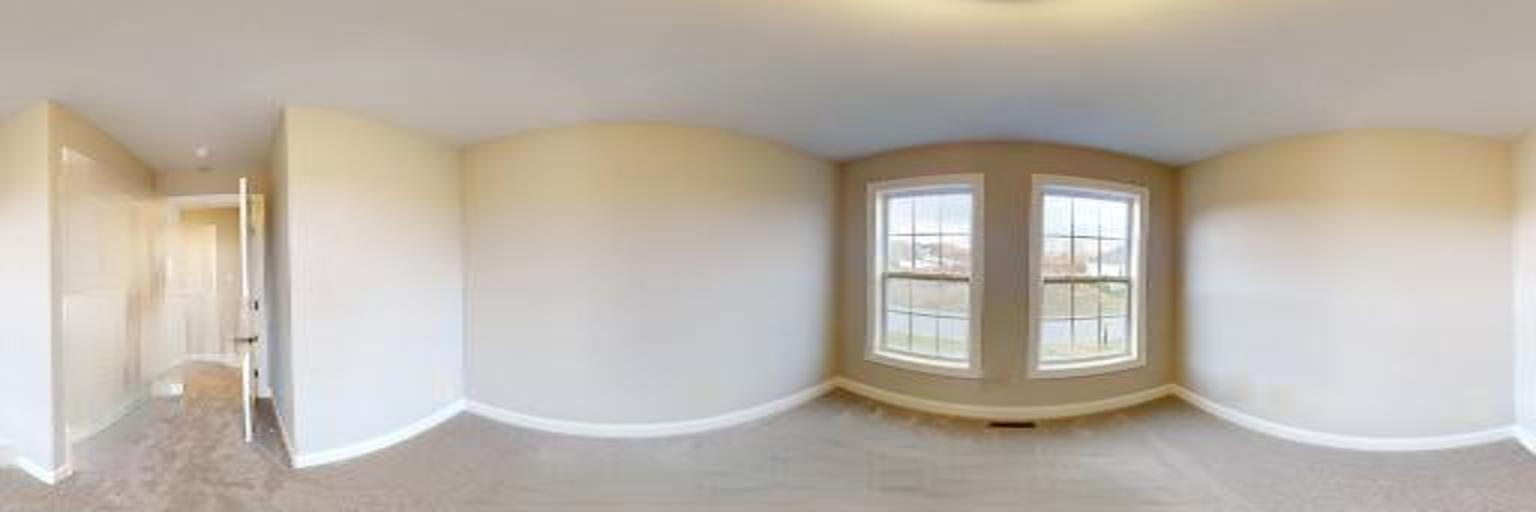}
  \put(2,2){\small{LGPN-Net~\cite{gao2022lgpn}}}
 \end{overpic}~
 \begin{overpic}[width=0.495\linewidth]{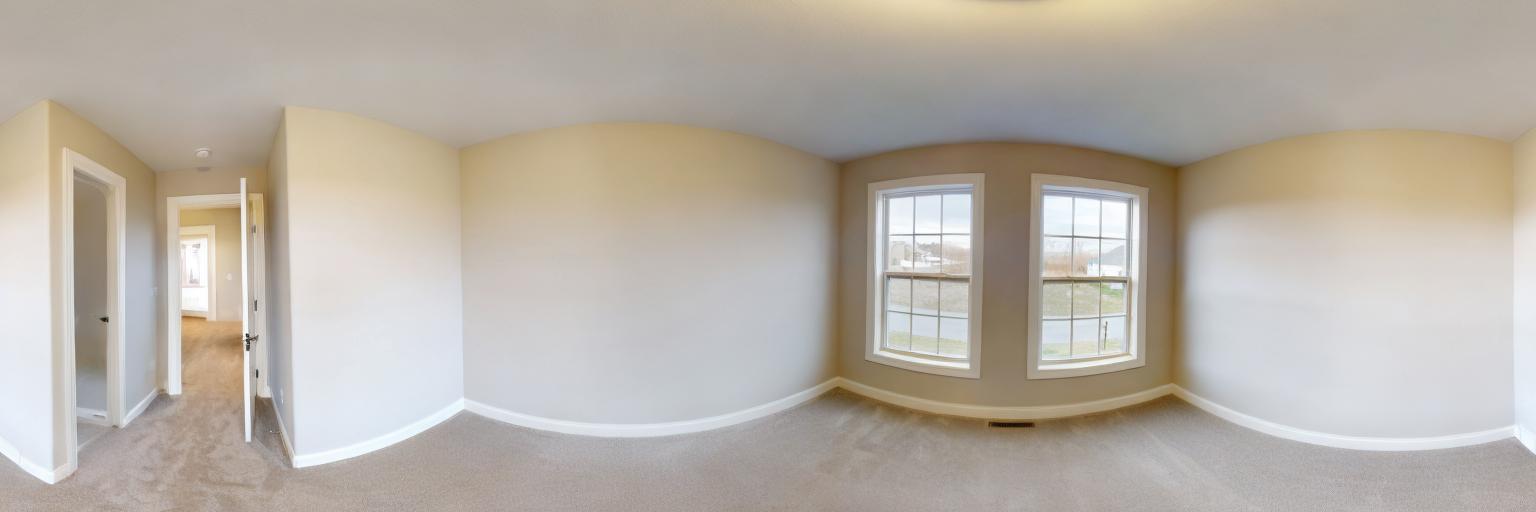}
  \put(2,2){\small{SD-2-inpaint}}
 \end{overpic}\\
 \begin{overpic}[width=0.495\linewidth]{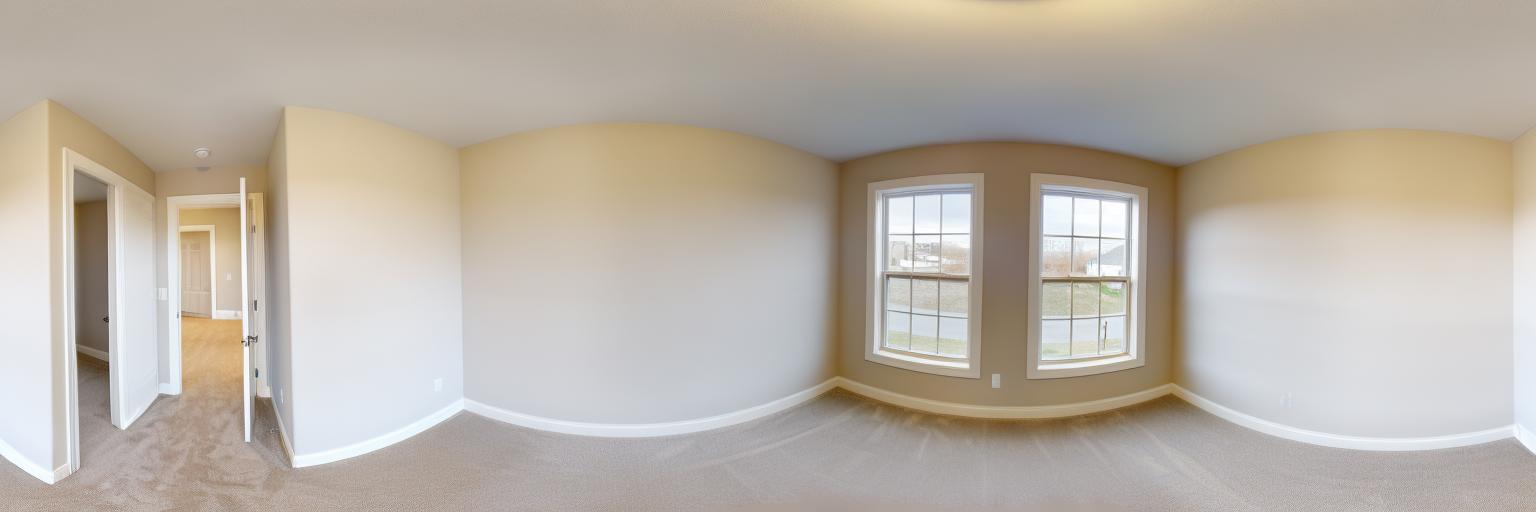}
  \put(2,2){\small{Ours-inpaint}}
 \end{overpic}~
 \begin{overpic}[width=0.495\linewidth]{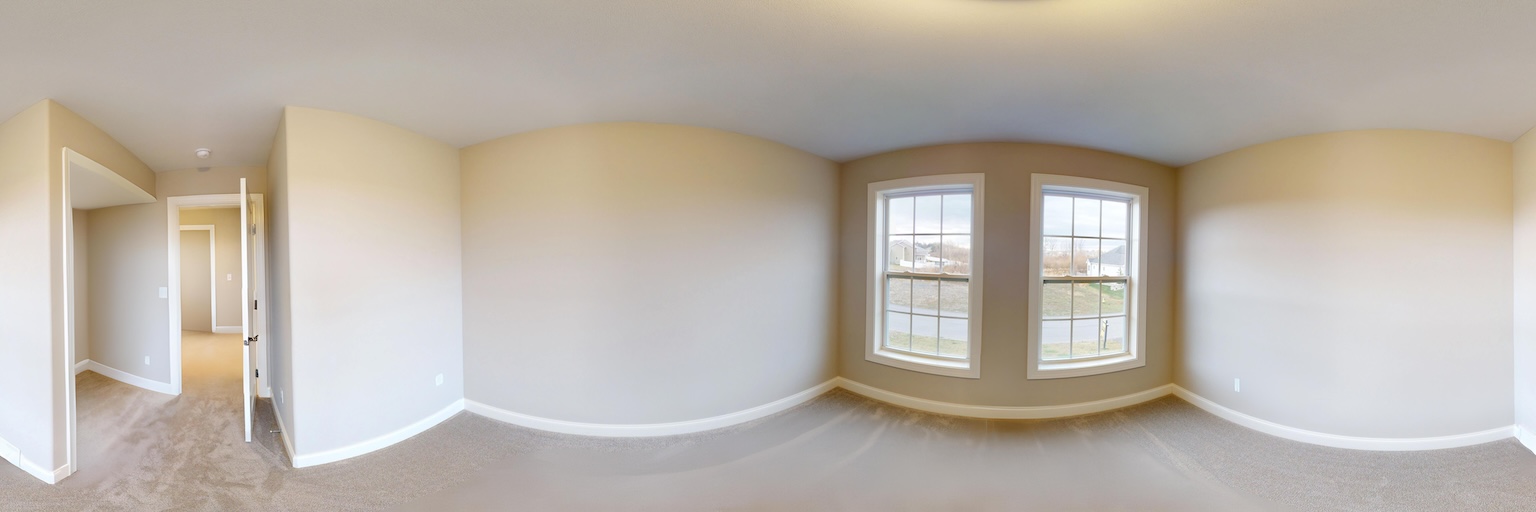}
  \put(2,2){\small{Ours-full}}
 \end{overpic}
 \caption{\textbf{Example of unfurnished space with synthetic furniture for quantitative evaluation.} The space is less complex than real furnished ones, letting LaMa and LGPN-Net produce higher frequency textures than SD and our inpainting, which is remedied by the blending in our full pipeline.}
 \label{fig:supp_quant_comp}
\end{figure*}

{\nopagebreak \onecolumn
{
\centering
\Large
\textbf{\thetitle}\\
\vspace{0.5em}Supplementary Material \\
\vspace{1.0em}
}
{\centering

\author{
Mira Slavcheva$^\assumption$
\quad
Dave Gausebeck$^\assumption$
\quad
Kevin Chen$^\assumption$
\quad
David Buchhofer
\quad
Azwad Sabik
\quad
Chen Ma\\
Sachal Dhillon
\quad
Olaf Brandt
\quad
Alan Dolhasz \vspace*{2mm} \\ 
Matterport\\
}
}

\section*{Visual results}
\blfootnote{
\hspace*{-5mm}$^\assumption$ denotes equal contribution.\\
\faEnvelope[regular] {\tt research@matterport.com}
}
In Figure~\ref{fig:supp_comp}, we provide larger versions of the defurnishing results from the main paper. 
In addition, we include results from the methods we compare to in Section~\ref{sec:comparisons}, \ie~LaMa~\cite{suvorov2021lama}, LGPN-Net~\cite{gao2022lgpn} and SD-inpainting~\cite{rombach2022highresolution}, with a smaller mask dilation of 10 pixels as well as with no mask dilation.

LaMa and LGPN-Net demonstrate similar trends; a non-dilated mask is absolutely insufficient---large dark patches are inpainted and the outlines of the inpainting mask are recognizable. Mask dilation remedies this effect, but the inpainted region tends to look like a blurry blob with increasing mask size. Conversely, the textures are sharpest with the non-dilated masks.

In all examples, SD-inpainting hallucinates objects when the inpainting mask is not dilated. 
With a mask dilated by 10 pixels, objects are still hallucinated in the first two examples. 
With a mask dilated by 20 pixels, only the first example shows very noticeable hallucinations, \eg~the tables on the right, but larger shadows cast by the objects that are being removed, \eg~the couch in the second example, remain in the output image and may even be extended by the inpainting.

Note that in the main paper, we chose the best-looking result for each of these three methods and each example separately.

Ours-inpaint builds upon SD-inpainting but tackles the hallucinations---and indeed none of the examples have hallucinations. Ours-full makes sure that original textures are preserved wherever possible, which is valuable because of the original image has more detail, \eg~the kitchen island in the first example, and because our method may sometimes remove more details than necessary since it is specifically trained to remove shadows outside of the inpainting mask. Intricate textures may still be an issue for our method, like the floor in the last image, where one plank is inpainted in a notably darker color.

In addition, in Figure~\ref{fig:supp_ours_dilation} we demonstrate that the inpainting component of our method is not influenced by mask dilation. There is no noticeable difference in the inpainted result in all examples but the first one, where the far away kitchen island is removed as the mask gets larger, while it should remain because it is built-in. This example demonstrates the usefulness of non-dilated masks for keeping far-away details intact.

Finally, in Figure~\ref{fig:supp_quant_comp} we show an example of an unfurnished space with synthetic furniture used for quantitative evaluation in Section~\ref{sec:comparisons}.

\section*{Prompt set}
The set of 32 prompts that we used for training is as follows:
\begin{align*}
    \{ X \;\; Y \;\; Z \},& \text{ where}\\
    X = \{ u \; V \},& \text{ where } u \in \{\emptyset, \text{``an''}\}, V \in \{ \text{``empty''}, \text{``unfurnished''} \},\\
    Y \in &\;\{ \text{``room''}, \text{``space''}, \text{``home''}, \text{``house''} \},\\
    Z \in \{ \emptyset, P \text{ if } u \neq \emptyset, P Q \text{ if } u \neq \emptyset \},& \text{ where } P = \text{``. uniformly blank''}, Q = \text{``, straight edges''}.
    \end{align*}
Note that during training, each image is assigned one prompt at random. Due to this randomness, the prompt in subsequent epochs for the same image might be different.
}

\begin{figure*}
 \centering
 \begin{overpic}[width=0.495\linewidth]{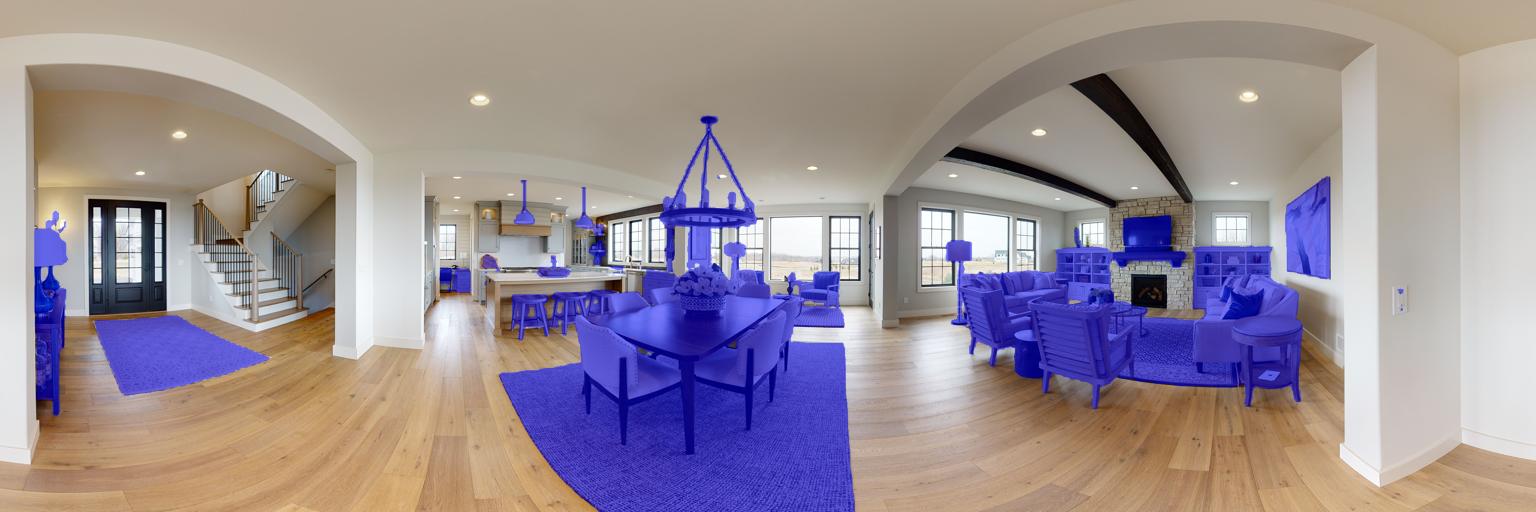}
    \put(2,2){\color{white}\small{Input panorama}}
 \end{overpic}~
 \begin{overpic}[width=0.495\linewidth]{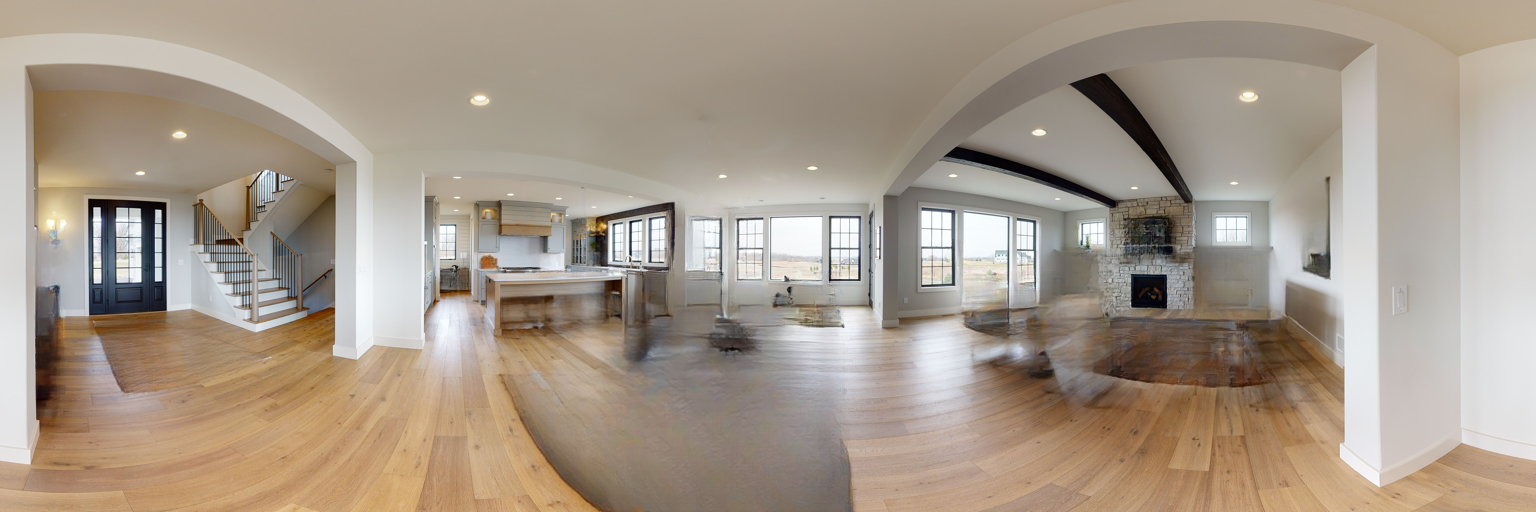}
    \put(2,2){\color{white}\small{LaMa no dilation}}
 \end{overpic}\\
 \begin{overpic}[width=0.495\linewidth]{fig/comp/ca9c4eca2d6f41aa8bf2d6a8a5407b15_eq_inpainted_comparesd}
    \put(2,2){\color{white}\small{Ours-inpaint}}
 \end{overpic}~
 \begin{overpic}[width=0.495\linewidth]{fig/comp/ca9c4eca2d6f41aa8bf2d6a8a5407b15_eq_mask_lama_dilate10_noref}
    \put(2,2){\color{white}\small{LaMa dilation 10px}}
 \end{overpic}\\
 \begin{overpic}[width=0.495\linewidth]{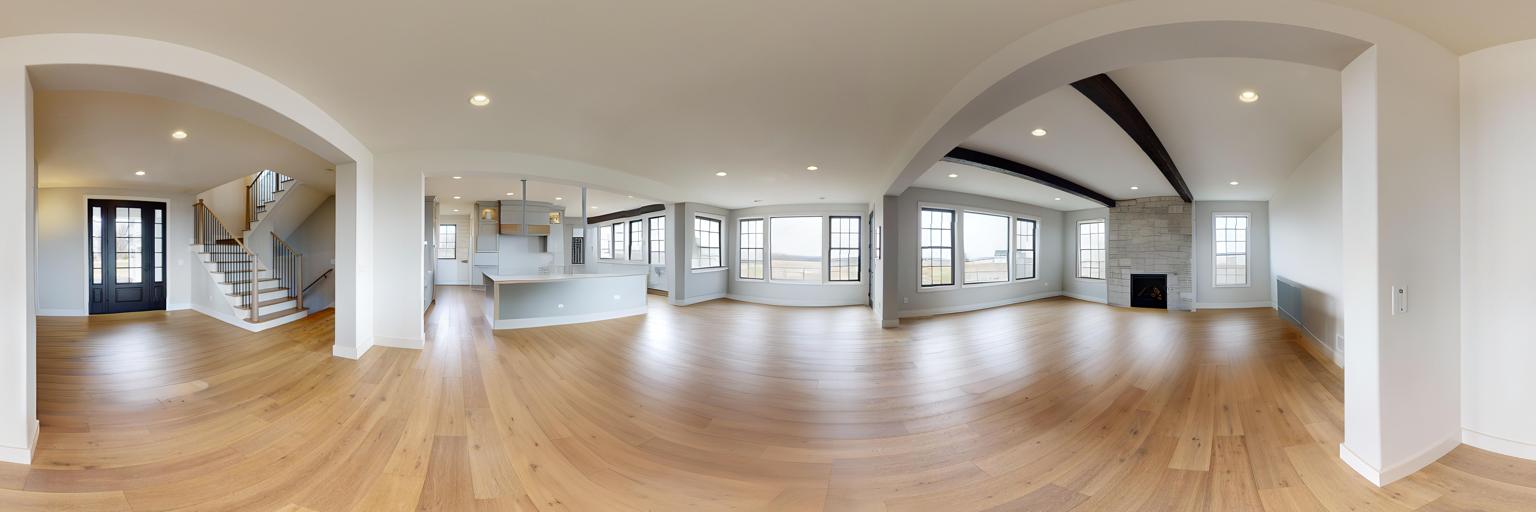}
    \put(2,2){\color{white}\small{Ours-full}}
 \end{overpic}~
 \begin{overpic}[width=0.495\linewidth]{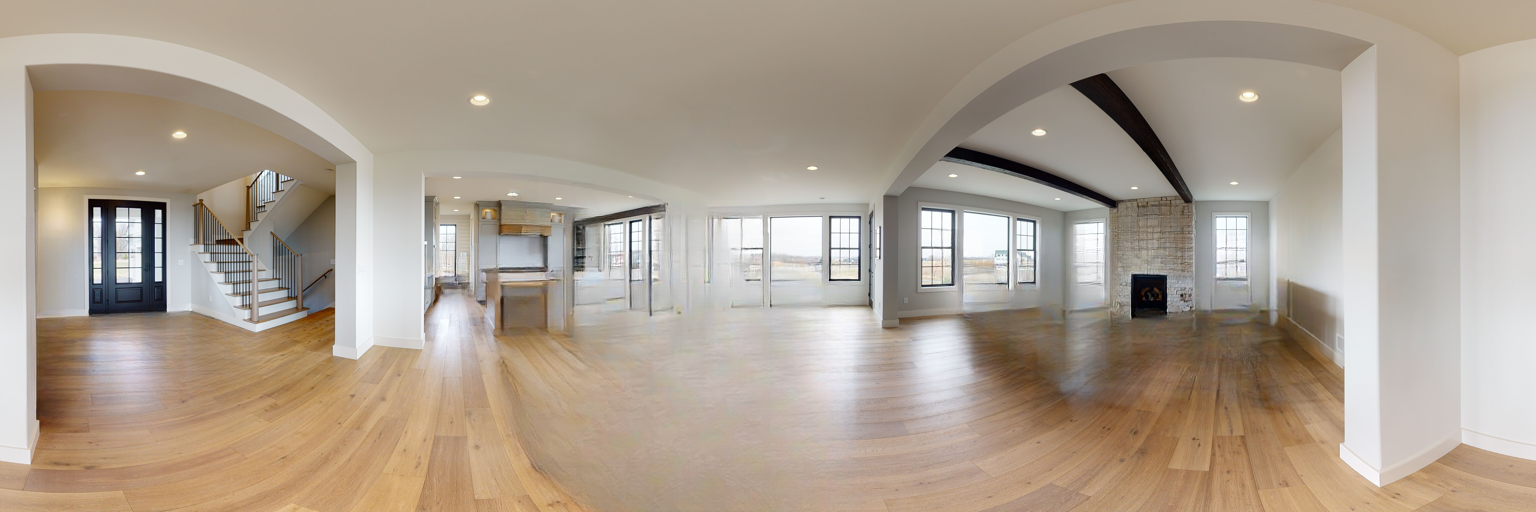}
    \put(2,2){\color{white}\small{LaMa dilation 20px}}
 \end{overpic}\\
 \begin{overpic}[width=0.495\linewidth]{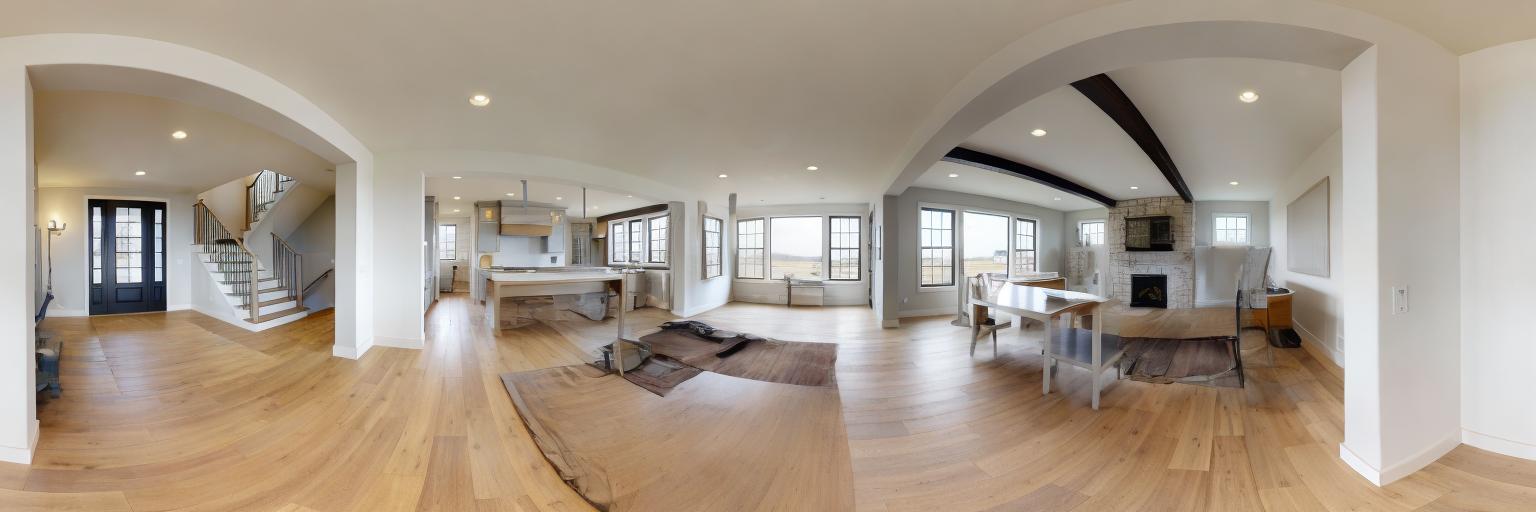}
    \put(2,2){\color{white}\small{SD-2-inpaint no dilation}}
 \end{overpic}~
 \begin{overpic}[width=0.495\linewidth,height=0.165\linewidth]{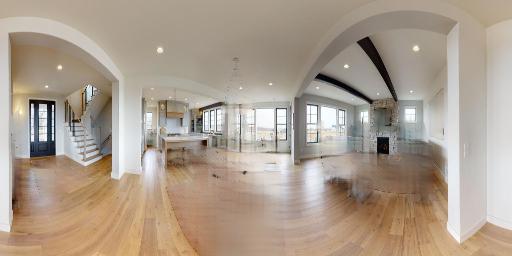}
    \put(2,2){\color{white}\small{LGPN-Net no dilation}}
 \end{overpic}\\
 \begin{overpic}[width=0.495\linewidth]{fig/comp/ca9c4eca2d6f41aa8bf2d6a8a5407b15_eq_vanilla_10dilate}
    \put(2,2){\color{white}\small{SD-2-inpaint dilation 10px}}
 \end{overpic}~
 \begin{overpic}[width=0.495\linewidth,height=0.165\linewidth]{fig/comp/ca9c4eca2d6f41aa8bf2d6a8a5407b15_eq_inpainted_lgpn_dilate10}
    \put(2,2){\color{white}\small{LGPN-Net dilation 10px}}
 \end{overpic}\\
 \begin{overpic}[width=0.495\linewidth]{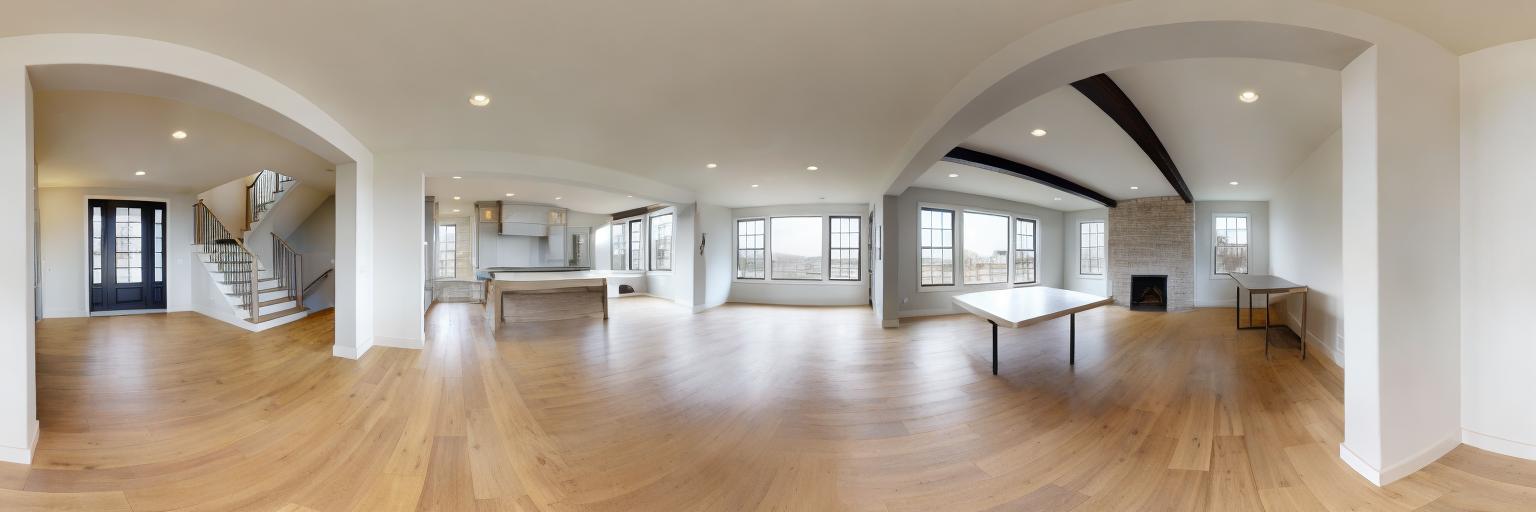}
    \put(2,2){\color{white}\small{SD-2-inpaint dilation 20px}}
 \end{overpic}~
 \begin{overpic}[width=0.495\linewidth,height=0.165\linewidth]{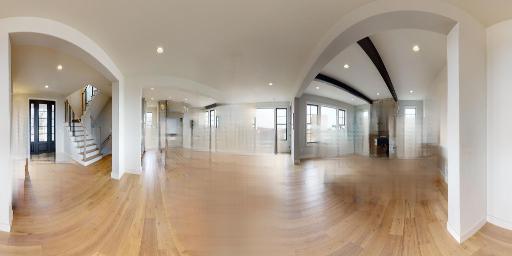}
    \put(2,2){\color{white}\small{LGPN-Net dilation 20px}}
 \end{overpic}
 \caption{\textbf{Additional defurnishing comparisons.} The non-dilated mask is overlaid in blue. Image best viewed digitally.}
 \label{fig:supp_comp}
\end{figure*}

\begin{figure*}\ContinuedFloat
 \centering
 \begin{overpic}[width=0.495\linewidth]{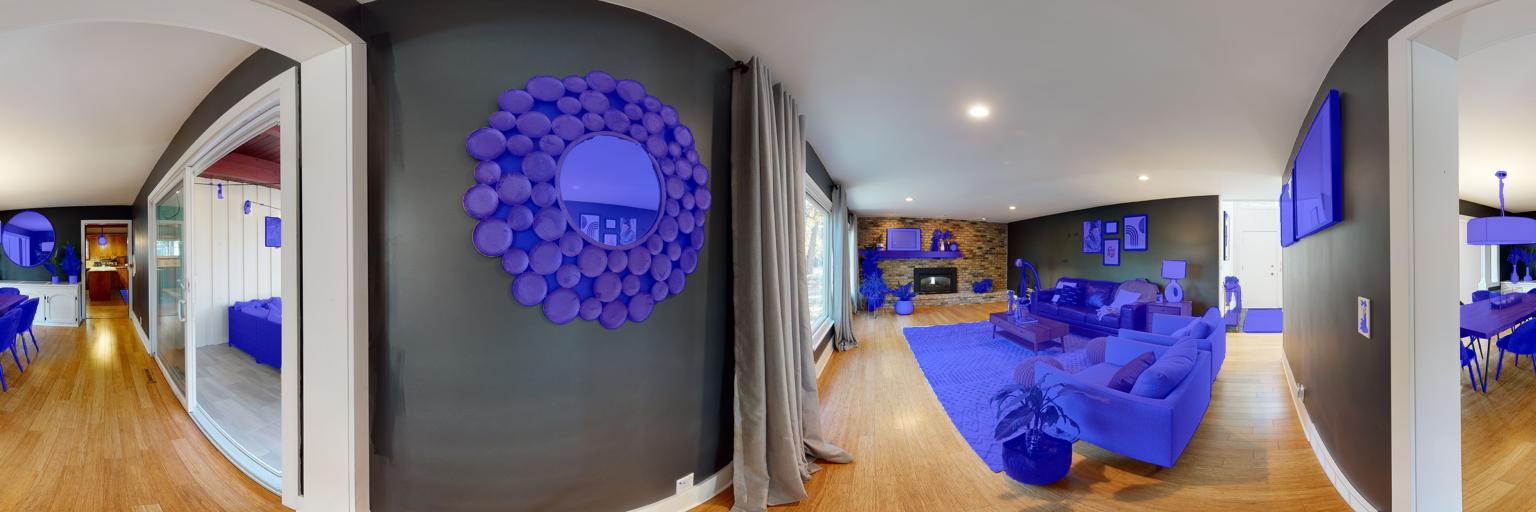}
    \put(2,2){\color{white}\small{Input panorama}}
 \end{overpic}~
 \begin{overpic}[width=0.495\linewidth]{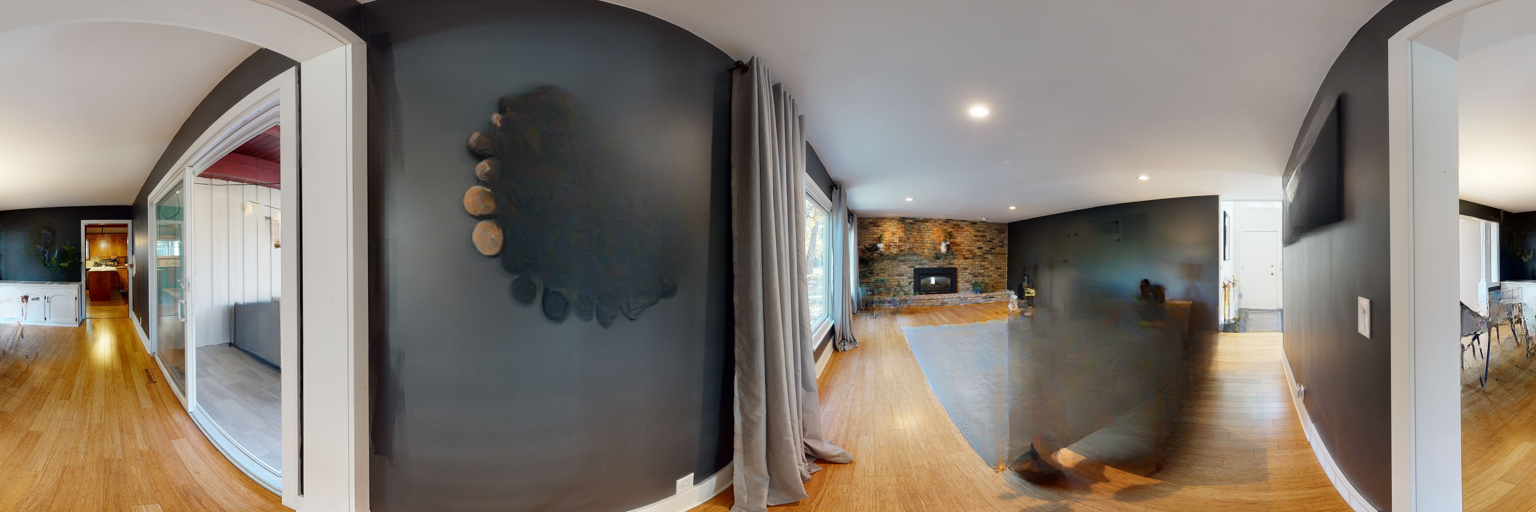}
    \put(2,2){\color{white}\small{LaMa no dilation}}
 \end{overpic}\\
 \begin{overpic}[width=0.495\linewidth]{fig/comp/a34d271e105744eabf60aa40f802c9d5_eq_inpainted_comparesd}
    \put(2,2){\color{white}\small{Ours-inpaint}}
 \end{overpic}~
 \begin{overpic}[width=0.495\linewidth]{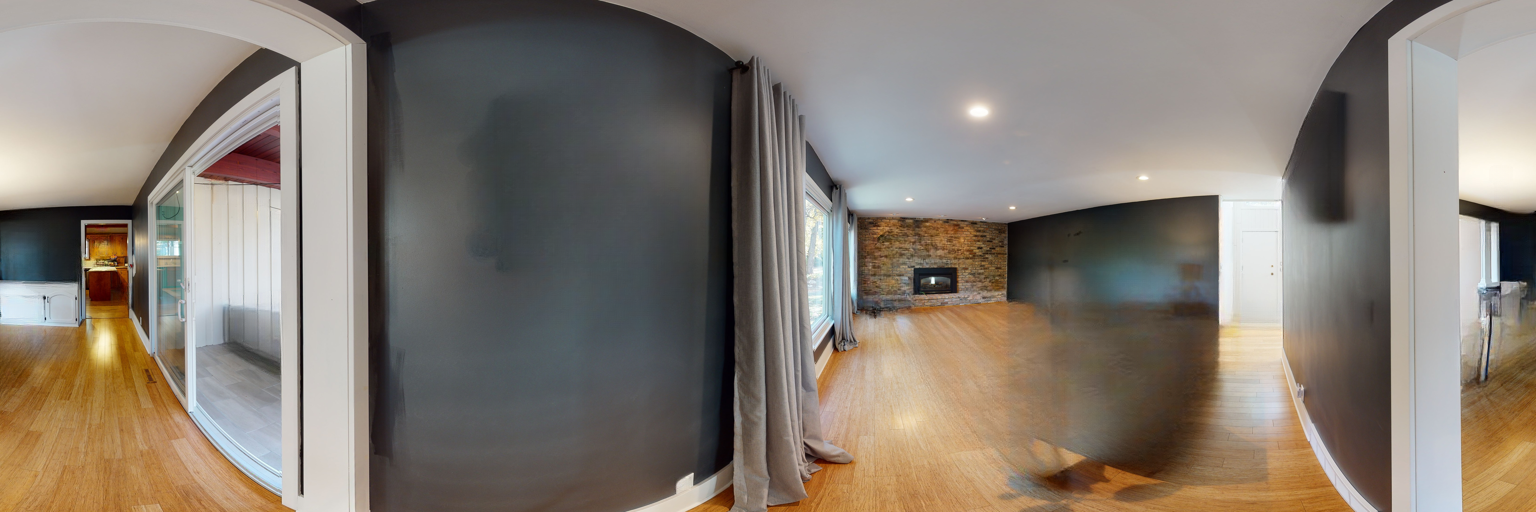}
    \put(2,2){\color{white}\small{LaMa dilation 10px}}
 \end{overpic}\\
 \begin{overpic}[width=0.495\linewidth]{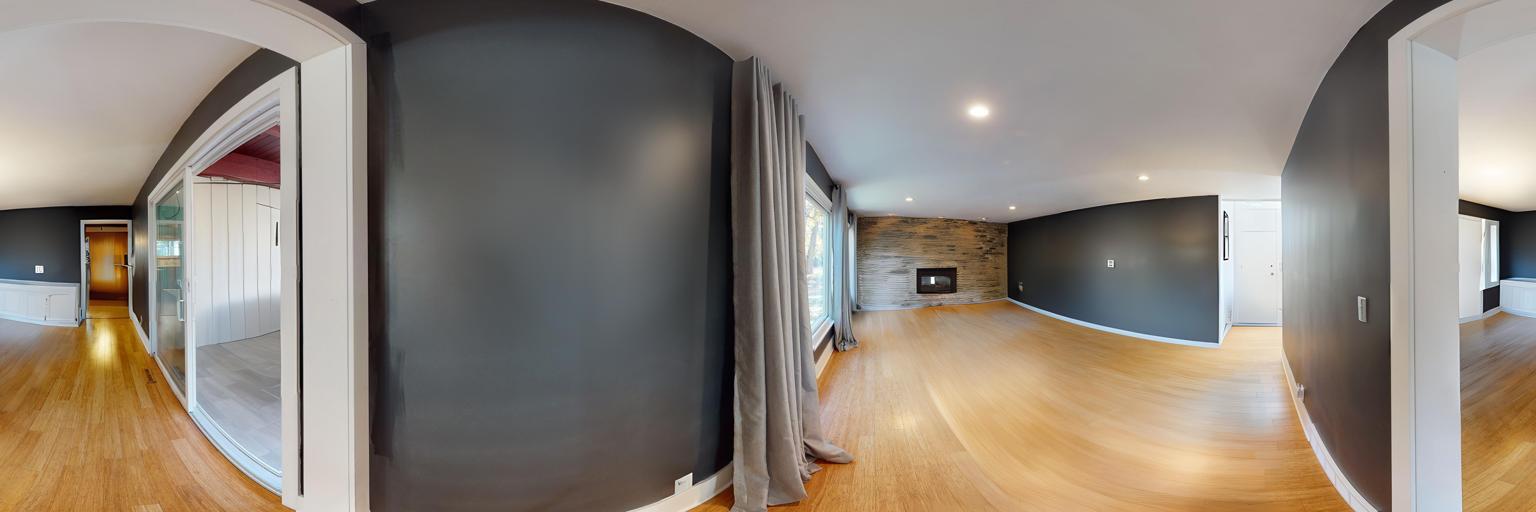}
    \put(2,2){\color{white}\small{Ours-full}}
 \end{overpic}~
 \begin{overpic}[width=0.495\linewidth]{fig/comp/a34d271e105744eabf60aa40f802c9d5_eq_mask_lama_dilate20_noref}
    \put(2,2){\color{white}\small{LaMa dilation 20px}}
 \end{overpic}\\
 \begin{overpic}[width=0.495\linewidth]{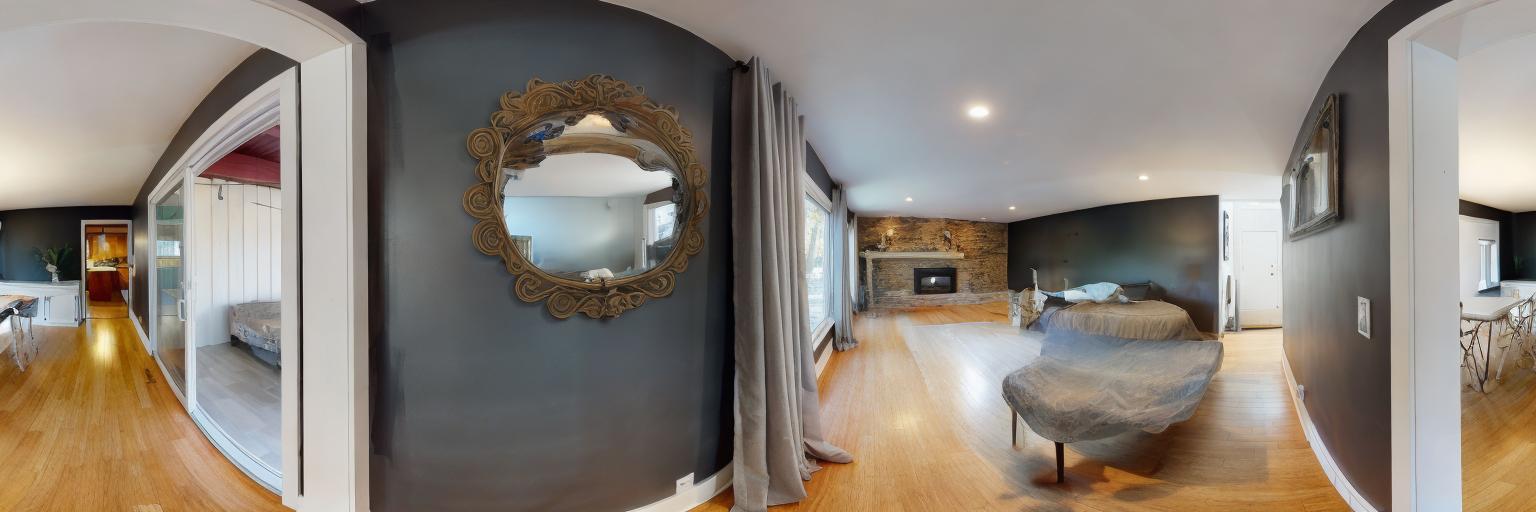}
    \put(2,2){\color{white}\small{SD-2-inpaint no dilation}}
 \end{overpic}~
 \begin{overpic}[width=0.495\linewidth,height=0.165\linewidth]{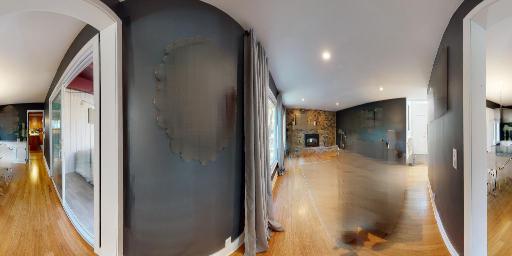}
    \put(2,2){\color{white}\small{LGPN-Net no dilation}}
 \end{overpic}\\
 \begin{overpic}[width=0.495\linewidth]{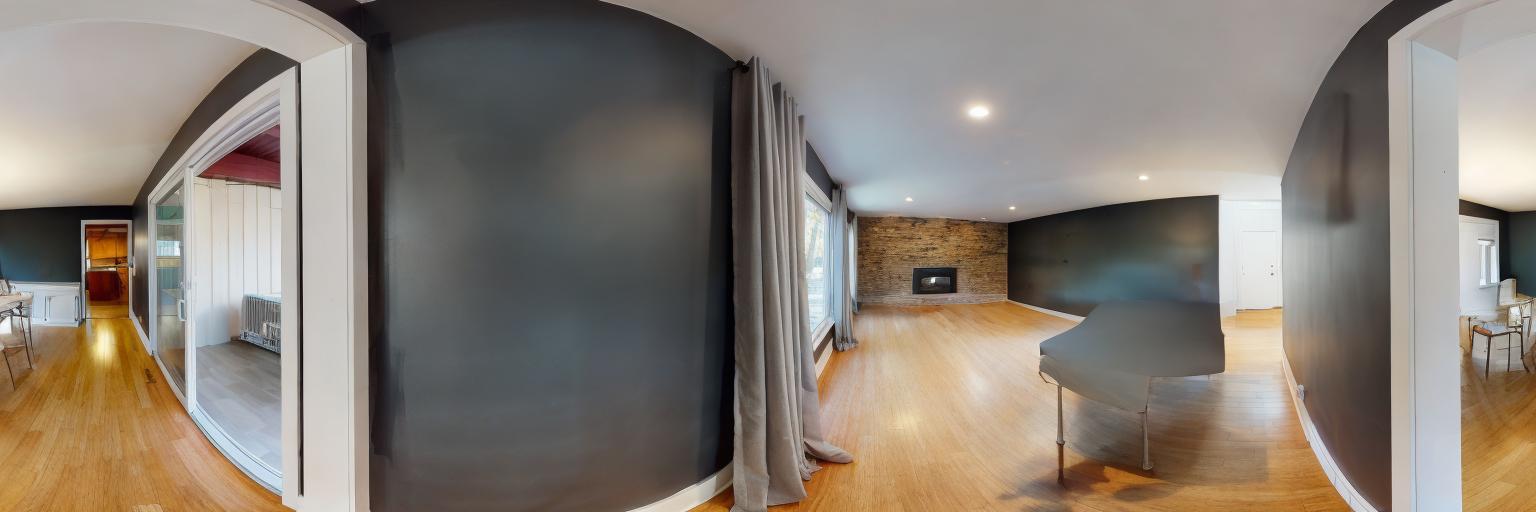}
    \put(2,2){\color{white}\small{SD-2-inpaint dilation 10px}}
 \end{overpic}~
 \begin{overpic}[width=0.495\linewidth,height=0.165\linewidth]{fig/comp/a34d271e105744eabf60aa40f802c9d5_eq_inpainted_lgpn_dilate10}
    \put(2,2){\color{white}\small{LGPN-Net dilation 10px}}
 \end{overpic}\\
 \begin{overpic}[width=0.495\linewidth]{fig/comp/a34d271e105744eabf60aa40f802c9d5_eq_vanilla_20dilate}
    \put(2,2){\color{white}\small{SD-2-inpaint dilation 20px}}
 \end{overpic}~
 \begin{overpic}[width=0.495\linewidth,height=0.165\linewidth]{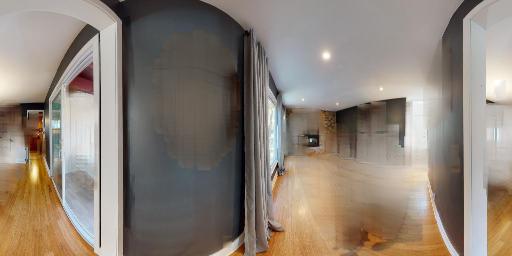}
    \put(2,2){\color{white}\small{LGPN-Net dilation 20px}}
 \end{overpic}
 \caption{\textbf{Additional defurnishing comparisons.} The non-dilated mask is overlaid in blue. Image best viewed digitally.}
\end{figure*}

\begin{figure*}\ContinuedFloat
 \centering
 \begin{overpic}[width=0.495\linewidth]{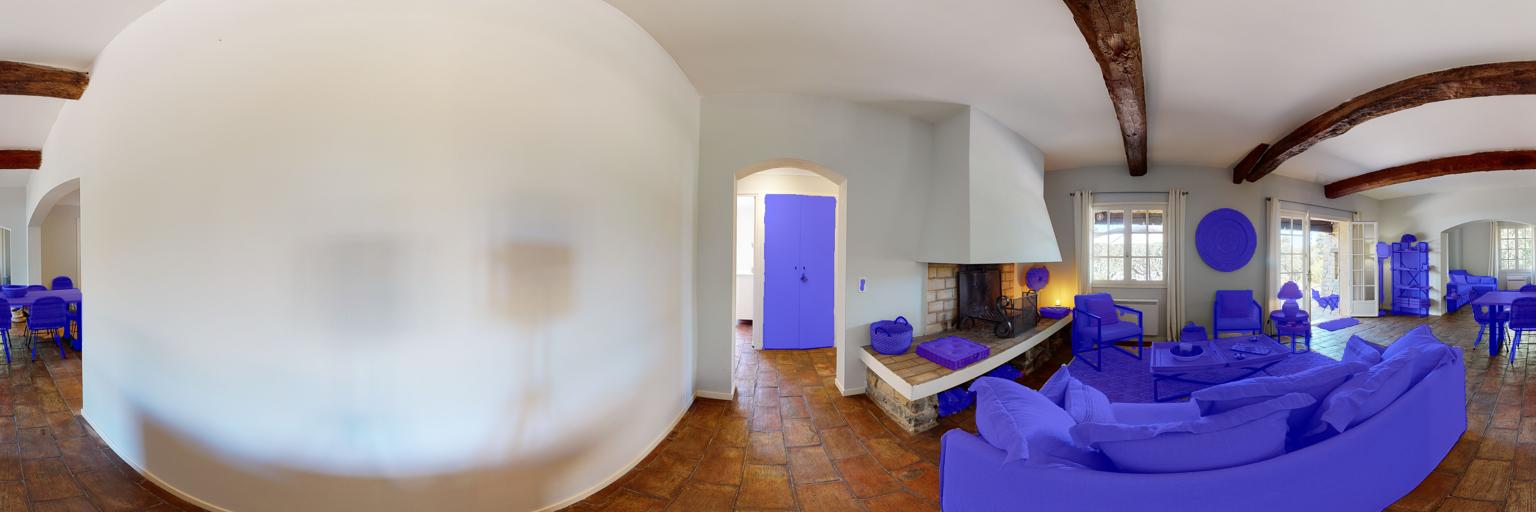}
    \put(2,2){\color{white}\small{Input panorama}}
 \end{overpic}~
 \begin{overpic}[width=0.495\linewidth]{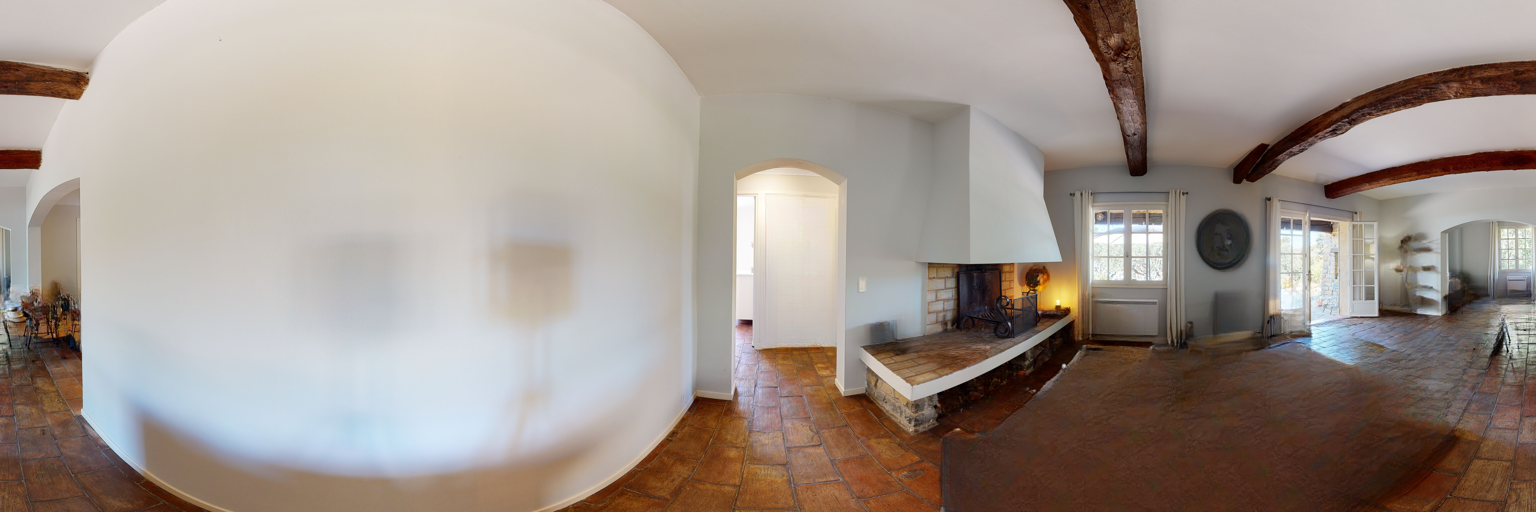}
    \put(2,2){\color{white}\small{LaMa no dilation}}
 \end{overpic}\\
 \begin{overpic}[width=0.495\linewidth]{fig/comp/9985900c24be46bfa6af793fb34e393d_eq_inpainted_comparesd}
    \put(2,2){\color{white}\small{Ours-inpaint}}
 \end{overpic}~
 \begin{overpic}[width=0.495\linewidth]{fig/comp/9985900c24be46bfa6af793fb34e393d_eq_mask_lama_dilate10_noref}
    \put(2,2){\color{white}\small{LaMa dilation 10px}}
 \end{overpic}\\
 \begin{overpic}[width=0.495\linewidth]{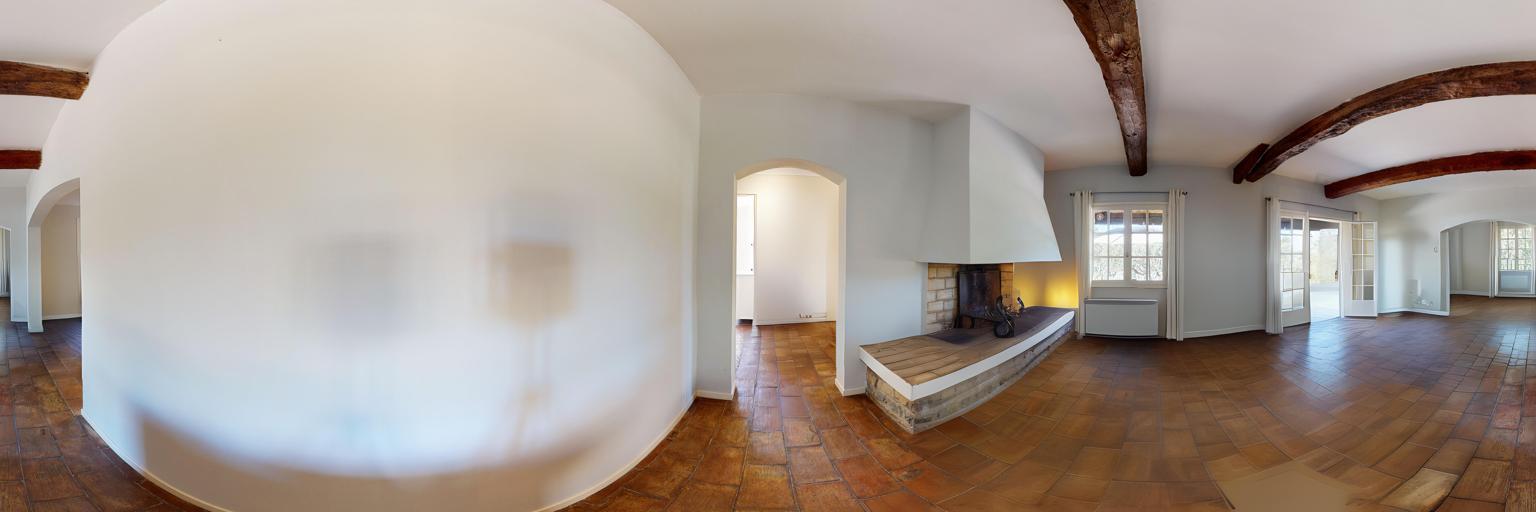}
    \put(2,2){\color{white}\small{Ours-full}}
 \end{overpic}~
 \begin{overpic}[width=0.495\linewidth]{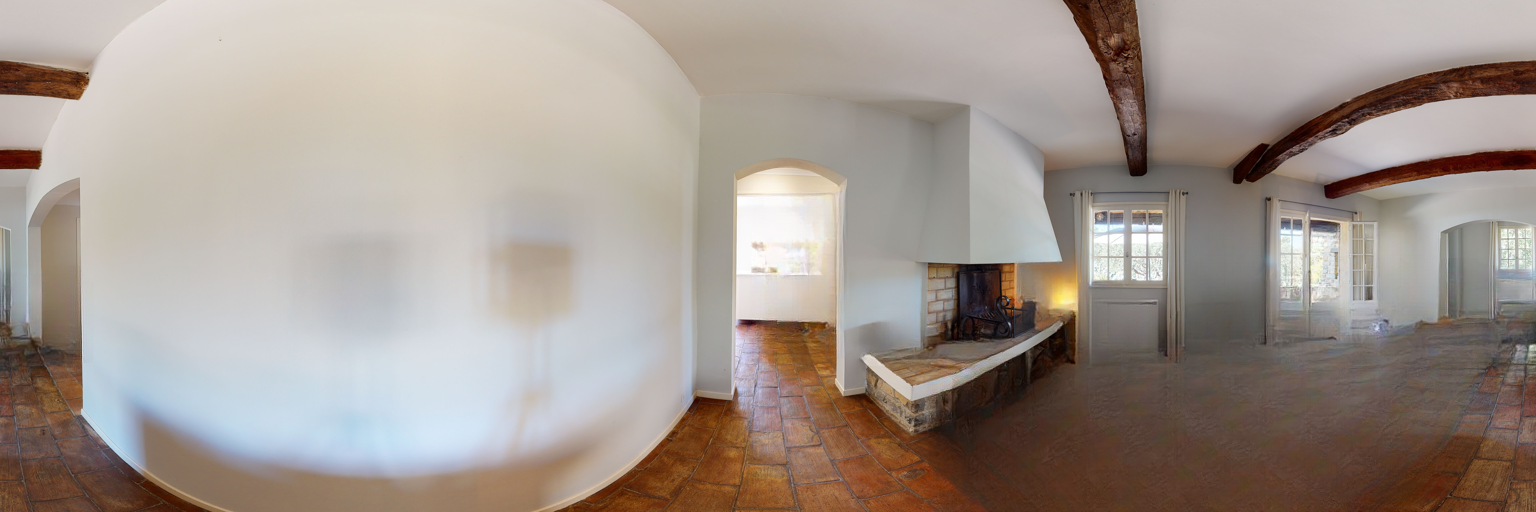}
    \put(2,2){\color{white}\small{LaMa dilation 20px}}
 \end{overpic}\\
 \begin{overpic}[width=0.495\linewidth]{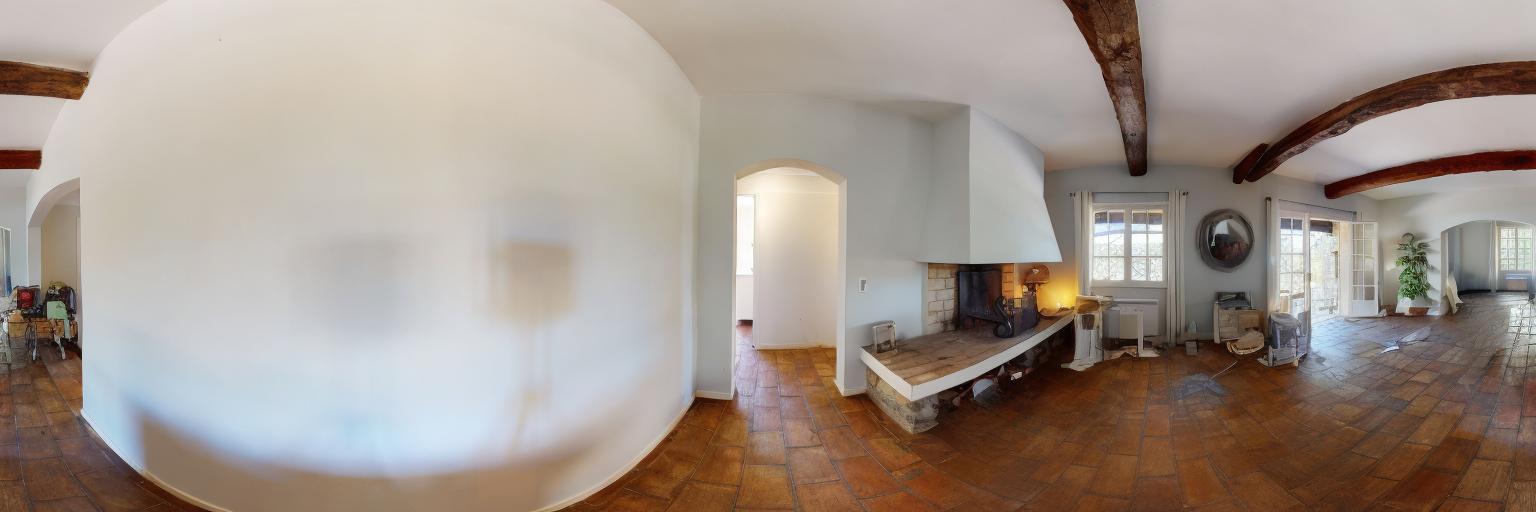}
    \put(2,2){\color{white}\small{SD-2-inpaint no dilation}}
 \end{overpic}~
 \begin{overpic}[width=0.495\linewidth,height=0.165\linewidth]{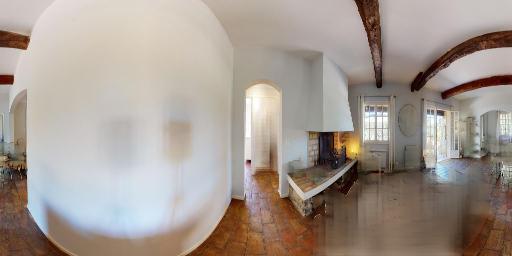}
    \put(2,2){\color{white}\small{LGPN-Net no dilation}}
 \end{overpic}\\
 \begin{overpic}[width=0.495\linewidth]{fig/comp/9985900c24be46bfa6af793fb34e393d_eq_vanilla_10dilate}
    \put(2,2){\color{white}\small{SD-2-inpaint dilation 10px}}
 \end{overpic}~
 \begin{overpic}[width=0.495\linewidth,height=0.165\linewidth]{fig/comp/9985900c24be46bfa6af793fb34e393d_eq_inpainted_lgpn_dilate10}
    \put(2,2){\color{white}\small{LGPN-Net dilation 10px}}
 \end{overpic}\\
 \begin{overpic}[width=0.495\linewidth]{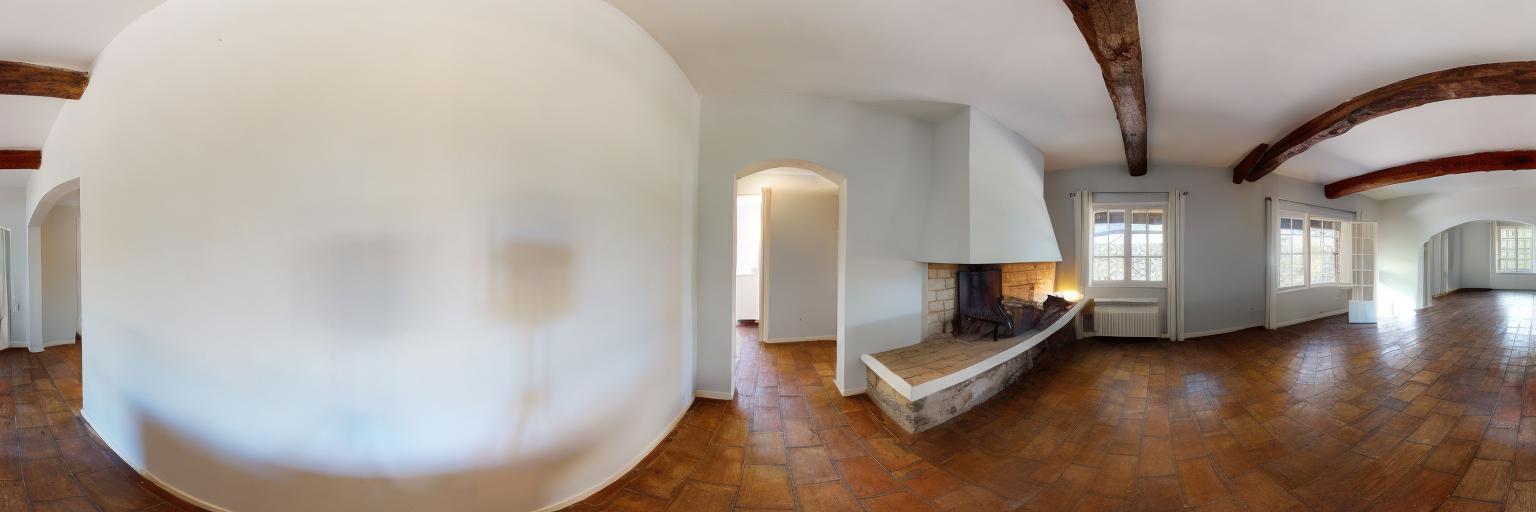}
    \put(2,2){\color{white}\small{SD-2-inpaint dilation 20px}}
 \end{overpic}~
 \begin{overpic}[width=0.495\linewidth,height=0.165\linewidth]{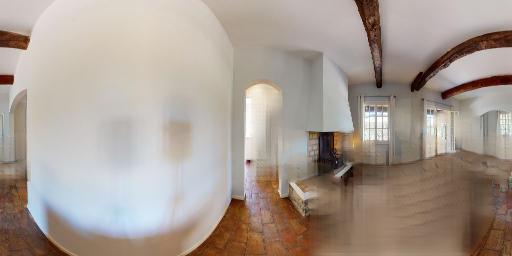}
    \put(2,2){\color{white}\small{LGPN-Net dilation 20px}}
 \end{overpic}
 \caption{\textbf{Additional defurnishing comparisons.} The non-dilated mask is overlaid in blue. Image best viewed digitally.}
\end{figure*}

\begin{figure*}\ContinuedFloat
 \centering
 \begin{overpic}[width=0.495\linewidth]{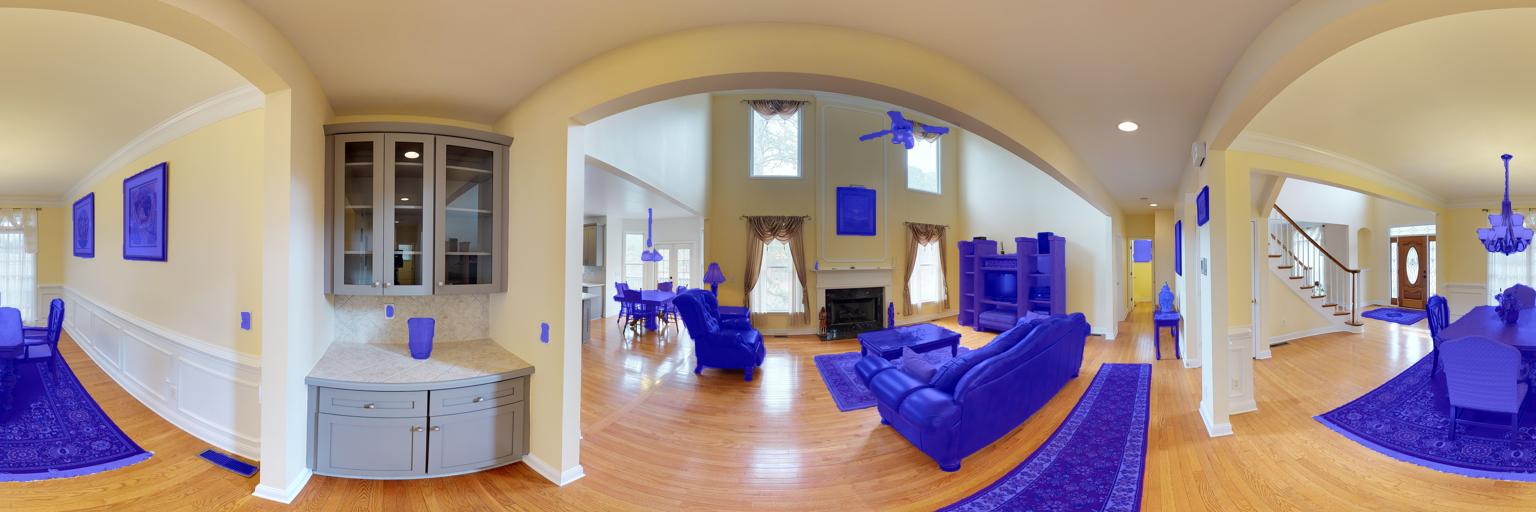}
    \put(2,2){\color{white}\small{Input panorama}}
 \end{overpic}~
 \begin{overpic}[width=0.495\linewidth]{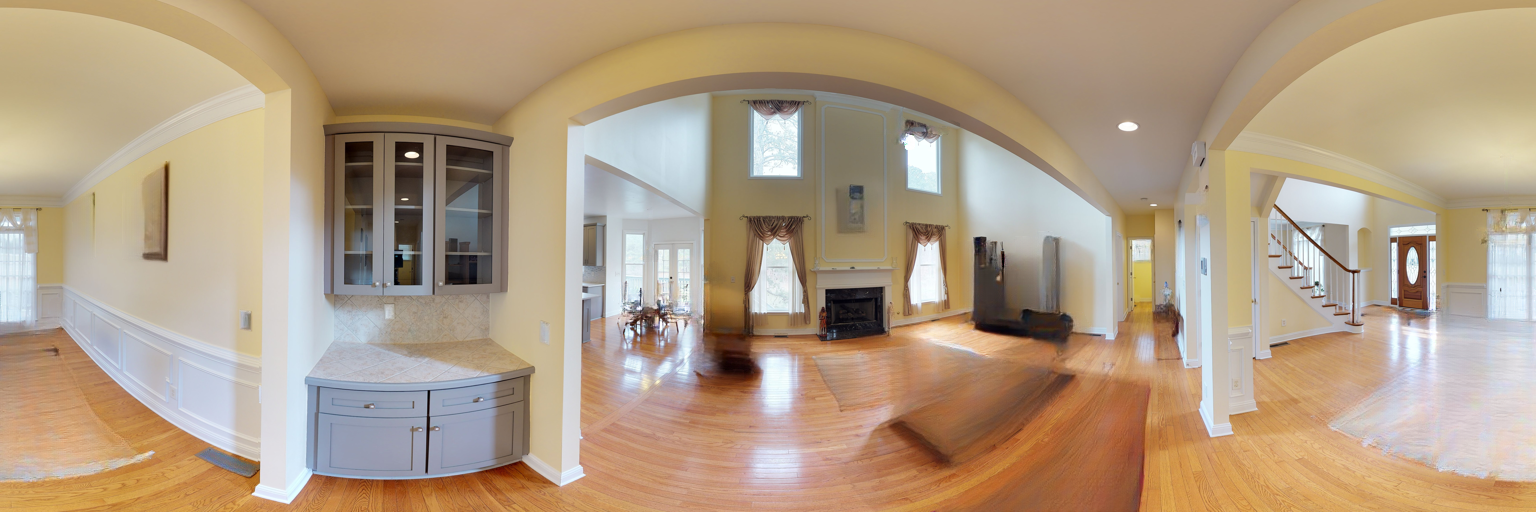}
    \put(2,2){\color{white}\small{LaMa no dilation}}
 \end{overpic}\\
 \begin{overpic}[width=0.495\linewidth]{fig/comp/eb59a0c1b01a4f7085fb3aec815cdde7_eq_inpainted_comparesd}
    \put(2,2){\color{white}\small{Ours-inpaint}}
 \end{overpic}~
 \begin{overpic}[width=0.495\linewidth]{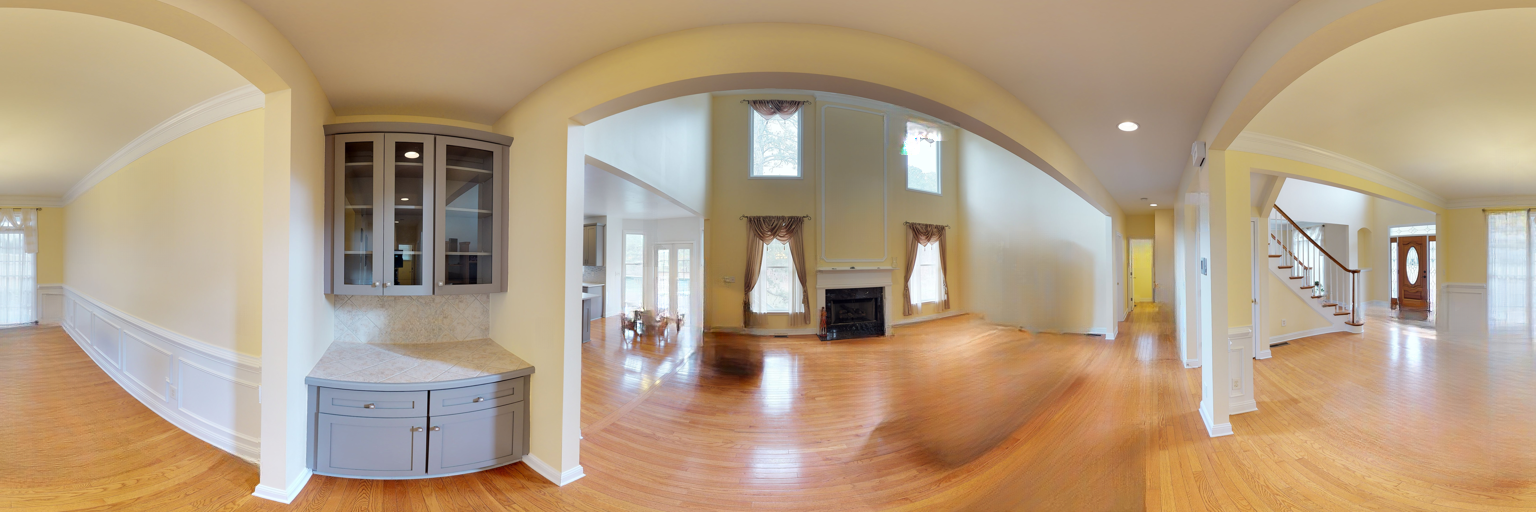}
    \put(2,2){\color{white}\small{LaMa dilation 10px}}
 \end{overpic}\\
 \begin{overpic}[width=0.495\linewidth]{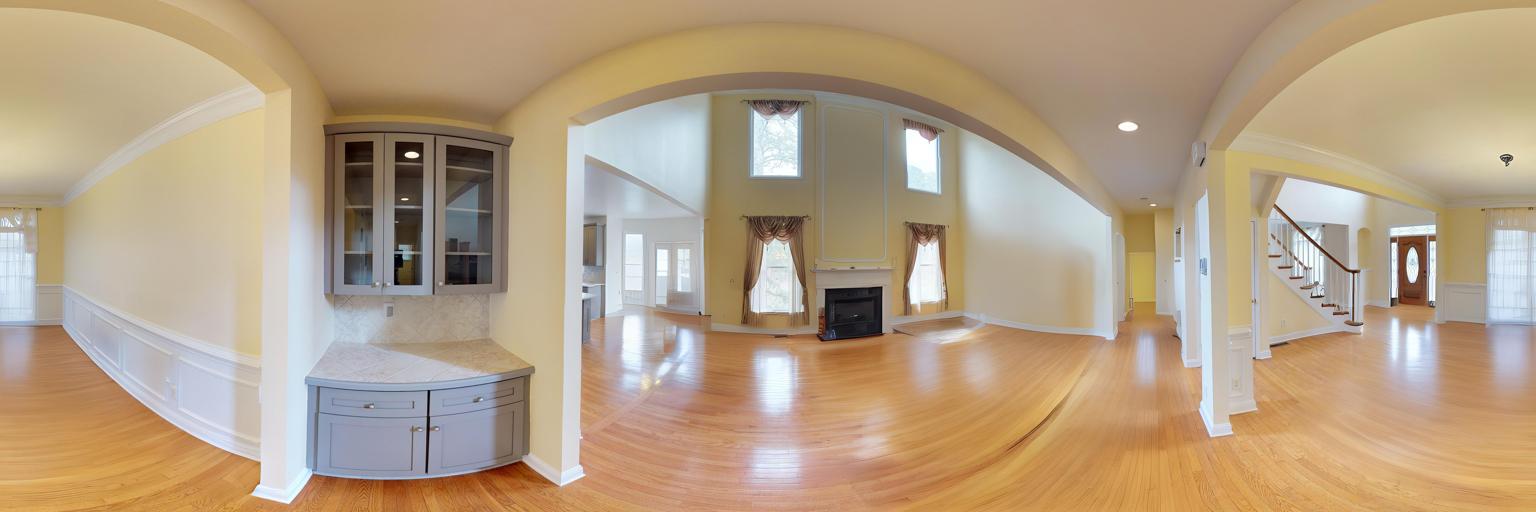}
    \put(2,2){\color{white}\small{Ours-full}}
 \end{overpic}~
 \begin{overpic}[width=0.495\linewidth]{fig/comp/eb59a0c1b01a4f7085fb3aec815cdde7_eq_mask_lama_dilate20_noref}
    \put(2,2){\color{white}\small{LaMa dilation 20px}}
 \end{overpic}\\
 \begin{overpic}[width=0.495\linewidth]{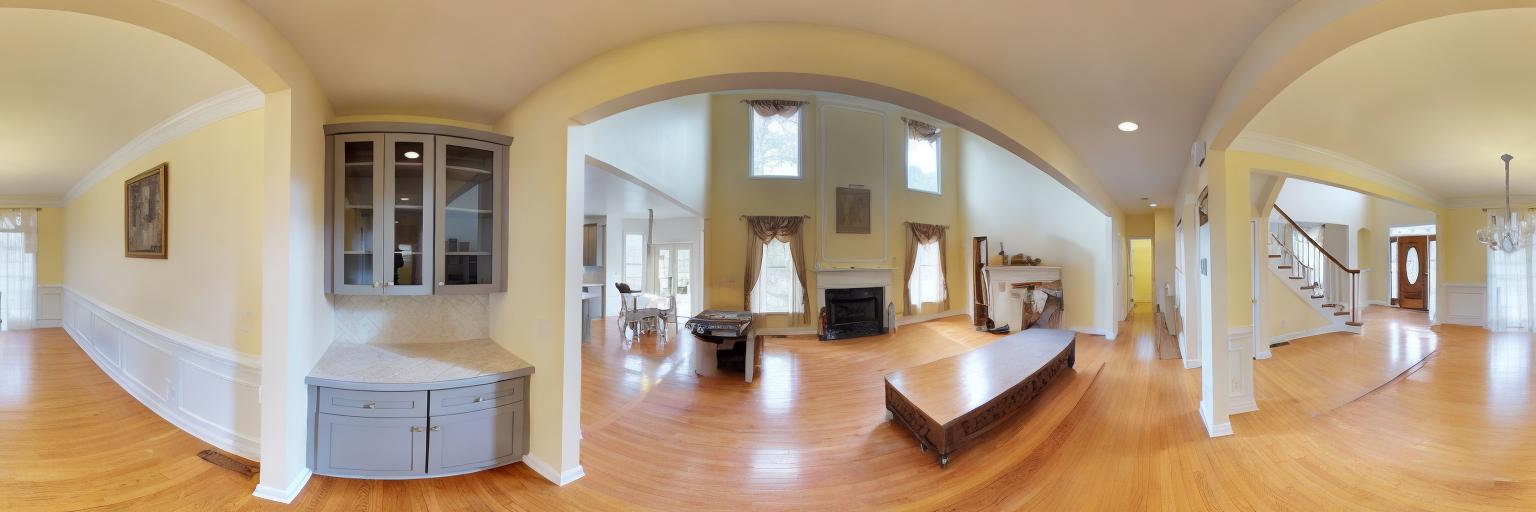}
    \put(2,2){\color{white}\small{SD-2-inpaint no dilation}}
 \end{overpic}~
 \begin{overpic}[width=0.495\linewidth,height=0.165\linewidth]{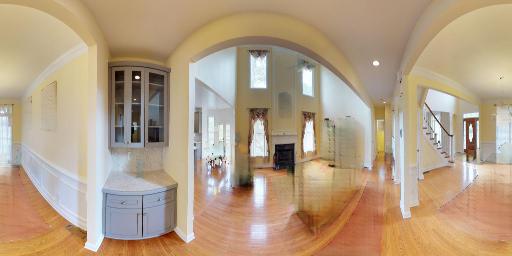}
    \put(2,2){\color{white}\small{LGPN-Net no dilation}}
 \end{overpic}\\
 \begin{overpic}[width=0.495\linewidth]{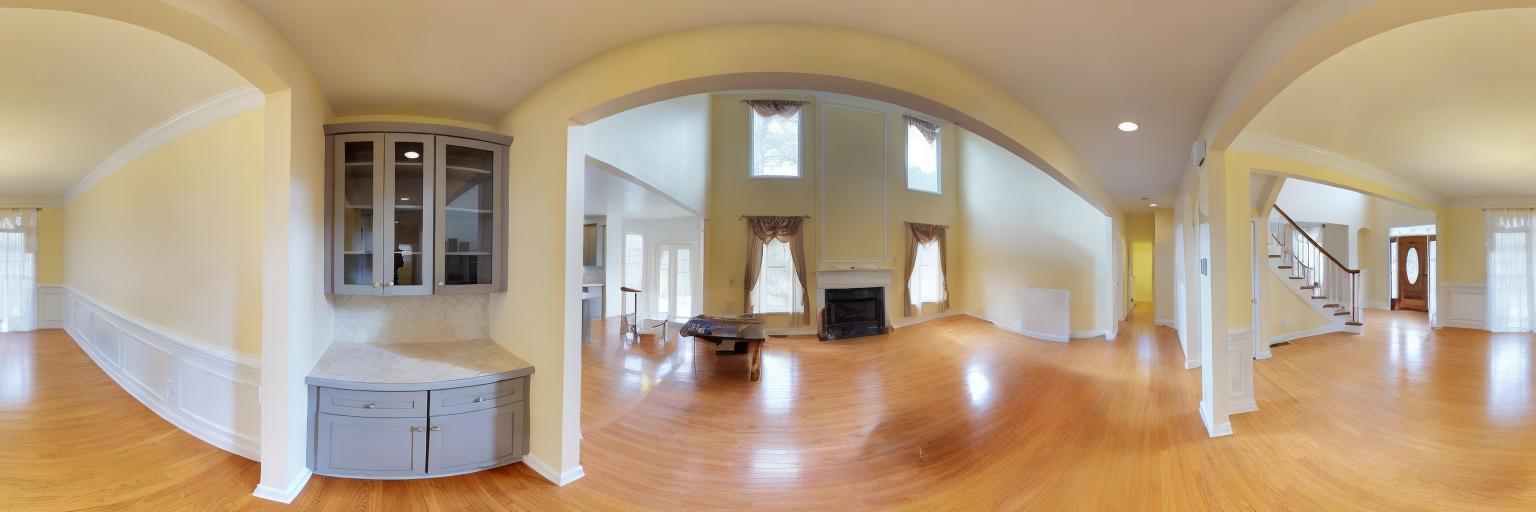}
    \put(2,2){\color{white}\small{SD-2-inpaint dilation 10px}}
 \end{overpic}~
 \begin{overpic}[width=0.495\linewidth,height=0.165\linewidth]{fig/comp/eb59a0c1b01a4f7085fb3aec815cdde7_eq_inpainted_lgpn_dilate10}
    \put(2,2){\color{white}\small{LGPN-Net dilation 10px}}
 \end{overpic}\\
 \begin{overpic}[width=0.495\linewidth]{fig/comp/eb59a0c1b01a4f7085fb3aec815cdde7_eq_vanilla_20dilate}
    \put(2,2){\color{white}\small{SD-2-inpaint dilation 20px}}
 \end{overpic}~
 \begin{overpic}[width=0.495\linewidth,height=0.165\linewidth]{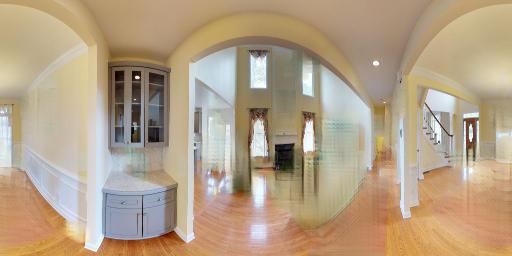}
    \put(2,2){\color{white}\small{LGPN-Net dilation 20px}}
 \end{overpic}
 \caption{\textbf{Additional defurnishing comparisons.} The non-dilated mask is overlaid in blue. Image best viewed digitally.}
\end{figure*}

\end{document}